\PassOptionsToPackage{capitalize}{cleveref}
\documentclass{mlr_template/applemlr}

\usepackage{amsmath}
\usepackage{enumerate}
\usepackage{algorithm}
\usepackage{algpseudocode}
\usepackage{amsfonts}
\usepackage{amsthm}
\usepackage{cleveref}
\usepackage{diagbox}
\usepackage{colortbl}
\usepackage{amssymb}
\usepackage{xspace}
\usepackage{wrapfig}
\usepackage{adjustbox}
\usepackage{tabularx}
\usepackage{booktabs}
\usepackage{mathtools}
\usepackage{tikz}
\usepackage{enumitem}
\usepackage{silence}
\usepackage{dsfont}
\usepackage[table]{xcolor}
\usepackage[dvipsnames]{xcolor}
\usepackage{multirow}
\usepackage{makecell}
\usepackage{xfakebold}
\usepackage{placeins} %
\usepackage{float}
\usepackage{flafter}
\usepackage{pdflscape}

\usepackage{siunitx}

\usepackage{afterpage}
\usepackage{stfloats}

\usepackage{amsmath,amsfonts,bm}

\def\eqref#1{equation~\ref{#1}}
\def\1{\bm{1}}

\DeclareMathAlphabet{\mathsfit}{\encodingdefault}{\sfdefault}{m}{sl}
\SetMathAlphabet{\mathsfit}{bold}{\encodingdefault}{\sfdefault}{bx}{n}

\definecolor{textgray}{HTML}{6E6E73}
\usetikzlibrary{positioning, calc}
\usetikzlibrary{decorations.pathmorphing}

\makeatletter
\patchcmd{\wrong@fontshape}{\@gobbletwo}{}{}{}
\makeatother
\numberwithin{equation}{section}
\makeatletter
\AtBeginDocument{
  \urlstyle{sf}
  
}
\makeatother

\definecolor{light}{RGB}{125, 125, 125}
\crefname{tcb@cnt@pbox}{code}{code}
\Crefname{tcb@cnt@pbox}{Code}{Code}
\crefname{assumption}{assumption}{assumption}
\Crefname{assumption}{Assumption}{Assumptions}
\crefname{appendix}{Appendix}{Appendices}
\Crefname{appendix}{Appendix}{Appendices}

\newtcolorbox[auto counter]{pbox}[2][]{
  colback=white,
  title=Code~\thetcbcounter: #2,
  #1,fonttitle=\sffamily,
  fontupper=\sffamily,
  arc=2pt,
  colframe=bgcolor,
  coltitle=fgcolor,
  colbacktitle=bgcolor,
  toptitle=0.25cm,
  bottomtitle=0.125cm
}

\makeatletter
\newcommand\applefootnote[1]{%
  \begingroup
  \renewcommand\thefootnote{}%
  \renewcommand\@makefntext[1]{\noindent##1}%
  \footnote{#1}%
  \addtocounter{footnote}{-1}%
  \endgroup
}
\makeatother

\definecolor{cverbbg}{gray}{0.90}

\usepackage{fontspec}
\usepackage{microtype}
\usepackage{graphicx}
\usepackage{booktabs,longtable,array,colortbl}
\usepackage{etoolbox}
\DeclareCaptionFont{figonesubcaption}{\fontsize{6.5pt}{7pt}\selectfont}
\DeclareCaptionFont{figmodelcaption}{\fontsize{8pt}{8.5pt}\selectfont}

\makeatletter
\def\@fnsymbol#1{%
  \ensuremath{%
    \ifcase#1
      \or *%
      \or \text{\tiny\faApple}%
      \or \dagger%
      \or \ddagger%
      \or \mathsection%
      \or \mathparagraph%
      \or \|%
      \or **%
      \or \dagger\dagger%
      \or \ddagger\ddagger%
      \else\@ctrerr
    \fi
  }%
}

\makeatother

\makeatletter
\patchcmd{\hyper@makecurrent}{%
    \ifx\Hy@param\Hy@chapterstring
        \let\Hy@param\Hy@chapapp
    \fi
}{%
    \iftoggle{inappendix}{%
        \@checkappendixparam{chapter}%
        \@checkappendixparam{section}%
        \@checkappendixparam{subsection}%
        \@checkappendixparam{subsubsection}%
        \@checkappendixparam{paragraph}%
        \@checkappendixparam{subparagraph}%
    }{}%
}{}{\errmessage{failed to patch}}

\newcommand*{\@checkappendixparam}[1]{%
    \def\@checkappendixparamtmp{#1}%
    \ifx\Hy@param\@checkappendixparamtmp
        \let\Hy@param\Hy@appendixstring
    \fi
}
\makeatletter

\newtoggle{inappendix}
\togglefalse{inappendix}

\title{GRPO Beyond English: A Large-Scale Study of GRPO in Non-English and Multilingual Settings }
\author{Konstantin Dobler$^{1}\!$}
\author{Federico Scozzafava}
\author{Jonathan Janke}
\author{Mohamed Ali}
\author{Simon Lehnerer}

\affiliation{Apple}
\affiliation{$^1$Hasso Plattner Institute \& ELLIS Unit Potsdam (work done during Apple internship)}

\usepackage{catchfile}
\abstract{Reinforcement Learning with Verifiable Rewards (RLVR), often optimized with Group Relative Policy Optimization (GRPO), has become a central recipe for improving the reasoning capabilities of pretrained language models but current studies remain heavily English-centric. 
We conduct a large-scale empirical study of multilingual and non-English GRPO across a wide range of base models, training languages, and different reasoning language rewards. We find that training to reason in the native language often leaves only a small gap to training for English reasoning. We further observe strong crosslingual transfer: training in one language often improves performance in many others. However, specific trends are highly model- and language-dependent. In some cases, training in a particular language induces severe regressions on out-of-domain capabilities in other languages. Our analysis shows that RLVR beyond English can provide broad crosslingual gains, but also requires broad evaluation to detect language-specific regressions.
}

{
\metadata[Correspondence]{\sffamily konstantin.dobler@hpi.de}
\date{\sffamily August 13, 2026}
}
\begin{document}

\maketitle

\section{Introduction}
Long chain-of-thought (CoT) reasoning has become a defining capability of contemporary large language models (LLMs). A major driver of recent progress is reinforcement learning with verifiable rewards (RLVR), where models explore diverse solution trajectories and are optimized using automatically checkable outcome feedback such as exact-match numeric answers. In this setting, Group Relative Policy Optimization \citep[GRPO]{shao2024deepseekmath} has emerged as an effective training method.

Despite the multilingual pre-training of many open models, GRPO/RLVR post-training is still studied primarily in English. 
Recent evidence suggests that, under RLVR, reasoning traces may drift toward a dominant pre-training language, often English, even when prompts are posed in other languages \citep{park2025crosslingualcollapselanguagecentricfoundation}. 
At the same time, work on test-time control finds that forcing models to reason in a non-English target language can improve interpretability and language faithfulness, but often at a severe cost in accuracy \citep{qi2025whenmodelsreason,yong2025crosslingualreasoningtesttimescaling}. 

In this work, we study GRPO training in multilingual and non-English settings and analyze crosslingual transfer, the effect of reasoning languages and multilingual training.
Crucially, we run experiments with broad coverage across three model families, including both ``\texttt{-Base}'' and post-trained variants as well as multiple model sizes. We run multilingual training experiments as well as separate monolingual trainings covering 11 different languages and nine different models.
As we will show in \cref{sec:results}, this large-scale setup reveals that the effects of GRPO can be highly model- and language-specific, and that conclusions drawn from a single model or a single training language can be misleading.
In summary, our contributions are:
\begin{itemize}[leftmargin=*,itemsep=1pt,topsep=2pt]
\item We conduct a large-scale controlled study of multilingual and non-English GRPO across nine base models, 11 training languages with monolingual and multilingual training settings, and English- and native-reasoning rewards.
\item We find that rewarding reasoning in a non-English language during training leaves only a small gap to English reasoning rewards on average and that substantial crosslingual transfer occurs across a large range of models and languages under GRPO.
\item We observe a language-dependent regression failure mode of GRPO with non-English prompt or rewarded reasoning languages: specific model-language combinations can cause large regressions on other unseen languages or tasks, even when performance is improved elsewhere.
\end{itemize}

\section{Related Work}

A directly relevant line of work studies how the language of reasoning affects performance of reasoning models. Inference-time interventions and analyses of multilingual reasoning reveal a tension: reasoning in the user's language can be an important desideratum for readability, whereas English or mixed-language reasoning often retains higher task accuracy \citep{qi2025whenmodelsreason,yong2025crosslingualreasoningtesttimescaling,tam2025languagemattersmultilingualinput,wang-etal-2025-language-mixing}. A similar tension appears during RLVR, where reasoning may drift toward a dominant pretraining language or language-mixing and language-consistency objectives are consequently used to steer the model toward the desired language \citep{park2025crosslingualcollapselanguagecentricfoundation,deepseekr1,hwang2025learngloballyspeaklocally,zhang2025mthinker}. Conversely, recent work treats variation in reasoning language as a source of exploration rather than solely as a failure of control \citep{wu2026polygrpo}. Together, these findings motivate our controlled comparison of English- and native-reasoning rewards across a broad range of models and languages under a common GRPO setup.

A second direction transfers strong English reasoning capabilities to other languages. Existing approaches differ operationally---through various methods such as translations, reference traces, judges, or teacher distributions---but all use different forms of crosslingual supervision signals to bridge the gap between English and non-English reasoning \citep{she2024mapo,faisal2025pbrlsvr,sutawika2026sp3f,zhang2025mthinker,huang2026tapo,liu2026trit,liu2026copsd}. In contrast, our study does not introduce a new alignment or translation mechanism and instead holds the GRPO training method fixed while systematically varying the prompt and rewarded reasoning languages.
Recent studies further indicate that RL-based post-training can transfer improvements across languages, including languages excluded from training, and in some settings generalizes more effectively than supervised fine-tuning \citep{hwang2025learngloballyspeaklocally,zhang2025mthinker,huang2026tapo,elhady2026crosslingualselfconsistency}.
Most closely related, \citet{huang2026beyond} also investigate crosslingual transfer and the impact of reasoning language under GRPO, but focus on a comparison of SFT and GRPO with a smaller scope of training languages and two base models.
Most prior studies either propose a particular transfer method or evaluate a comparatively small set of models or training languages, while our study covers a wide range of models and languages.
We discuss broader related work in \cref{app:related-work}.

\section{Experimental Setup}
\paragraph{Datasets.} A large-scale study of GRPO in non-English training settings requires large-scale available multilingual datasets suitable for RLVR training. 
We use the Multilingual Reasoning Gym \citep{dobler2026multilingualreasoninggym}, which can procedurally generate ``infinite'' data samples of varying difficulty for 14 different languages.
GRPO training is especially sensitive to matching the difficulty of problems in the data to the model capability \citep{an2025polaris}. Merely translating an English dataset is therefore not necessarily optimal regardless of data scarcity: models exhibit different capabilities in different languages; therefore, ``optimal'' difficulty in English might be too hard in other languages.
The Multilingual Reasoning Gym provides a difficulty curriculum for each task that can be adjusted based on the current model's performance. Specifically, we initialize all tasks at the lowest difficulty and increase the difficulty for a particular task whenever more than 70\% of the last 32 rollouts were correct (and also allow a decrease if less than 10\% of these were correct).

However, even though the difficulty can be adjusted, some tasks are too difficult for smaller models even at the lowest difficulty levels. Therefore, we do an initial pass over all tasks at an easy (default) difficulty level using \texttt{Qwen3-14B-Base}. We filter out all tasks on which \texttt{Qwen3-14B-Base} has a pass@8 score of zero. Additionally, we construct a set of held-out tasks by sampling tasks from each task domain (\textit{e.g.}, arithmetic, graphs, logic, geometry).
We also exclude all tasks from training that contain heavy English content (\textit{e.g.}, English word problems) and use these only for evaluation.
This leaves us with a total of 62 tasks used for training in our main training configuration and 30 tasks unseen during training used for out-of-distribution evaluation (see \cref{app:tasklist}).

\paragraph{Models.} Recent works have shown that RL training performance heavily depends on the base model used \citep{shao2026spuriousrewardsrethinkingtraining}. Thus, we run experiments on a broad range of models from the \texttt{Qwen}, \texttt{Gemma} and \texttt{SmolLM} families: \texttt{Qwen3-Base-\{1.7B/4B/8B\}}, \texttt{Qwen3-\{1.7B/4B/8B\}}, \texttt{SmolLM3-3B}, and \texttt{gemma-3-\{1b/4b\}-it} \citep{yang2025qwen3technicalreport, bakouch2025smollm3, gemmateam2025gemma3technicalreport}.
This setup includes not only different model families, but also models at varying degrees of post-training, different model sizes up to 8 billion parameters and varying numbers of covered languages during pretraining, resulting in a total of 9 different base models.

\paragraph{Training.}
We run RL training for 500 steps in each language, with a learning rate of \num{1e-6}, 64 prompts per step and 8 rollouts per prompt (leading to a total batch size per step of 512).
We use the DAPO-style loss formulation \citep{yu2025dapoopensourcellmreinforcement} without a KL-divergence penalty.
Following standard RLVR practice, we give a binary correctness reward when the extracted answer is verified as correct \citep{lambert2025tulu3pushingfrontiers,kimiteam2025kimik15scalingreinforcement,yu2025dapoopensourcellmreinforcement,he2025skyworkopenreasoner1}. Only when the answer is correct, we give up to two additional rewards, each weighted at one half of the correctness reward: (1) a format reward for adhering to the specified output format and (2) a reasoning language reward which either rewards matching the prompt language (``native reasoning'') or reasoning in English (``English reasoning''). This setup allowed for robust training across languages and models, avoiding collapse or overfitting to the format reward observed in preliminary experiments for low-resource languages.

In this work, we want to study the effects and interplay of training data language as well as reasoning language. To uncover model- and training-language-specific phenomena, we run monolingual trainings in the 11 different languages which are present in MGSM (English, Chinese, German, French, Spanish, Russian, Japanese, Thai, Swahili, Telugu and Bengali) and a multilingual run where a sample's language is uniformly sampled from that set. This reserves three languages of the Multilingual Reasoning Gym (Italian, Korean and Portuguese) as completely heldout languages. 
We train English-reward variants for all training languages. For model–language pairs with sufficient initial support for the (non-English) target reasoning language (see \cref{sec:reasoning-lang-effect-exclusions}), we additionally train a native-reward variant. This yields up to 23 trained variants per base model.
We report further details in \cref{app:technical-details}.  

\paragraph{Evaluation.} 
\phantomsection
\label{sec:eval-setup}

We evaluate all trained checkpoints on unseen samples of seen tasks of the Multilingual Reasoning Gym \citep{dobler2026multilingualreasoninggym}, the 30 unseen tasks of the Multilingual Reasoning Gym, and the MGSM \citep{shi2022mgsm} and PolyMath \citep{wang2025polymath} benchmarks. PolyMath comprises four different difficulty levels, where the lowest level is roughly comparable to that of MGSM. Indeed, we see that performance on the low-difficulty split of PolyMath and MGSM is often similar, while the more difficult splits of PolyMath sometimes show starkly different trends (see our analysis in \cref{sec:crosslingual-transfer-forgetting}). Thus, we group MGSM and the low difficulty split of PolyMath as ``Easy Math'' and the remaining PolyMath splits as ``Hard Math''.
We evaluate each checkpoint on all available languages in order to study crosslingual transfer effects.
We use 8 rollouts per prompt for all tasks and report the avg@8 performance. We use a maximum rollout length of 16k tokens and the recommended sampling parameters. We report further details in \cref{app:technical-details}.

\begin{figure*}
    \centering
    \begin{subfigure}[t]{0.49\linewidth}
        \centering
        \caption{English reasoning reward}
        \includegraphics[width=\linewidth, trim={0pt 5pt 0pt 0pt}, clip]{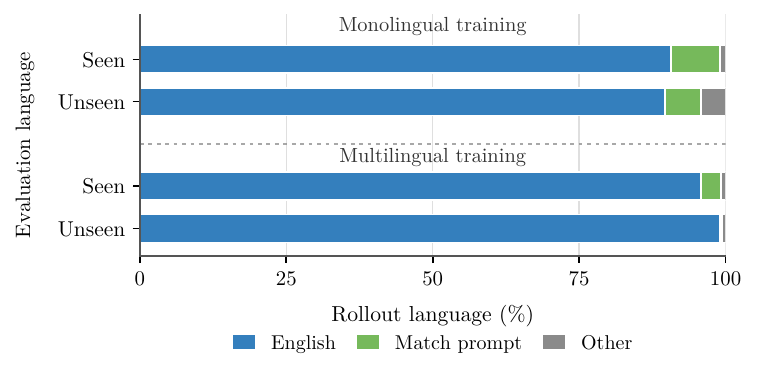}
        \label{fig:reasoning-language-adherence-english-reward}
    \end{subfigure}
    \hfill
    \begin{subfigure}[t]{0.49\linewidth}
        \centering
        \caption{Native reasoning reward}
        \includegraphics[width=\linewidth, trim={0pt 5pt 0pt 0pt}, clip]{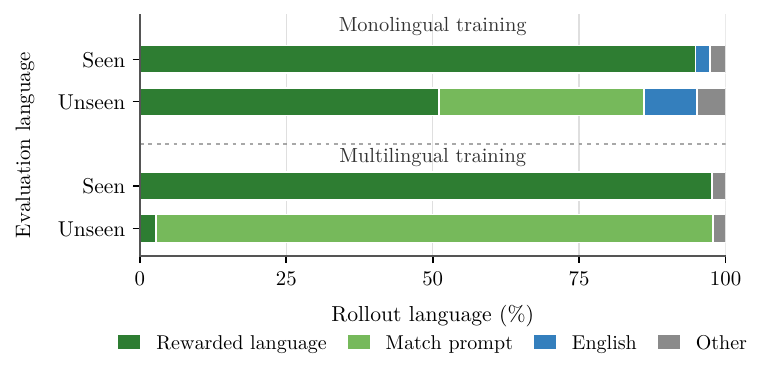}
        \label{fig:reasoning-language-adherence-native-reward}
    \end{subfigure}
    \vspace{-2em}

    \caption{Reasoning languages detected during evaluation.
    We classify into four mutually exclusive buckets: (1) the rewarded language(s), (2) the prompt language (if different from rewarded language), (3) English (if different from both), and (4) all other languages. Results are averaged over all evaluated benchmarks and training languages.}
    \label{fig:reasoning-language-adherence-stacks}
\end{figure*}

\section{Results}
\label{sec:results}

We first discuss the effect of reasoning language in \cref{sec:reasoning-lang-effect}, then crosslingual transfer and multilingual GRPO training in \cref{sec:crosslingual-transfer}, and finally surprising language-specific capability regressions in \cref{sec:crosslingual-transfer-forgetting}.

\subsection{Effect of Reasoning Language}
\label{sec:reasoning-lang-effect}

Prior work has shown that \textit{native-language reasoning} degrades performance of reasoning models \citep{yong2025crosslingualreasoningtesttimescaling, qi2025whenmodelsreason}. Small-scale training can reliably ensure language consistency in reasoning traces, but does not eliminate the performance penalty \citep{qi2025whenmodelsreason}. 
However, these works use no or only very small-scale training in the tested reasoning languages. We therefore investigate whether larger-scale RL training with non-English reasoning languages can close the gap. We implement this via reward shaping: we compare English-reasoning rewards with native-language rewards, which reward reasoning in the prompt language. This approach is chosen to avoid confounders from techniques such as assistant hijacking \citep{yong2025crosslingualreasoningtesttimescaling}.

We distinguish the \textit{training language} used in RLVR prompts from the \textit{reasoning language} used in generated reasoning traces. An evaluation language is \textit{seen} if it occurred during training and \textit{unseen} otherwise.

\paragraph{Language consistency after reasoning language rewards.}
\phantomsection
\label{sec:reasoning-lang-effect-exclusions}
On seen evaluation languages, both rewards strongly control the reasoning language: 90.7\% of rollouts from monolingually trained English-reward models use English, while 94.9\% of native-reward models use the training language. On unseen languages, English-reward models still reason predominantly in English after monolingual (89.7\%) and multilingual training (99.1\%). Monolingual native-reward models divide their reasoning between the training language (51.0\%), unseen prompt language (35.0\%), and English (9.0\%). Multilingual native-reward models consistently match the prompt language on both seen and unseen languages (\cref{fig:reasoning-language-adherence-stacks}).

A reasoning-language reward needs the initial model to at least sometimes use the target language. We thus apply native-language rewards only to model--language pairs where at least 2.5\% of initial traces use that language. \texttt{Qwen3-Base} models consistently include native language reasoning. For \texttt{SmolLM-3B} and most \texttt{Qwen3} (non-``Base'') models, this retains Chinese, while German and Russian are also retained for \texttt{Qwen3-8B}. \Cref{fig:reasoning-language-adherence-native-reward} includes only these valid model-language combinations for the native-reward training, while English rewards are trained for all pairs. Full details are in \cref{app:matrices}.
For a small set of training languages, \texttt{Qwen3-4B-Base} and \texttt{gemma-3-1b-it} continue to match the prompt language despite English-reward training. Interestingly, when English rewards are effective, they deliver consistent benchmark improvements for both models but yield performance close to native rewards when ineffective (see \cref{app:matrices-qwen3-4b-base-sec}). We analyze these performance benefits of English reasoning over native-language reasoning in more detail next. All comparisons between English and native reasoning rewards are restricted to model-language pairs for which both rewards successfully control the rollout language\footnote{See also \cref{app:reasoning-reward-unfiltered} for discussion.}.

\begin{figure*}[htbp]
    \centering
    \begin{minipage}[t]{0.49\linewidth}
        \centering
        \includegraphics[width=\linewidth]{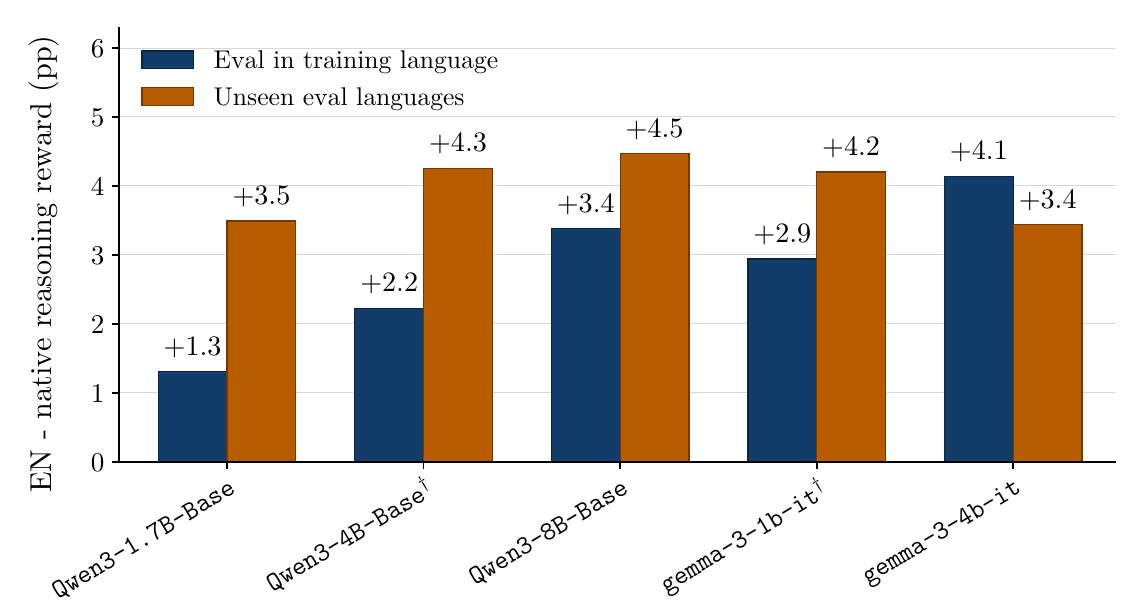}
        \vspace{-1.7em}
        \caption{Per model advantage (avg@8) of English over native-language reasoning rewards, averaged across all benchmarks and (monolingual) training languages for each model. 
        Seen and unseen evaluation languages are reported separately.
        \textsuperscript{$\dagger$}Model average excludes some training languages; see \cref{sec:reasoning-lang-effect-exclusions}.}
        \label{fig:reason_reward_model_summary}
    \end{minipage}
    \hfill
    \begin{minipage}[t]{0.49\linewidth}
        \centering
        \includegraphics[width=\linewidth]{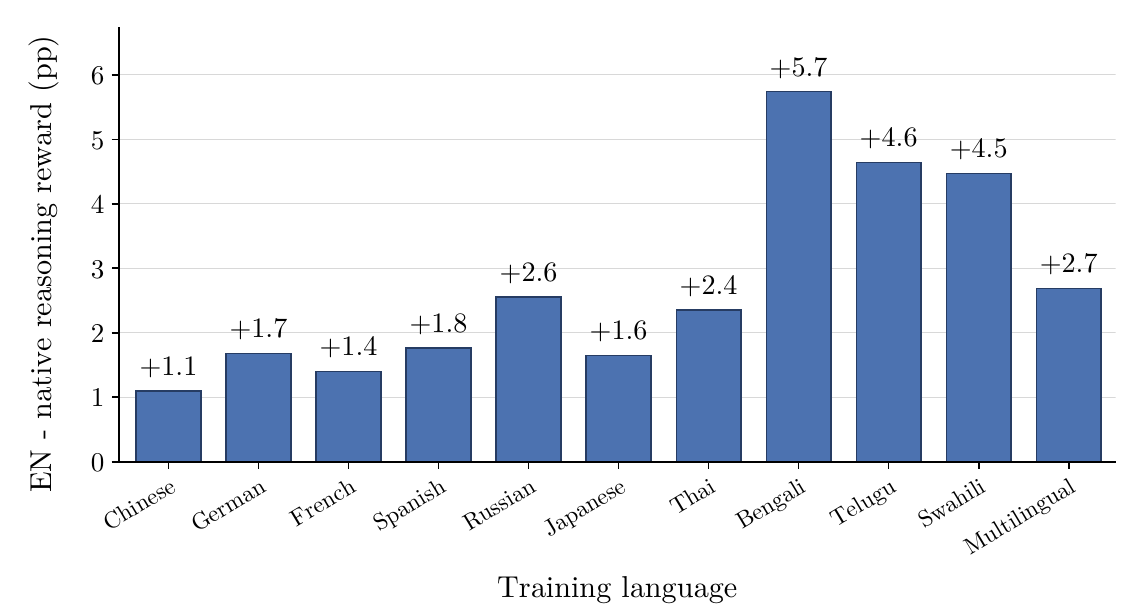}
        \vspace{-1.7em}
        \caption{Per training-language advantage (avg@8) of English over native-language reasoning rewards on seen evaluation languages, averaged over the models from \cref{fig:reason_reward_model_summary}. The multilingual bar averages over all 11 evaluation languages seen during multilingual training.}
        \label{fig:reason_reward_seen_language_summary}
    \end{minipage}

\end{figure*}

\paragraph{Performance comparison of reasoning language rewards.} 
In \cref{fig:reason_reward_model_summary,fig:reason_reward_seen_language_summary}, we compare training with English and native-language reasoning rewards. Indeed, English reasoning rewards perform better on average. On languages seen during training, the mean advantage ranges from 1.3pp to 4.1pp across models.
The effect is not uniform across languages: Bengali, Telugu, and Swahili exhibit substantially larger English-reward advantages than French, Chinese, Japanese, or German.
This suggests that training models to reason in English can partly compensate for weaker target-language reasoning capabilities.
Notably, the absolute difference is smaller than in previous work without sustained native-language reasoning training and might provide an acceptable performance trade-off if native-language reasoning is an important desideratum. While we observe comparatively ``small'' deltas, especially on higher-resource languages, \citet{qi2025whenmodelsreason} report up to double-digit accuracy drops. \citet{huang2026beyond} similarly find drops of up to 10pp from native-language rewards in Chinese or German, but combine these rewards with language-forcing prompts and limited model and language coverage.
While the aggregate differences in our broader evaluation are smaller, specific model--language combinations can still show large English-reward advantages. 
For example, \cref{fig:reward-specific-forgetting-panels:gemma-1b-easy} shows double-digit advantages for English reasoning with \texttt{gemma-3-1b-it} on MGSM and PolyMath-low (``Easy Math''), especially for Spanish, Thai, and Swahili. Multilingual training shows a smaller but still substantial 6.6pp advantage for English reasoning.

Note that native-reward models consistently reason in the prompt language on seen languages, but may switch between prompt and training languages on unseen languages (\cref{fig:reasoning-language-adherence-native-reward}). English-reward models continue to reason in English (\cref{fig:reasoning-language-adherence-english-reward}). Their advantage on unseen languages may therefore partly reflect a ``train-evaluation mismatch'' by native-reward models, which sometimes reason in a language not used during training.
However, English-reward advantages persist on seen languages, where this mismatch does not occur, and under multilingual training, where native-reward models were trained to reason in every evaluation language used in \cref{fig:reason_reward_seen_language_summary}.
We leave an investigation of more aggressive inference-time interventions to enforce reasoning in the prompt language \citep{yong2025crosslingualreasoningtesttimescaling} to future work.

\subsection{Crosslingual Transfer Under GRPO}
\label{sec:crosslingual-transfer}

\begin{table*}
\centering
\begin{subtable}[t]{0.53\linewidth}
\centering
\caption{\textbf{In-domain tasks:} Avg@8 improvement over the base model, in percentage points, on Multilingual Reasoning Gym tasks \textit{seen during training}.}
\label{tab:pullin_crosslingual_matrix_reward_average_rg_heldout_train}
\scriptsize
\setlength{\tabcolsep}{3pt}
\renewcommand{\arraystretch}{1.15}
\newcommand{\MatrixResizeWidth}{\linewidth}
\resizebox{\MatrixResizeWidth}{!}{%
\begin{tabular}{lrrrrrrrrrrr@{\hspace{1.2em}}rrr}
\toprule
Train $\downarrow$ / Eval $\rightarrow$  & en & zh & de & fr & es & ru & ja & th & bn & te & sw & it & ko & pt \\
\midrule
en & \cellcolor[HTML]{347FBD}\underline{\textbf{25.9}} & \cellcolor[HTML]{B4D2EA}20.5 & \cellcolor[HTML]{8AB7DB}22.9 & \cellcolor[HTML]{9AC1E1}21.8 & \cellcolor[HTML]{9AC1E1}22.7 & \cellcolor[HTML]{A1C6E3}21.1 & \cellcolor[HTML]{A1C6E3}21.5 & \cellcolor[HTML]{B2D1E9}22.6 & \cellcolor[HTML]{D5E7F6}20.7 & \cellcolor[HTML]{DAEAF7}18.9 & \cellcolor[HTML]{DCECF8}17.3 & \cellcolor[HTML]{9DC3E2}22.1 & \cellcolor[HTML]{86B5DA}23.5 & \cellcolor[HTML]{93BDDE}21.6 \\
\midrule
zh & \cellcolor[HTML]{9FC4E3}22.4 & \cellcolor[HTML]{347FBD}\underline{\textbf{27.5}} & \cellcolor[HTML]{87B5DA}23.1 & \cellcolor[HTML]{92BCDE}22.1 & \cellcolor[HTML]{9CC3E2}22.6 & \cellcolor[HTML]{82B2D8}22.8 & \cellcolor[HTML]{609CCD}24.6 & \cellcolor[HTML]{9BC2E1}23.7 & \cellcolor[HTML]{B6D3EB}22.6 & \cellcolor[HTML]{C7DEF1}20.3 & \cellcolor[HTML]{D1E5F4}18.3 & \cellcolor[HTML]{A6C9E5}21.8 & \cellcolor[HTML]{3F86C1}26.1 & \cellcolor[HTML]{8CB8DC}21.9 \\
\midrule
de & \cellcolor[HTML]{76AAD4}23.8 & \cellcolor[HTML]{A8CAE6}21.1 & \cellcolor[HTML]{347FBD}\underline{\textbf{26.4}} & \cellcolor[HTML]{6DA4D1}23.5 & \cellcolor[HTML]{83B2D9}23.8 & \cellcolor[HTML]{8CB8DC}22.3 & \cellcolor[HTML]{83B2D9}22.9 & \cellcolor[HTML]{97BFE0}23.9 & \cellcolor[HTML]{C5DDF0}21.6 & \cellcolor[HTML]{CEE3F3}19.8 & \cellcolor[HTML]{C5DDF0}19.4 & \cellcolor[HTML]{71A7D3}23.6 & \cellcolor[HTML]{639ECE}24.8 & \cellcolor[HTML]{79ABD5}22.7 \\
\midrule
fr & \cellcolor[HTML]{8DB9DC}23.0 & \cellcolor[HTML]{ADCEE7}20.8 & \cellcolor[HTML]{6CA4D1}24.2 & \cellcolor[HTML]{347FBD}\underline{\textbf{25.6}} & \cellcolor[HTML]{74A8D3}24.5 & \cellcolor[HTML]{85B4DA}22.6 & \cellcolor[HTML]{90BBDD}22.3 & \cellcolor[HTML]{9BC2E1}23.7 & \cellcolor[HTML]{C3DCEF}21.8 & \cellcolor[HTML]{CBE1F2}20.0 & \cellcolor[HTML]{C9E0F1}19.0 & \cellcolor[HTML]{4A8EC5}25.0 & \cellcolor[HTML]{78ABD5}24.1 & \cellcolor[HTML]{609CCD}23.8 \\
\midrule
es & \cellcolor[HTML]{70A6D2}24.0 & \cellcolor[HTML]{99C1E1}21.9 & \cellcolor[HTML]{5E9BCC}24.7 & \cellcolor[HTML]{4F90C6}24.6 & \cellcolor[HTML]{347FBD}\underline{\textbf{27.5}} & \cellcolor[HTML]{76AAD4}23.4 & \cellcolor[HTML]{81B1D8}23.0 & \cellcolor[HTML]{91BBDE}24.2 & \cellcolor[HTML]{B3D1EA}22.8 & \cellcolor[HTML]{C6DEF0}20.4 & \cellcolor[HTML]{CDE2F3}18.6 & \cellcolor[HTML]{347FBD}\underline{\textbf{25.8}} & \cellcolor[HTML]{5393C8}25.4 & \cellcolor[HTML]{347FBD}\underline{\textbf{25.6}} \\
\midrule
ru & \cellcolor[HTML]{7CAED6}23.6 & \cellcolor[HTML]{8DB9DC}22.6 & \cellcolor[HTML]{73A8D3}23.9 & \cellcolor[HTML]{72A7D3}23.3 & \cellcolor[HTML]{83B2D9}23.8 & \cellcolor[HTML]{347FBD}\underline{\textbf{26.9}} & \cellcolor[HTML]{77ABD5}23.5 & \cellcolor[HTML]{92BCDE}24.1 & \cellcolor[HTML]{AECEE8}23.1 & \cellcolor[HTML]{C5DDF0}20.5 & \cellcolor[HTML]{CAE0F2}18.9 & \cellcolor[HTML]{79ABD5}23.4 & \cellcolor[HTML]{3A83BF}26.3 & \cellcolor[HTML]{659FCE}23.6 \\
\midrule
ja & \cellcolor[HTML]{ABCCE7}22.0 & \cellcolor[HTML]{8FBADD}22.5 & \cellcolor[HTML]{99C1E1}22.4 & \cellcolor[HTML]{A4C8E4}21.4 & \cellcolor[HTML]{ACCDE7}21.9 & \cellcolor[HTML]{96BEDF}21.7 & \cellcolor[HTML]{347FBD}\underline{\textbf{26.8}} & \cellcolor[HTML]{A9CBE6}23.0 & \cellcolor[HTML]{AACBE6}23.4 & \cellcolor[HTML]{BBD7ED}21.2 & \cellcolor[HTML]{CAE0F2}18.9 & \cellcolor[HTML]{A3C7E4}21.9 & \cellcolor[HTML]{3C84C0}26.3 & \cellcolor[HTML]{99C0E0}21.4 \\
\midrule
th & \cellcolor[HTML]{98C0E0}22.7 & \cellcolor[HTML]{9DC3E2}21.7 & \cellcolor[HTML]{76AAD4}23.8 & \cellcolor[HTML]{89B6DB}22.4 & \cellcolor[HTML]{96BEDF}22.9 & \cellcolor[HTML]{8CB8DC}22.3 & \cellcolor[HTML]{72A7D3}23.7 & \cellcolor[HTML]{347FBD}\underline{\textbf{28.6}} & \cellcolor[HTML]{AECEE8}23.1 & \cellcolor[HTML]{C3DCEF}20.6 & \cellcolor[HTML]{C2DBEF}19.6 & \cellcolor[HTML]{88B6DB}22.9 & \cellcolor[HTML]{4288C2}26.1 & \cellcolor[HTML]{8DB9DC}21.9 \\
\midrule
bn & \cellcolor[HTML]{B0CFE8}21.9 & \cellcolor[HTML]{A2C6E4}21.5 & \cellcolor[HTML]{97BFE0}22.4 & \cellcolor[HTML]{9BC2E1}21.8 & \cellcolor[HTML]{ABCCE7}21.9 & \cellcolor[HTML]{8BB7DB}22.3 & \cellcolor[HTML]{7BADD6}23.3 & \cellcolor[HTML]{A2C7E4}23.4 & \cellcolor[HTML]{347FBD}\underline{\textbf{30.7}} & \cellcolor[HTML]{A1C6E3}23.1 & \cellcolor[HTML]{BFD9EE}19.9 & \cellcolor[HTML]{9BC2E1}22.2 & \cellcolor[HTML]{488CC4}25.8 & \cellcolor[HTML]{8EBADD}21.8 \\
\midrule
te & \cellcolor[HTML]{D2E5F4}20.8 & \cellcolor[HTML]{B7D4EB}20.3 & \cellcolor[HTML]{9EC4E2}22.2 & \cellcolor[HTML]{B4D2EA}20.8 & \cellcolor[HTML]{BDD8ED}21.0 & \cellcolor[HTML]{AACBE6}20.7 & \cellcolor[HTML]{88B6DB}22.7 & \cellcolor[HTML]{B0CFE9}22.7 & \cellcolor[HTML]{86B4DA}25.6 & \cellcolor[HTML]{347FBD}\underline{\textbf{31.2}} & \cellcolor[HTML]{B8D5EC}20.5 & \cellcolor[HTML]{B3D1EA}21.4 & \cellcolor[HTML]{4389C2}26.0 & \cellcolor[HTML]{9DC3E2}21.2 \\
\midrule
sw & \cellcolor[HTML]{DCECF8}20.4 & \cellcolor[HTML]{DCECF8}18.3 & \cellcolor[HTML]{DCECF8}19.7 & \cellcolor[HTML]{DCECF8}19.3 & \cellcolor[HTML]{DCECF8}19.6 & \cellcolor[HTML]{DCECF8}18.0 & \cellcolor[HTML]{DCECF8}18.6 & \cellcolor[HTML]{DCECF8}20.6 & \cellcolor[HTML]{DCECF8}20.3 & \cellcolor[HTML]{DCECF8}18.7 & \cellcolor[HTML]{347FBD}\underline{\textbf{32.2}} & \cellcolor[HTML]{DCECF8}20.0 & \cellcolor[HTML]{DCECF8}20.4 & \cellcolor[HTML]{DCECF8}18.5 \\
\midrule
multilingual & \cellcolor[HTML]{76AAD4}23.8 & \cellcolor[HTML]{83B3D9}23.1 & \cellcolor[HTML]{74A8D3}23.9 & \cellcolor[HTML]{73A8D3}23.2 & \cellcolor[HTML]{79ACD5}24.2 & \cellcolor[HTML]{67A0CF}24.2 & \cellcolor[HTML]{70A6D2}23.9 & \cellcolor[HTML]{8CB8DC}24.4 & \cellcolor[HTML]{A0C5E3}24.0 & \cellcolor[HTML]{A9CBE6}22.6 & \cellcolor[HTML]{B6D3EB}20.7 & \cellcolor[HTML]{619CCD}24.2 & \cellcolor[HTML]{347FBD}\underline{\textbf{26.6}} & \cellcolor[HTML]{5D9ACB}23.9 \\
\bottomrule
\end{tabular}%
}
\end{subtable}
\hfill
\begin{subtable}[t]{0.46\linewidth}
\centering
\caption{\textbf{Out-of-domain tasks:} Avg@8 improvement over the base model, in percentage points, on Multilingual Reasoning Gym tasks \textit{not seen during training}.}
\label{tab:pullin_crosslingual_matrix_reward_average_rg_heldout_rg_heldout}
\scriptsize
\setlength{\tabcolsep}{3pt}
\renewcommand{\arraystretch}{1.15}
\newcommand{\MatrixResizeWidth}{\linewidth}
\resizebox{\MatrixResizeWidth}{!}{%
\begin{tabular}{lrrrrrrrrrrr@{\hspace{1.2em}}rrr}
\toprule
Train $\downarrow$ / Eval $\rightarrow$ & en & zh & de & fr & es & ru & ja & th & bn & te & sw & it & ko & pt \\
\midrule
en & \cellcolor[HTML]{347FBD}\underline{\textbf{9.1}} & \cellcolor[HTML]{5997CA}8.5 & \cellcolor[HTML]{4B8EC5}8.7 & \cellcolor[HTML]{68A1CF}8.8 & \cellcolor[HTML]{468BC3}9.4 & \cellcolor[HTML]{9DC3E2}9.0 & \cellcolor[HTML]{5D9ACB}9.3 & \cellcolor[HTML]{5393C8}8.7 & \cellcolor[HTML]{76AAD4}8.8 & \cellcolor[HTML]{C7DEF0}7.0 & \cellcolor[HTML]{AFCFE8}5.9 & \cellcolor[HTML]{4F90C6}9.2 & \cellcolor[HTML]{5292C7}9.7 & \cellcolor[HTML]{498DC4}9.3 \\
\midrule
zh & \cellcolor[HTML]{85B4DA}7.6 & \cellcolor[HTML]{3E85C0}8.9 & \cellcolor[HTML]{68A1CF}8.3 & \cellcolor[HTML]{7AADD6}8.6 & \cellcolor[HTML]{6FA5D2}8.8 & \cellcolor[HTML]{76AAD4}9.9 & \cellcolor[HTML]{6FA5D2}9.1 & \cellcolor[HTML]{92BCDE}7.9 & \cellcolor[HTML]{9FC4E2}8.5 & \cellcolor[HTML]{DCECF8}6.8 & \cellcolor[HTML]{AACCE7}6.0 & \cellcolor[HTML]{639DCD}9.0 & \cellcolor[HTML]{6AA2D0}9.4 & \cellcolor[HTML]{5B98CB}9.0 \\
\midrule
de & \cellcolor[HTML]{659FCE}8.2 & \cellcolor[HTML]{4C8FC5}8.7 & \cellcolor[HTML]{5091C7}8.7 & \cellcolor[HTML]{4C8EC5}9.2 & \cellcolor[HTML]{4D8FC6}9.3 & \cellcolor[HTML]{98C0E0}9.1 & \cellcolor[HTML]{609BCC}9.3 & \cellcolor[HTML]{5594C9}8.7 & \cellcolor[HTML]{8FBADD}8.6 & \cellcolor[HTML]{C3DCEF}7.1 & \cellcolor[HTML]{90BBDD}6.3 & \cellcolor[HTML]{5192C7}9.2 & \cellcolor[HTML]{4D8FC6}9.8 & \cellcolor[HTML]{4B8EC5}9.2 \\
\midrule
fr & \cellcolor[HTML]{74A9D4}7.9 & \cellcolor[HTML]{76AAD4}8.0 & \cellcolor[HTML]{85B4D9}7.9 & \cellcolor[HTML]{77AAD4}8.7 & \cellcolor[HTML]{7BADD6}8.6 & \cellcolor[HTML]{B2D1E9}8.6 & \cellcolor[HTML]{9FC5E3}8.5 & \cellcolor[HTML]{70A6D2}8.3 & \cellcolor[HTML]{CEE3F3}8.3 & \cellcolor[HTML]{D6E8F6}6.9 & \cellcolor[HTML]{C8DFF1}5.6 & \cellcolor[HTML]{70A6D2}8.8 & \cellcolor[HTML]{70A6D2}9.3 & \cellcolor[HTML]{4F91C7}9.2 \\
\midrule
es & \cellcolor[HTML]{629DCD}8.2 & \cellcolor[HTML]{78ABD5}8.0 & \cellcolor[HTML]{5B98CB}8.5 & \cellcolor[HTML]{5695C9}9.0 & \cellcolor[HTML]{4B8EC5}9.3 & \cellcolor[HTML]{8DB9DC}9.4 & \cellcolor[HTML]{73A8D3}9.0 & \cellcolor[HTML]{5695C9}8.6 & \cellcolor[HTML]{A2C6E4}8.5 & \cellcolor[HTML]{C5DDF0}7.0 & \cellcolor[HTML]{DCECF8}5.4 & \cellcolor[HTML]{498DC4}9.3 & \cellcolor[HTML]{5595C9}9.7 & \cellcolor[HTML]{4E90C6}9.2 \\
\midrule
ru & \cellcolor[HTML]{4B8EC5}8.7 & \cellcolor[HTML]{498CC4}8.8 & \cellcolor[HTML]{4288C2}8.9 & \cellcolor[HTML]{347FBD}\underline{\textbf{9.4}} & \cellcolor[HTML]{347FBD}\underline{\textbf{9.7}} & \cellcolor[HTML]{347FBD}\underline{\textbf{11.3}} & \cellcolor[HTML]{347FBD}\underline{\textbf{9.8}} & \cellcolor[HTML]{67A0CF}8.4 & \cellcolor[HTML]{629DCD}8.9 & \cellcolor[HTML]{B4D2EA}7.2 & \cellcolor[HTML]{AECEE8}5.9 & \cellcolor[HTML]{3680BE}9.5 & \cellcolor[HTML]{3781BE}10.0 & \cellcolor[HTML]{347FBD}\underline{\textbf{9.6}} \\
\midrule
ja & \cellcolor[HTML]{90BBDD}7.4 & \cellcolor[HTML]{5896C9}8.5 & \cellcolor[HTML]{629DCD}8.4 & \cellcolor[HTML]{7AADD6}8.6 & \cellcolor[HTML]{7DAED7}8.6 & \cellcolor[HTML]{87B5DA}9.5 & \cellcolor[HTML]{5594C8}9.4 & \cellcolor[HTML]{7CAED6}8.2 & \cellcolor[HTML]{DCECF8}8.2 & \cellcolor[HTML]{92BCDE}7.6 & \cellcolor[HTML]{BFD9EE}5.7 & \cellcolor[HTML]{8CB8DC}8.4 & \cellcolor[HTML]{6EA5D2}9.3 & \cellcolor[HTML]{6BA3D0}8.7 \\
\midrule
th & \cellcolor[HTML]{4E90C6}8.6 & \cellcolor[HTML]{347FBD}\underline{\textbf{9.1}} & \cellcolor[HTML]{347FBD}\underline{\textbf{9.1}} & \cellcolor[HTML]{3882BE}9.4 & \cellcolor[HTML]{4187C2}9.5 & \cellcolor[HTML]{5A98CA}10.5 & \cellcolor[HTML]{3E85C0}9.7 & \cellcolor[HTML]{347FBD}\underline{\textbf{9.0}} & \cellcolor[HTML]{3A83BF}9.1 & \cellcolor[HTML]{95BEDF}7.6 & \cellcolor[HTML]{8BB8DC}6.4 & \cellcolor[HTML]{347FBD}\underline{\textbf{9.6}} & \cellcolor[HTML]{347FBD}\underline{\textbf{10.1}} & \cellcolor[HTML]{3B83BF}9.5 \\
\midrule
bn & \cellcolor[HTML]{70A6D2}8.0 & \cellcolor[HTML]{6BA3D0}8.2 & \cellcolor[HTML]{74A8D3}8.1 & \cellcolor[HTML]{82B1D8}8.5 & \cellcolor[HTML]{87B5DA}8.4 & \cellcolor[HTML]{8EB9DC}9.4 & \cellcolor[HTML]{5F9BCC}9.3 & \cellcolor[HTML]{4B8EC5}8.8 & \cellcolor[HTML]{347FBD}\underline{\textbf{9.2}} & \cellcolor[HTML]{7DAED7}7.8 & \cellcolor[HTML]{73A8D3}6.6 & \cellcolor[HTML]{8EB9DD}8.4 & \cellcolor[HTML]{5C99CB}9.6 & \cellcolor[HTML]{6DA4D1}8.7 \\
\midrule
te & \cellcolor[HTML]{C3DBEF}6.5 & \cellcolor[HTML]{D2E6F5}6.6 & \cellcolor[HTML]{A7C9E5}7.4 & \cellcolor[HTML]{CFE4F4}7.6 & \cellcolor[HTML]{B7D4EB}7.7 & \cellcolor[HTML]{B4D2EA}8.5 & \cellcolor[HTML]{D9EAF7}7.7 & \cellcolor[HTML]{A4C7E4}7.7 & \cellcolor[HTML]{3982BF}9.1 & \cellcolor[HTML]{347FBD}\underline{\textbf{8.6}} & \cellcolor[HTML]{83B2D9}6.5 & \cellcolor[HTML]{B0D0E9}8.0 & \cellcolor[HTML]{86B4DA}9.0 & \cellcolor[HTML]{A8CAE6}7.7 \\
\midrule
sw & \cellcolor[HTML]{DCECF8}6.0 & \cellcolor[HTML]{DCECF8}6.4 & \cellcolor[HTML]{DCECF8}6.6 & \cellcolor[HTML]{DCECF8}7.5 & \cellcolor[HTML]{DCECF8}7.1 & \cellcolor[HTML]{DCECF8}7.7 & \cellcolor[HTML]{DCECF8}7.7 & \cellcolor[HTML]{DCECF8}7.1 & \cellcolor[HTML]{C9DFF1}8.3 & \cellcolor[HTML]{BED9ED}7.1 & \cellcolor[HTML]{347FBD}\underline{\textbf{7.4}} & \cellcolor[HTML]{DCECF8}7.4 & \cellcolor[HTML]{DCECF8}7.9 & \cellcolor[HTML]{DCECF8}6.9 \\
\midrule
multilingual & \cellcolor[HTML]{609BCC}8.3 & \cellcolor[HTML]{75A9D4}8.1 & \cellcolor[HTML]{83B2D9}7.9 & \cellcolor[HTML]{81B1D8}8.5 & \cellcolor[HTML]{67A0CF}8.9 & \cellcolor[HTML]{79ACD5}9.8 & \cellcolor[HTML]{71A7D3}9.0 & \cellcolor[HTML]{73A8D3}8.3 & \cellcolor[HTML]{488CC4}9.1 & \cellcolor[HTML]{93BDDF}7.6 & \cellcolor[HTML]{97BFE0}6.2 & \cellcolor[HTML]{78ABD5}8.7 & \cellcolor[HTML]{5493C8}9.7 & \cellcolor[HTML]{609BCC}8.9 \\
\bottomrule
\end{tabular}%
}
\end{subtable}
\caption{Avg@8 improvement over base models, in percentage points, on seen and unseen Multilingual Reasoning Gym tasks. Results average all nine base models under the English reasoning reward. Cells are colored within each evaluation-language column from \textcolor[HTML]{97BFE0}{light blue} to \textcolor[HTML]{347FBD}{dark blue}. For multilingual training, Italian (it), Korean (ko), and Brazilian Portuguese (pt) are unseen. Full per-model results are available in \autoref{app:matrices}.
}
\label{tab:pullin_crosslingual_matrix_reward_average_rg_heldout}
\end{table*}

\begin{figure*}
    \centering
    \captionsetup[subfigure]{skip=0pt,font=figonesubcaption}
    \includegraphics[width=0.8\textwidth]{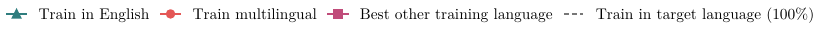}\\[-0.35em]
    \begin{minipage}{0.02\textwidth}
        \centering
        \raisebox{0pt}[9em][0pt]{\rotatebox{90}{\footnotesize Recovered target gain (\%)}}
    \end{minipage}\hfill
    \begin{minipage}{0.97\textwidth}
        \centering
        \begin{minipage}[t]{0.325\linewidth}
            \includegraphics[width=\linewidth]{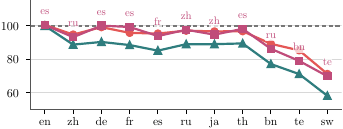}
            \subcaption{\texttt{Qwen3-1.7B-Base}}
        \end{minipage}\hfill
        \begin{minipage}[t]{0.325\linewidth}
            \includegraphics[width=\linewidth]{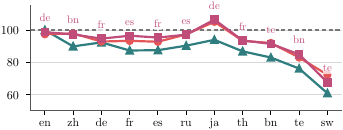}
            \subcaption{\texttt{Qwen3-4B-Base}}
        \end{minipage}\hfill
        \begin{minipage}[t]{0.325\linewidth}
            \includegraphics[width=\linewidth]{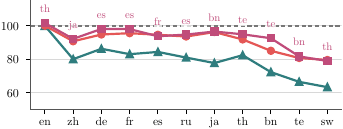}
            \subcaption{\texttt{Qwen3-8B-Base}}
        \end{minipage}\\[0.05em]
        \begin{minipage}[t]{0.325\linewidth}
            \includegraphics[width=\linewidth]{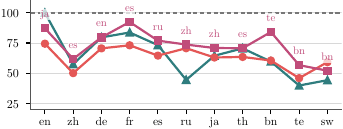}
            \subcaption{\texttt{Qwen3-1.7B}}
        \end{minipage}\hfill
        \begin{minipage}[t]{0.325\linewidth}
            \includegraphics[width=\linewidth]{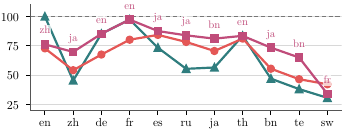}
            \subcaption{\texttt{Qwen3-4B}}
        \end{minipage}\hfill
        \begin{minipage}[t]{0.325\linewidth}
            \includegraphics[width=\linewidth]{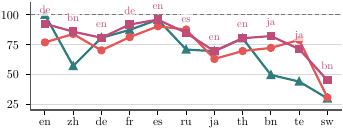}
            \subcaption{\texttt{Qwen3-8B}}
        \end{minipage}\\[0.05em]
        \begin{minipage}[t]{0.325\linewidth}
            \includegraphics[width=\linewidth]{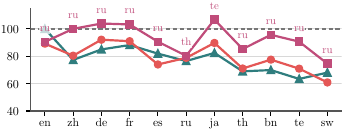}
            \subcaption{\texttt{gemma-3-1b-it}}
        \end{minipage}\hfill
        \begin{minipage}[t]{0.325\linewidth}
            \includegraphics[width=\linewidth]{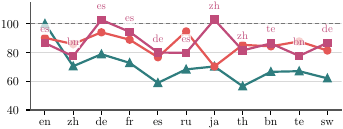}
            \subcaption{\texttt{gemma-3-4b-it}}
        \end{minipage}\hfill
        \begin{minipage}[t]{0.325\linewidth}
            \includegraphics[width=\linewidth]{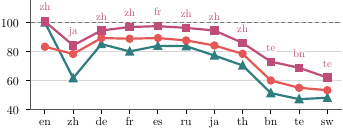}
            \subcaption{\texttt{SmolLM3-3B}}
        \end{minipage}
    \end{minipage}
    \caption{In-domain (seen tasks) Multilingual Reasoning Gym performance under English reasoning rewards. Each language label on the x-axis represents a target evaluation language. Values show different training strategies' performance as a percentage of direct target-language training. Labels on the best-other line show the best other source language for transfer (selected separately for each evaluation language as an oracle).}
    \label{fig:transfer-strategy-comparison}
    \vspace{-1em}
\end{figure*}

We next analyze crosslingual transfer by training monolingual GRPO checkpoints in eleven languages and then evaluating them across all languages. We also compare against multilingual training. We use English-reward variants, which generally performed better in \cref{sec:reasoning-lang-effect}.

\paragraph{Crosslingual transfer on seen tasks (in-domain).} \Cref{tab:pullin_crosslingual_matrix_reward_average_rg_heldout_train} reports the full training-language $\times$ evaluation-language matrix on in-domain Multilingual Reasoning Gym tasks, averaged over all models. We observe \textit{substantial crosslingual transfer}: much of the same-language improvement transfers to other languages. Still, training in an evaluation language generally performs best in that language. Trends hold for each individual model: our per-model breakdown in \cref{fig:transfer-strategy-comparison} compares English-only, multilingual, and an oracle best transfer-language against target-language training. Only a few cases slightly favor training in another language. Notably, English is often \textit{not} the best transfer language, and training in lower-resource languages can achieve strong crosslingual transfer: training in Bengali, for example, achieves the best transfer to Chinese for \texttt{Qwen3-4B-Base} and \texttt{Qwen3-8B}.

In particular, for French tasks, Spanish-only training nearly recovers the 25.6pp target-language training gain (-1.0pp). For lower-resource languages such as Swahili or Telugu, target-language data provides larger benefits and transfer from other languages is less effective. For example, Swahili training improves Swahili evaluation by 32.2pp on average, compared with 20.5pp from the next-best language, Telugu.
Training in Swahili in particular also provides substantially worse (but still positive) transfer to other languages, but this is not the case for languages such as Telugu or Bengali. In line with our results, \citet{huang2026beyond} find strong crosslingual transfer from English, Chinese, and German, but do not include low-resource training languages in their study.
Overall, crosslingual transfer is remarkably strong across languages, especially for high- or medium-resource languages.

\paragraph{Multilingual training.} The performance of multilingual training varies by model family (\cref{tab:pullin_crosslingual_matrix_reward_average_rg_heldout_train,fig:transfer-strategy-comparison}). For \texttt{Qwen3-Base}, it is competitive with selecting the best source language for transfer separately for each evaluation language, averaging 0.3pp lower across model--language cells and requires only one single model. For \texttt{Qwen3} (non-``Base''), it averages 1.3pp below. However, despite training on all languages in \cref{fig:transfer-strategy-comparison}, multilingual training still underperforms monolingual same-language training.

\begin{table}[t]
    \centering
    \small
    \setlength{\tabcolsep}{4pt}
    \resizebox{0.65\linewidth}{!}{%
    \begin{tabular}{@{}lcccc@{}}
        \toprule
        & \multicolumn{2}{c}{Easy Math} & \multicolumn{2}{c}{Hard Math} \\
        \cmidrule(lr){2-3}\cmidrule(lr){4-5}
        Model & Seen language &  Crosslingual Transfer & Seen language & Crosslingual Transfer \\
        \midrule
        \texttt{Qwen3-1.7B-Base} & 58.6 ({\setlength{\fboxsep}{1pt}\colorbox[HTML]{498E4C}{\strut +52.5}}) & 57.6 ({\setlength{\fboxsep}{1pt}\colorbox[HTML]{498E4C}{\strut +51.4}}) & 7.4 ({\setlength{\fboxsep}{1pt}\colorbox[HTML]{A3C9A4}{\strut +6.5}}) & 7.4 ({\setlength{\fboxsep}{1pt}\colorbox[HTML]{A2C8A3}{\strut +6.6}}) \\
        \texttt{Qwen3-4B-Base} & 77.9 ({\setlength{\fboxsep}{1pt}\colorbox[HTML]{2E7D32}{\strut +61.0}}) & 76.8 ({\setlength{\fboxsep}{1pt}\colorbox[HTML]{2E7D32}{\strut +59.9}}) & 17.5 ({\setlength{\fboxsep}{1pt}\colorbox[HTML]{509353}{\strut +14.1}}) & 17.4 ({\setlength{\fboxsep}{1pt}\colorbox[HTML]{509353}{\strut +14.0}}) \\
        \texttt{Qwen3-8B-Base} & 83.3 ({\setlength{\fboxsep}{1pt}\colorbox[HTML]{519454}{\strut +49.9}}) & 82.7 ({\setlength{\fboxsep}{1pt}\colorbox[HTML]{4F9353}{\strut +49.3}}) & 21.3 ({\setlength{\fboxsep}{1pt}\colorbox[HTML]{2E7D32}{\strut +17.2}}) & 21.2 ({\setlength{\fboxsep}{1pt}\colorbox[HTML]{2E7D32}{\strut +17.0}}) \\
        \cmidrule(lr){1-5}
        \texttt{Qwen3-1.7B} & 71.4 ({\setlength{\fboxsep}{1pt}\colorbox[HTML]{E0F0E0}{\strut +4.2}}) & 70.6 ({\setlength{\fboxsep}{1pt}\colorbox[HTML]{E1F1E1}{\strut +3.5}}) & 13.7 ({\setlength{\fboxsep}{1pt}\colorbox[HTML]{D45C5C}{\strut -12.0}}) & 12.4 ({\setlength{\fboxsep}{1pt}\colorbox[HTML]{D86B6B}{\strut -14.3}}) \\
        \texttt{Qwen3-4B} & 81.9 ({\setlength{\fboxsep}{1pt}\colorbox[HTML]{EAF7EA}{\strut +1.0}}) & 81.5 ({\setlength{\fboxsep}{1pt}\colorbox[HTML]{EAF7EA}{\strut +0.5}}) & 22.1 ({\setlength{\fboxsep}{1pt}\colorbox[HTML]{C62828}{\strut -15.7}}) & 18.1 ({\setlength{\fboxsep}{1pt}\colorbox[HTML]{C62828}{\strut -20.7}}) \\
        \texttt{Qwen3-8B} & 87.0 ({\setlength{\fboxsep}{1pt}\colorbox[HTML]{EAF7EA}{\strut +0.9}}) & 86.5 ({\setlength{\fboxsep}{1pt}\colorbox[HTML]{EAF7EA}{\strut +0.5}}) & 24.6 ({\setlength{\fboxsep}{1pt}\colorbox[HTML]{CC3E3E}{\strut -14.1}}) & 21.5 ({\setlength{\fboxsep}{1pt}\colorbox[HTML]{CD4444}{\strut -18.1}}) \\
        \cmidrule(lr){1-5}
        \texttt{gemma-3-1b-it} & 27.0 ({\setlength{\fboxsep}{1pt}\colorbox[HTML]{CBE3CB}{\strut +11.0}}) & 27.0 ({\setlength{\fboxsep}{1pt}\colorbox[HTML]{C9E2CA}{\strut +10.9}}) & 2.4 ({\setlength{\fboxsep}{1pt}\colorbox[HTML]{EAF7EA}{\strut +0.0}}) & 2.4 ({\setlength{\fboxsep}{1pt}\colorbox[HTML]{EAF7EA}{\strut +0.0}}) \\
        \texttt{gemma-3-4b-it} & 76.9 ({\setlength{\fboxsep}{1pt}\colorbox[HTML]{C9E1C9}{\strut +11.6}}) & 77.2 ({\setlength{\fboxsep}{1pt}\colorbox[HTML]{C6E0C7}{\strut +11.9}}) & 10.6 ({\setlength{\fboxsep}{1pt}\colorbox[HTML]{CBE3CC}{\strut +2.8}}) & 10.7 ({\setlength{\fboxsep}{1pt}\colorbox[HTML]{D0E6D0}{\strut +2.4}}) \\
        \cmidrule(lr){1-5}
        \texttt{SmolLM3-3B} & 66.5 ({\setlength{\fboxsep}{1pt}\colorbox[HTML]{AFD1B0}{\strut +19.8}}) & 65.8 ({\setlength{\fboxsep}{1pt}\colorbox[HTML]{AFD1B0}{\strut +19.1}}) & 24.1 ({\setlength{\fboxsep}{1pt}\colorbox[HTML]{E3F2E3}{\strut +0.7}}) & 23.9 ({\setlength{\fboxsep}{1pt}\colorbox[HTML]{FCF3F3}{\strut -1.2}}) \\
        \bottomrule
    \end{tabular}
    }
    \caption{Easy and Hard Math performance under English reasoning rewards. Cells report Avg@8 accuracy, with change from the original model in percentage points in parentheses, averaged over monolingual runs in the 11 training languages. ``Seen'' denotes evaluation in the training language. Crosslingual transfer denotes evaluation in any other language. Full per-model and per-language results are available in \cref{app:matrices}.}
    \label{tab:easy-hard-math-performance}
\end{table}

\paragraph{Crosslingual transfer on unseen tasks (out-of-domain).}
\Cref{tab:pullin_crosslingual_matrix_reward_average_rg_heldout_rg_heldout} reports the full training-language $\times$ evaluation-language matrix on unseen Multilingual Reasoning Gym tasks, aggregated over all models.
Crosslingual transfer remains strong on unseen tasks, but other trends are less clear. For example, same-language training is no longer consistently best. Transfer from other languages can outperform same-language training in some cases, without ever being finetuned on samples from the evaluation language!
\Cref{tab:easy-hard-math-performance} further reports performance on MGSM and PolyMath, both unseen during training, aggregated over all training languages. As described in \cref{sec:eval-setup}, we group MGSM and PolyMath-low as ``Easy Math'' and the remaining splits as ``Hard Math''. Training on Multilingual Reasoning Gym tasks in one language improves unseen math tasks in that language, and the improvements also transfer to others. However, for specific models, particularly on Hard Math tasks, we observe regressions in performance compared to the base model, while performance on other tasks remains stable or improves. In \cref{sec:crosslingual-transfer-forgetting}, we observe that training in specific languages can collapse capabilities in others on specific tasks, while other training languages, particularly English, may largely mitigate the effect or improve performance.

\begin{table*}
    \vspace{-3mm}
    \centering
    \begin{minipage}[t]{0.32\textwidth}
        \centering
        \subcaption{\texttt{Qwen3-1.7B}}
        \vspace{0.10em}
        \centering
\begingroup
\tiny
\setlength{\tabcolsep}{1.7pt}
\renewcommand{\arraystretch}{0.9}
\resizebox{\linewidth}{!}{%
\begin{tabular}{lrrrrrrrrr}
\toprule
Train $\downarrow$ / Eval $\rightarrow$  & en & zh & de & fr & ru & ja & bn & te & sw \\
\midrule
en & \cellcolor[HTML]{79AE7B}5.1 & \cellcolor[HTML]{5B9A5E}11.7 & \cellcolor[HTML]{4D9150}8.5 & \cellcolor[HTML]{4A8F4E}10.3 & \cellcolor[HTML]{94BF96}9.2 & \cellcolor[HTML]{529455}6.1 & \cellcolor[HTML]{2E7D32}10.7 & \cellcolor[HTML]{80B383}6.6 & \cellcolor[HTML]{3A853D}3.9 \\
\midrule
zh & \cellcolor[HTML]{5D9B60}6.4 & \cellcolor[HTML]{2E7D32}15.3 & \cellcolor[HTML]{2E7D32}10.2 & \cellcolor[HTML]{2E7D32}12.2 & \cellcolor[HTML]{2E7D32}16.4 & \cellcolor[HTML]{4D9150}6.3 & \cellcolor[HTML]{37833B}10.4 & \cellcolor[HTML]{358138}9.2 & \cellcolor[HTML]{74AB77}3.7 \\
\midrule
fr & \cellcolor[HTML]{2E7D32}8.5 & \cellcolor[HTML]{38833B}14.6 & \cellcolor[HTML]{5C9B5F}7.7 & \cellcolor[HTML]{448B48}10.7 & \cellcolor[HTML]{B5D4B6}6.9 & \cellcolor[HTML]{5B9A5E}5.7 & \cellcolor[HTML]{80B282}7.7 & \cellcolor[HTML]{529556}8.2 & \cellcolor[HTML]{D2E8D3}3.2 \\
\midrule
bn & \cellcolor[HTML]{C4DEC5}1.7 & \cellcolor[HTML]{A9CDAB}5.3 & \cellcolor[HTML]{A6CBA8}3.7 & \cellcolor[HTML]{B2D3B3}3.6 & \cellcolor[HTML]{EAF7EA}3.1 & \cellcolor[HTML]{2E7D32}7.5 & \cellcolor[HTML]{7AAE7C}7.9 & \cellcolor[HTML]{2E7D32}9.4 & \cellcolor[HTML]{5D9B60}3.8 \\
\midrule
te & \cellcolor[HTML]{C62828}-8.1 & \cellcolor[HTML]{DB7777}-2.4 & \cellcolor[HTML]{CCE4CD}1.6 & \cellcolor[HTML]{C62828}-1.6 & \cellcolor[HTML]{D0E6D1}4.9 & \cellcolor[HTML]{C62828}-13.0 & \cellcolor[HTML]{7DB07F}7.8 & \cellcolor[HTML]{458C49}8.6 & \cellcolor[HTML]{69A36B}3.7 \\
\midrule
sw & \cellcolor[HTML]{CF4848}-6.9 & \cellcolor[HTML]{C62828}-3.8 & \cellcolor[HTML]{C62828}-1.2 & \cellcolor[HTML]{ABCEAD}4.1 & \cellcolor[HTML]{CEE5CF}5.1 & \cellcolor[HTML]{E39494}-6.4 & \cellcolor[HTML]{EAF7EA}3.7 & \cellcolor[HTML]{EAF7EA}3.1 & \cellcolor[HTML]{2E7D32}4.0 \\
\midrule
multi & \cellcolor[HTML]{8AB98C}4.3 & \cellcolor[HTML]{569759}12.1 & \cellcolor[HTML]{6DA66F}6.8 & \cellcolor[HTML]{5B9A5E}9.3 & \cellcolor[HTML]{609E63}12.8 & \cellcolor[HTML]{7CB07E}4.4 & \cellcolor[HTML]{83B485}7.6 & \cellcolor[HTML]{96C098}5.9 & \cellcolor[HTML]{EAF7EA}3.1 \\
\bottomrule
\end{tabular}%
}
\endgroup

    \end{minipage}\hfill
    \begin{minipage}[t]{0.34\textwidth}
        \centering
        \subcaption{\texttt{Qwen3-4B}}
        \vspace{0.10em}
        
\centering
\begingroup
\tiny
\setlength{\tabcolsep}{1.7pt}
\renewcommand{\arraystretch}{0.9}
\resizebox{\linewidth}{!}{%
\begin{tabular}{lrrrrrrrrr}
\toprule
Train $\downarrow$ / Eval $\rightarrow$  & en & zh & de & fr & ru & ja & bn & te & sw \\
\midrule
en & \cellcolor[HTML]{8DBB8F}2.2 & \cellcolor[HTML]{8DBB8F}3.1 & \cellcolor[HTML]{4D9150}3.7 & \cellcolor[HTML]{6FA771}3.6 & \cellcolor[HTML]{BFDBC0}3.1 & \cellcolor[HTML]{3A853E}4.8 & \cellcolor[HTML]{95C096}5.5 & \cellcolor[HTML]{4D9150}7.8 & \cellcolor[HTML]{A7CCA8}6.8 \\
\midrule
zh & \cellcolor[HTML]{86B688}2.4 & \cellcolor[HTML]{468C49}5.4 & \cellcolor[HTML]{569759}3.5 & \cellcolor[HTML]{58985B}4.3 & \cellcolor[HTML]{86B688}7.1 & \cellcolor[HTML]{4D9150}4.3 & \cellcolor[HTML]{84B587}6.4 & \cellcolor[HTML]{4B904E}7.8 & \cellcolor[HTML]{2E7D32}14.6 \\
\midrule
fr & \cellcolor[HTML]{4C9150}3.8 & \cellcolor[HTML]{448B48}5.4 & \cellcolor[HTML]{2E7D32}4.4 & \cellcolor[HTML]{2E7D32}5.5 & \cellcolor[HTML]{91BD93}6.3 & \cellcolor[HTML]{65A168}3.6 & \cellcolor[HTML]{65A168}8.1 & \cellcolor[HTML]{2E7D32}8.6 & \cellcolor[HTML]{9AC39C}7.7 \\
\midrule
bn & \cellcolor[HTML]{EFC3C3}-5.4 & \cellcolor[HTML]{2E7D32}6.2 & \cellcolor[HTML]{9AC39C}1.9 & \cellcolor[HTML]{67A26A}3.8 & \cellcolor[HTML]{629F65}9.7 & \cellcolor[HTML]{3C8640}4.7 & \cellcolor[HTML]{2E7D32}11.1 & \cellcolor[HTML]{478D4A}7.9 & \cellcolor[HTML]{A4CAA6}7.0 \\
\midrule
te & \cellcolor[HTML]{D76767}-13.6 & \cellcolor[HTML]{D56060}-10.9 & \cellcolor[HTML]{E59E9E}-5.3 & \cellcolor[HTML]{C72B2B}-8.8 & \cellcolor[HTML]{F4D5D5}-1.0 & \cellcolor[HTML]{FDF7F7}-0.5 & \cellcolor[HTML]{94BF95}5.6 & \cellcolor[HTML]{559658}7.6 & \cellcolor[HTML]{C3DEC4}5.1 \\
\midrule
sw & \cellcolor[HTML]{C62828}-19.2 & \cellcolor[HTML]{C62828}-14.8 & \cellcolor[HTML]{C62828}-11.8 & \cellcolor[HTML]{C62828}-8.9 & \cellcolor[HTML]{C62828}-5.2 & \cellcolor[HTML]{C62828}-14.0 & \cellcolor[HTML]{EAF7EA}0.8 & \cellcolor[HTML]{EAF7EA}3.6 & \cellcolor[HTML]{AACEAC}6.6 \\
\midrule
multi & \cellcolor[HTML]{2E7D32}4.5 & \cellcolor[HTML]{307E34}6.1 & \cellcolor[HTML]{75AB77}2.8 & \cellcolor[HTML]{6FA771}3.6 & \cellcolor[HTML]{2E7D32}13.3 & \cellcolor[HTML]{2E7D32}5.1 & \cellcolor[HTML]{7BAF7E}6.9 & \cellcolor[HTML]{418944}8.1 & \cellcolor[HTML]{EAF7EA}2.6 \\
\bottomrule
\end{tabular}%
}
\endgroup

    \end{minipage}\hfill
    \begin{minipage}[t]{0.30\textwidth}
        \centering
        \subcaption{\texttt{Qwen3-8B}}
        \vspace{0.10em}
        \centering
\begingroup
\tiny
\setlength{\tabcolsep}{1.7pt}
\renewcommand{\arraystretch}{0.9}
\resizebox{\linewidth}{!}{%
\begin{tabular}{lrrrrrrrrr}
\toprule
Train $\downarrow$ / Eval $\rightarrow$  & en & zh & de & fr & ru & ja & bn & te & sw \\
\midrule
en & \cellcolor[HTML]{57985A}3.0 & \cellcolor[HTML]{5B9A5E}5.2 & \cellcolor[HTML]{7DB07F}2.6 & \cellcolor[HTML]{468D49}4.9 & \cellcolor[HTML]{2E7D32}10.2 & \cellcolor[HTML]{2E7D32}5.6 & \cellcolor[HTML]{2E7D32}5.4 & \cellcolor[HTML]{4A8F4D}5.4 & \cellcolor[HTML]{A6CBA7}5.9 \\
\midrule
zh & \cellcolor[HTML]{FAEBEB}-0.5 & \cellcolor[HTML]{2E7D32}6.8 & \cellcolor[HTML]{2E7D32}4.5 & \cellcolor[HTML]{5A9A5D}4.6 & \cellcolor[HTML]{418A45}9.2 & \cellcolor[HTML]{5A995D}4.6 & \cellcolor[HTML]{6AA46C}4.3 & \cellcolor[HTML]{AACDAB}4.2 & \cellcolor[HTML]{AACEAC}5.8 \\
\midrule
fr & \cellcolor[HTML]{4F9252}3.2 & \cellcolor[HTML]{75AB77}4.2 & \cellcolor[HTML]{307F34}4.4 & \cellcolor[HTML]{3C863F}5.1 & \cellcolor[HTML]{38843C}9.7 & \cellcolor[HTML]{5C9B5F}4.5 & \cellcolor[HTML]{3D8741}5.1 & \cellcolor[HTML]{2E7D32}5.7 & \cellcolor[HTML]{AACEAC}5.8 \\
\midrule
bn & \cellcolor[HTML]{C9E2CA}0.7 & \cellcolor[HTML]{58985B}5.3 & \cellcolor[HTML]{89B88B}2.3 & \cellcolor[HTML]{2E7D32}5.3 & \cellcolor[HTML]{4B904E}8.7 & \cellcolor[HTML]{509354}4.8 & \cellcolor[HTML]{468C49}4.9 & \cellcolor[HTML]{2E7D32}5.7 & \cellcolor[HTML]{C5DFC6}4.8 \\
\midrule
te & \cellcolor[HTML]{E7A5A5}-2.3 & \cellcolor[HTML]{CAE2CA}1.2 & \cellcolor[HTML]{A4CAA6}1.7 & \cellcolor[HTML]{CFE5CF}2.7 & \cellcolor[HTML]{BAD8BB}2.6 & \cellcolor[HTML]{EAF7EA}1.1 & \cellcolor[HTML]{6DA66F}4.2 & \cellcolor[HTML]{4A8F4D}5.4 & \cellcolor[HTML]{9FC6A0}6.2 \\
\midrule
sw & \cellcolor[HTML]{C62828}-5.4 & \cellcolor[HTML]{C62828}-2.7 & \cellcolor[HTML]{EAF7EA}0.0 & \cellcolor[HTML]{EAF7EA}2.2 & \cellcolor[HTML]{C62828}-6.7 & \cellcolor[HTML]{9CC49E}2.9 & \cellcolor[HTML]{EAF7EA}1.9 & \cellcolor[HTML]{EAF7EA}3.4 & \cellcolor[HTML]{2E7D32}10.4 \\
\midrule
multi & \cellcolor[HTML]{2E7D32}3.8 & \cellcolor[HTML]{78AD7A}4.1 & \cellcolor[HTML]{96C198}2.0 & \cellcolor[HTML]{76AC78}4.1 & \cellcolor[HTML]{338037}9.9 & \cellcolor[HTML]{6AA46C}4.2 & \cellcolor[HTML]{C0DCC1}2.7 & \cellcolor[HTML]{AACDAB}4.2 & \cellcolor[HTML]{EAF7EA}3.4 \\
\bottomrule
\end{tabular}%
}
\endgroup

    \end{minipage}
    \vspace{-0.5em}
    \caption{Language-dependent regressions on English-content Multilingual Reasoning Gym tasks (unseen during training) under English reasoning rewards. Panels excerpt the corresponding crosslingual transfer matrices: rows are training languages, columns are evaluation languages, and cells report avg@8 change from the original model in percentage points. Full results are available in the Appendix: \cref{tab:appendix_crosslingual_qwen3_1_7b_5_english_content_add,tab:appendix_crosslingual_qwen3_4b_5_english_content_add,tab:appendix_crosslingual_qwen3_8b_5_english_content_add}.}
    \label{fig:language-dependent-forgetting-panels}
    \vspace{-1em}
\end{table*}

\begin{figure*}
    \centering
    \captionsetup[subfigure]{font=figmodelcaption}
    \begin{minipage}[t]{0.49\textwidth}
        \centering
        \subcaption{\texttt{Qwen3-1.7B}}
        \vspace{-0.22em}
        \includegraphics[width=\linewidth, clip]{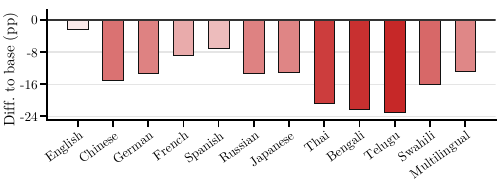}
    \end{minipage}\hfill
    \begin{minipage}[t]{0.49\textwidth}
        \centering
        \subcaption{\texttt{Qwen3-4B}}
        \vspace{-0.22em}
        \includegraphics[width=\linewidth, clip]{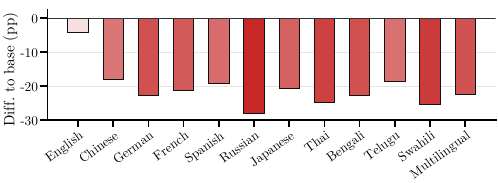}
    \end{minipage}

    \begin{minipage}[t]{0.49\textwidth}
        \centering
        \subcaption{\texttt{Qwen3-8B}}
        \vspace{-0.22em}
        \includegraphics[width=\linewidth, clip]{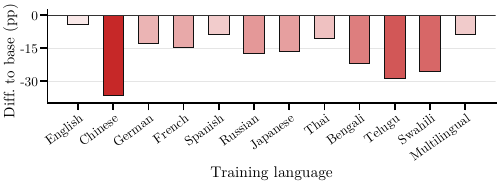}
    \end{minipage}\hfill
    \begin{minipage}[t]{0.49\textwidth}
        \centering
        \subcaption{\texttt{SmolLM3-3B}}
        \vspace{-0.22em}
        \includegraphics[width=\linewidth, clip]{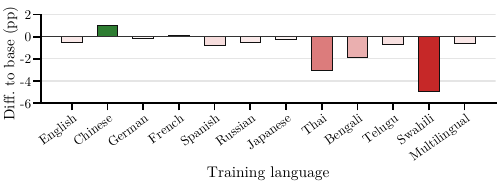}
    \end{minipage}
    \vspace{-0.2em}
    \caption{Hard Math language-dependent regressions under English reasoning rewards. Bars report avg@8 change from the original model in percentage points, averaged over all evaluation languages. Full results are available in the Appendix: \cref{tab:appendix_crosslingual_qwen3_1_7b_8_appendix_hard_math_macro,tab:appendix_crosslingual_qwen3_4b_8_appendix_hard_math_macro,tab:appendix_crosslingual_qwen3_8b_8_appendix_hard_math_macro,tab:appendix_crosslingual_smollm3_3b_8_appendix_hard_math_macro}.}
    \label{fig:hard-math-language-dependent-forgetting}
\end{figure*}

\subsection{Language-Dependent Task Regressions}
\label{sec:crosslingual-transfer-forgetting}

\paragraph{Training language.} We observe that specific combinations of base models and training languages exhibit extreme regression on specific types of tasks, such as difficult math problems or unseen tasks with English content from the Multilingual Reasoning Gym.
Recent work has analyzed reinforcement learning and GRPO as inducing less catastrophic forgetting than SFT \citep{chen2025retainingdoingroleonpolicy,shenfeld2025rlsrazoronlinereinforcement,lai2026reinforcementfinetuningnaturallymitigates}. We observe a different phenomenon of \textit{language-dependent task regressions} under GRPO.
For the \texttt{Qwen3} (non-``Base'') model family, we observe extreme regression on English content tasks when training on Swahili or Telugu (see \cref{fig:language-dependent-forgetting-panels}).
However, performance is preserved in the respective training language. We do not observe this effect for multilingual training or monolingual training in any other language.

A second, broader pattern appears on Hard Math: across many models, we observe extreme regression on difficult math tasks across many evaluated languages if English prompts are not included in the training data. We show examples of this in \cref{fig:hard-math-language-dependent-forgetting}. Monolingual training on English prompts mostly alleviates this regression incurring only small performance drops on average, while multilingual training (which includes a smaller fraction of English prompts) is less effective and can still result in substantial regressions depending on the model.
Note that we observe such extreme regression on difficult math tasks for training languages for which the same trained checkpoints \textit{do not show} such regressions  on other tasks (\textit{cf.} \cref{tab:easy-hard-math-performance,fig:language-dependent-forgetting-panels}).

This suggests that these task regressions are highly model- and language-specific. In our setting, these regressions are limited to tasks which are unseen during training.
Our observation is related to prior work on multilingual catastrophic forgetting and crosslingual interference in supervised finetuning. Prior studies show that finetuning multilingual models on one language or task can degrade generation or classification performance in other languages, especially in zero-shot crosslingual settings \citep{vu2022overcoming,liu2025conditionscatastrophicforgettingmultilingual}. However, our results show a more specific failure mode: for the same model, some training languages may induce regressions while others do not. Furthermore, we observe large regressions on particular unseen languages or task families while preserving, or even improving, performance in other evaluation languages or tasks. 
Following recent large-scale reasoning-RL recipes, we do not add an explicit KL penalty to the GRPO objective \citep{yu2025dapoopensourcellmreinforcement,minimax2025minimaxm1scalingtesttimecompute, hu2025openreasonerzeroopensourceapproach,he2025skyworkopenreasoner1,hou2026singlerolloutasynchronousoptimizationagentic}. Although KL regularization could mitigate the language-dependent regressions we observed during training, it would also constrain the policy's departure from the base model, potentially limiting the attainable performance gains.

\begin{figure*}
    \centering
    \captionsetup[subfigure]{font=figmodelcaption}
    \vspace{-0.6em}
    \begin{minipage}[t]{0.36\textwidth}
        \centering
        \includegraphics[width=\linewidth, trim={0pt 0pt 0pt 2pt}, clip]{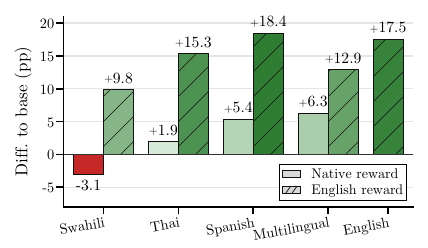}
        \vspace{-1.8em}
        \subcaption{\texttt{gemma-3-1b-it} on Easy Math.}
        \label{fig:reward-specific-forgetting-panels:gemma-1b-easy}
    \end{minipage}\hfill
    \begin{minipage}[t]{0.34\textwidth}
        \centering
        \includegraphics[width=\linewidth, trim={0pt 0pt 0pt 2pt}, clip]{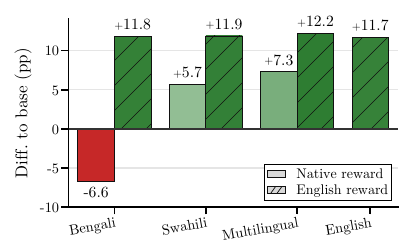}
        \vspace{-1.8em}

        \subcaption{\texttt{gemma-3-4b-it} on Easy Math.}
    \end{minipage}\hfill
    \begin{minipage}[t]{0.28\textwidth}
        \centering
        \includegraphics[width=\linewidth, trim={2pt 0pt 2pt 2pt}, clip]{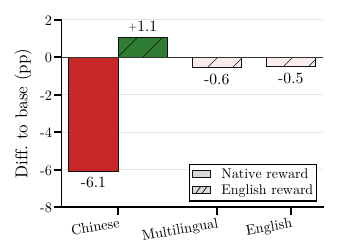}
        \vspace{-1.8em}
        \subcaption{\texttt{SmolLM3-3B} on Hard Math.}
    \end{minipage}
     \vspace{-0.5em}
    \caption{Reward-specific regression examples. Bars report avg@8 change from the original model in percentage points, averaged over all evaluation languages. Full results are available in the Appendix: \cref{tab:appendix_crosslingual_gemma_3_1b_it_7_appendix_easy_math_macro,tab:appendix_crosslingual_gemma_3_4b_it_7_appendix_easy_math_macro,tab:appendix_crosslingual_smollm3_3b_8_appendix_hard_math_macro}.}
    \label{fig:reward-specific-forgetting-panels}
    \vspace{-1em}

\end{figure*}

\paragraph{Reasoning language.}
\phantomsection
\label{sec:reasoning-lang-effect-forgetting}
Furthermore, we also observe language-dependent regressions based on the reasoning language reward used during training in combination with certain training languages, rather than just the training language itself.
We specifically analyze cases with a large delta between English and native reasoning reward performance. In \cref{fig:reward-specific-forgetting-panels}, we show that in some cases, native-language reasoning can lead to substantial degradation of previously learned capabilities, while training with English reasoning rewards delivers improvements or at least largely mitigates the issue. For example, native-reasoning reward training \texttt{gemma-3-1b-it} on Swahili prompts and \texttt{gemma-3-4b-it} on Bengali prompts shows a large drop in performance on Easy Math tasks (MGSM and PolyMath-low) compared to the original model, while English reasoning rewards actually improve performance substantially. For \texttt{SmolLM3-3B} on Chinese, we observe a similar effect on Hard Math tasks (PolyMath-medium, high, and top), although English reasoning does not improve substantially and only mitigates the forgetting.
Crucially, these effects are model- and language-specific: for example, \texttt{gemma-3-1b-it} on Bengali does not show such regressions, and \texttt{gemma-3-4b-it} on Swahili does not show such regressions either. This underscores the importance of our broad coverage of models, training and evaluation languages.

\section{Conclusion}
We studied multilingual GRPO training across nine base models, 11 training languages, multilingual training, and English- versus native-reasoning rewards. Training with native-reasoning rewards yields models with often only a small performance drop compared to training with English reasoning rewards, although reasoning in English yields larger benefits for lower-resource target languages.
GRPO also transfers strongly across languages. Training in one language often substantially improves performance in others. However, direct target-language training remains important for in-domain performance, especially for lower-resource languages.
At the same time, transfer gains can hide severe regressions. Some model-language combinations degrade on unseen English-content tasks or difficult math tasks, even when performance in the training language is preserved. Multilingual and non-English RLVR should therefore be evaluated not only by average gains, but also by language- and task-specific regressions on a broad panel of evaluation languages and tasks.
\FloatBarrier

\section*{Limitations}
Our study uses the Multilingual Reasoning Gym \citep{dobler2026multilingualreasoninggym} for training, which provides procedurally generated, template-based RLVR tasks with a dynamic difficulty curriculum. This enables controlled multilingual training at scale, but the data does not have the same variability as ``in-the-wild'' user queries or other real-world scenarios due to the template-based nature of the data. 
We also use a shared training recipe across many model, language, and reward configurations rather than tuning each setting individually. Separately tuning all model-language combinations was unfortunately infeasible for this study. However, some negative results, including cases of language-dependent task regressions, may therefore be reduced by targeted hyperparameter tuning.
Similarly, we run all experiments with a single seed only due to the breadth of our study.
Finally, our experiments focus on RLVR with automatically verifiable rewards in math, logic and reasoning puzzle tasks. We do not study reward models, tool-use or more ambiguous and open-ended tasks that are not automatically verifiable.

\section*{Acknowledgments}
Konstantin Dobler acknowledges the support of the European Laboratory for Learning and Intelligent Systems (ELLIS) PhD program and Federal Ministry of Research, Technology and Space under the funding code “KI-Servicezentrum Berlin-Brandenburg” 16IS22092.
We further thank Adam Golinski and Federico Danieli for their valuable feedback.
\bibliographystyle{plainnat}
\bibliography{custom}

\appendix
\toggletrue{inappendix}
\crefalias{section}{appendix}
\crefalias{subsection}{appendix}
\crefalias{subsubsection}{appendix}
\crefalias{paragraph}{appendix}
\crefalias{subparagraph}{appendix}
\clearpage

\section{Training Details}

\paragraph{Technical Details.}
\phantomsection
\label{app:technical-details}
We use the DAPO-style loss formulation \citep{yu2025dapoopensourcellmreinforcement} and do not scale the rewards by the group standard deviation following \citet{liu2025understandingr1zeroliketrainingcritical}.
We run RL training for 500 steps in each language. Each step has a total batch size of 512. As we use 8 rollouts per prompt, this results in 64 unique prompts seen per step. 
We use a learning rate of 1e-6, maximum gradient norm of 0.1, and the AdamW optimizer \citep{kingma2017adammethodstochasticoptimization, loshchilov2019decoupledweightdecayregularization} with $\beta_1=0.9$, $\beta_2=0.999$ and $\epsilon=1e-8$. 
We use a temperature of 1.0 for sampling rollouts and a maximum rollout length of 16384 tokens.
We mask completions exceeding the maximum rollout length.
We do not apply a Kullback-Leibler divergence (KL) penalty.
For multilingual training, after sampling a task, we sample a language uniformly at random from the set of training languages. 
We run training on Nvidia A100 80GB GPU nodes with HSDP model sharding using \texttt{accelerate} \citep{accelerate} and data parallel rollout generation via \texttt{vllm} \citep{kwon2023efficient}.

For evaluation, we generate 8 rollouts per sample and report the avg@8 score. We use the recommended sampling parameters by the publisher for each model family and a maximum rollout length of 16384 tokens. For the Multilingual Reasoning Gym, we use the default difficulty level for each task. We generate a test set of 25 samples per task using a different random seed than we use during training to test in-domain performance on unseen samples. 
For MGSM and PolyMath, we use the official evaluation sets provided by the respective authors and we use \texttt{Math-Verify} for comparing answers to the reference solution \citep{Kydlicek_Math-Verify_Math_Verification}.
We run evaluations on a single Nvidia A100 80GB GPU node with data parallel rollout generation via \texttt{vllm} \citep{kwon2023efficient}.

For language detection, we use the \texttt{langdetect} Python package \citep{nakatani2010langdetect} following the approach by \citet{wang2025polymath}. 

\paragraph{Multilingual Reasoning Gym setup.}
\phantomsection
\label{app:tasklist}
We first separate tasks which (1) contain English content and (2) are too difficult for \texttt{Qwen3-14B-Base} even at the (easy) default task difficulty curriculum level. This amounts to 22 filtered tasks out of 92 total tasks, which we use as additional unseen evaluation tasks. From the remaining 70 tasks, we sample 8 heldout tasks and use the following 62 tasks for training:
\begin{itemize}
  \item \textbf{Algebra}: \texttt{complex\_arithmetic}, \texttt{intermediate\_integration}, \texttt{polynomial\_equations}, \texttt{polynomial\_multiplication}, \texttt{simple\_equations}
  \item \textbf{Algorithmic}: \texttt{ab}, \texttt{base\_conversion}, \texttt{binary\_alternation}, \texttt{binary\_matrix}, \texttt{count\_primes}, \texttt{game\_of\_life\_halting}, \texttt{graph\_color}, \texttt{isomorphic\_strings}, \texttt{jugs}, \texttt{manipulate\_matrix}, \texttt{number\_filtering}, \texttt{number\_sorting}, \texttt{palindrome\_generation}, \texttt{pool\_matrix}, \texttt{ransom\_note}, \texttt{rotate\_matrix}, \texttt{spiral\_matrix}, \texttt{string\_insertion}, \texttt{string\_manipulation}, \texttt{string\_splitting}, \texttt{string\_synthesis}
  \item \textbf{ARC}: \texttt{arc\_agi}, \texttt{rearc}
  \item \textbf{Arithmetic}: \texttt{basic\_arithmetic}, \texttt{bitwise\_arithmetic}, \texttt{chain\_sum}, \texttt{count\_bits}, \texttt{decimal\_arithmetic}, \texttt{decimal\_chain\_sum}, \texttt{fraction\_simplification}, \texttt{gcd}, \texttt{lcm}, \texttt{leg\_counting}, \texttt{prime\_factorization}, \texttt{products}
  \item \textbf{Cognition}: \texttt{color\_cube\_rotation}, \texttt{modulo\_grid}, \texttt{number\_sequences}
  \item \textbf{Games}: \texttt{boxnet}, \texttt{countdown}, \texttt{tower\_of\_hanoi}, \texttt{knight\_swap}, \texttt{mahjong\_puzzle}, \texttt{maze}, \texttt{mini\_sudoku}, \texttt{puzzle24}, \texttt{survo}
  \item \textbf{Geometry}: \texttt{advanced\_geometry}, \texttt{simple\_geometry}
  \item \textbf{Graphs}: \texttt{course\_schedule}, \texttt{largest\_island}, \texttt{quantum\_lock}
  \item \textbf{Induction}: \texttt{list\_functions}
  \item \textbf{Logic}: \texttt{circuit\_logic}, \texttt{propositional\_logic}, \texttt{self\_reference}, \texttt{syllogisms}
\end{itemize}
We use the following 8 tasks sampled as heldout tasks (these tasks did pass the difficulty and English content filter):
\begin{itemize}
  \item \textbf{Algebra}: \texttt{simpleintegration}
  \item \textbf{Algorithmic}: \texttt{rottenoranges}
  \item \textbf{ARC}: \texttt{arc1d}
  \item \textbf{Arithmetic}: \texttt{powerfunction}
  \item \textbf{Cognition}: \texttt{rectanglecount}
  \item \textbf{Games}: \texttt{tsumego}
  \item \textbf{Graphs}: \texttt{shortestpath}
  \item \textbf{Logic}: \texttt{aliceinwonderland}
\end{itemize}
We also evaluate on 9 English content tasks:
\begin{itemize}
  \item \textbf{Algorithmic}: \texttt{caesar\_cipher}, \texttt{group\_anagrams}, \texttt{letter\_counting}, \texttt{letter\_jumble}, \texttt{sentence\_reordering}, \texttt{spell\_backward}, \texttt{word\_sequence\_reversal}, \texttt{word\_sorting}
  \item \textbf{Arithmetic}: \texttt{number\_format}
\end{itemize}

We furthermore evaluate on 13 ``too difficult'' tasks:
\begin{itemize}
  \item \textbf{Algorithmic}: \texttt{cryptarithm}, \texttt{game\_of\_life}, \texttt{palindrome\_partitioning}
  \item \textbf{Arithmetic}: \texttt{dice}
  \item \textbf{Code}: \texttt{bf}
  \item \textbf{Cognition}: \texttt{rubiks\_cube}
  \item \textbf{Games}: \texttt{emoji\_mystery}, \texttt{futoshiki}, \texttt{kakurasu}, \texttt{n\_queens}, \texttt{rush\_hour}, \texttt{sokoban}, \texttt{sudoku}
\end{itemize}

\section{Sensitivity to language-control filtering}
\label{app:reasoning-reward-unfiltered}
The English-versus-native reward comparison in \cref{fig:reason_reward_model_summary,fig:reason_reward_seen_language_summary} is restricted to pairs in which both checkpoints successfully control the reasoning language. As a sensitivity analysis, we repeat the same aggregation over all 50 trained, paired non-English monolingual configurations for the five models shown in those figures, irrespective of post-training language control. The English-reward advantage remains positive, decreasing only from 2.8 to 2.4pp on seen languages and from 4.0 to 3.4pp on unseen languages. The main analysis excludes eight configurations in which one checkpoint does not control its intended reasoning language. Including these configurations does not provide a clean comparison of reasoning languages because the ineffective variant instead reasons in the same language as its paired counterpart.

\section{Additional Related Work}
\label{app:related-work}
\paragraph{Multilingual reasoning benchmarks.}
Multilingual reasoning has been studied through translated benchmarks and multilingual instruction tuning. MGSM shows that chain-of-thought prompting can transfer across languages, while also revealing substantial multilingual gaps even on relatively simple mathematical problems \citep{shi2022mgsm}. Later benchmarks such as MathOctopus, MMATH, and PolyMath extend the evaluation to stronger models and more difficult settings, finding large performance disparities across languages and frequent language-consistency issues in reasoning traces \citep{chen2024mathoctopus,luo2025mmath,wang2025polymath}.

\paragraph{Multilingual reasoning transfer before RLVR.}
Several methods improve multilingual reasoning without multilingual reinforcement learning. CLP uses crosslingual prompting for zero-shot chain-of-thought reasoning \citep{qin2023clp}, xCoT transfers reasoning from high-resource to low-resource languages through crosslingual instruction tuning \citep{chai2024xcot}, and mCoT studies multilingual chain-of-thought instruction tuning for reasoning consistency \citep{lai2024mcotmultilingualinstructiontuning}. LangBridge combines multilingual understanding with an English reasoning-specialized backbone \citep{yoon2024langbridge}, while SLAM selectively aligns language-relevant layers for efficient multilingual reasoning \citep{fan2025slamefficientmultilingualreasoning}. Finally, \citet{ranaldi2025selftraining} propose self-training through a more language-agnostic intermediate reasoning space. These approaches provide complementary context for our focus on the crosslingual transfer under GRPO rather than prompting or supervised alignment.

\section{Full Cross-Lingual Transfer Matrices}
\label{app:matrices}
\subsection*{}

\paragraph{Reading the matrices.}
Each cell reports Avg@8 percentage-point improvement over the base model. Rows are training-language and reasoning-reward settings; columns are evaluation languages. Cells in each column are colored based on relative performance, ranging from {\color[HTML]{C62828} red (worst performance)} through {\color[HTML]{4A90D9} blue (mean performance)} to {\color[HTML]{2E7D32} green (best performance)}. Bold entries mark the larger reasoning-reward value within a training language for that evaluation language, and underlined entries mark the best value in the column. The final Avg column reports each row's mean over the displayed evaluation-language columns. Reward labels followed by ``ineffective'' mark model, training-language, and reward combinations where the model had non-negligible base support for the target reasoning language but the trained model did not converge to that language. 

\paragraph{Macro definitions.}
Easy Math is the macro average of MGSM and PolyMath-low. Since PolyMath-low includes more languages than MGSM, in the main paper we average only over the languages with support in both datasets. We provide the full results here in the Appendix for completeness. Hard Math is the macro average of PolyMath-medium, PolyMath-high, and PolyMath-top. Average over all unseen tasks from Multilingual Reasoning Gym (MRG) is the equal-weight macro average of MRG heldout, English-content, and very difficult MRG tasks. Macro average over all unseen tasks is the equal-weight macro average of MRG all-unseen, Easy Math, and Hard Math.

\onecolumn

\newcolumntype{V}{@{\hspace{3pt}}@{\hspace{3pt}}}

\FloatBarrier
\subsection{Qwen3-1.7B-Base}

\begin{table}[!htbp]
\centering
\begin{minipage}[t]{0.49\textwidth}
\centering
\caption{Tasks seen during training from Multilingual Reasoning Gym -- Avg@8 percentage-point improvement over the base model \texttt{Qwen3-1.7B-Base}}
\label{tab:appendix_crosslingual_qwen3_1_7b_base_1_train_filtered_noheldout}
\tiny
\setlength{\tabcolsep}{2pt}
\renewcommand{\arraystretch}{0.92}
\newcommand{\MatrixResizeWidth}{\linewidth}
\resizebox{\MatrixResizeWidth}{!}{%
\begin{tabular}{llrrrrrrrrrrrrrrVr}
\toprule
Train & Reward & en & zh & de & fr & es & ru & ja & th & bn & te & sw & it & ko & pt & Avg \\
\midrule
\multirow{1}{*}{en} & English & \cellcolor[HTML]{5FBD81}\textbf{37.3} & \cellcolor[HTML]{9CC7E6}\textbf{34.4} & \cellcolor[HTML]{B8D3EC}\textbf{34.5} & \cellcolor[HTML]{C6DAE9}\textbf{34.4} & \cellcolor[HTML]{E4EBEA}\textbf{34.1} & \cellcolor[HTML]{A9CCF2}\textbf{34.5} & \cellcolor[HTML]{A5CAEE}\textbf{34.4} & \cellcolor[HTML]{D4E2E8}\textbf{34.0} & \cellcolor[HTML]{F0E7B9}\textbf{30.7} & \cellcolor[HTML]{F5D596}\textbf{28.0} & \cellcolor[HTML]{EA6C62}\textbf{19.4} & \cellcolor[HTML]{D8E4E8}\textbf{34.3} & \cellcolor[HTML]{A2C9EC}\textbf{33.0} & \cellcolor[HTML]{BFD7EB}\textbf{33.2} & \cellcolor[HTML]{B1CFEF}\textbf{32.6} \\
\midrule
\multirow{2}{*}{zh} & Native & \cellcolor[HTML]{98C5E1}34.5 & \cellcolor[HTML]{67BD90}38.0 & \cellcolor[HTML]{D9E4E8}33.8 & \cellcolor[HTML]{D1E0E8}34.1 & \cellcolor[HTML]{B7D2ED}35.4 & \cellcolor[HTML]{A2C9EB}34.8 & \cellcolor[HTML]{C8DBE9}33.3 & \cellcolor[HTML]{EBECDD}32.9 & \cellcolor[HTML]{E85C5C}25.4 & \cellcolor[HTML]{EC8A6D}25.2 & \cellcolor[HTML]{E85C5C}19.0 & \cellcolor[HTML]{F0F3F0}33.7 & \cellcolor[HTML]{D2E1E8}31.5 & \cellcolor[HTML]{9BC7E5}34.5 & \cellcolor[HTML]{D3E2E8}31.9 \\
 & English & \cellcolor[HTML]{6DBD9F}\textbf{36.6} & \cellcolor[HTML]{5CBE78}\underline{\textbf{38.8}} & \cellcolor[HTML]{75BEAD}\textbf{37.0} & \cellcolor[HTML]{81C0C2}\textbf{36.9} & \cellcolor[HTML]{85C1C8}\textbf{37.6} & \cellcolor[HTML]{6BBE99}\textbf{37.8} & \cellcolor[HTML]{7DBFBB}\textbf{36.6} & \cellcolor[HTML]{83C0C6}\textbf{36.3} & \cellcolor[HTML]{DDE7E9}\textbf{33.4} & \cellcolor[HTML]{EBE9CE}\textbf{30.3} & \cellcolor[HTML]{EA7565}\textbf{19.7} & \cellcolor[HTML]{91C3D9}\textbf{36.5} & \cellcolor[HTML]{70BEA4}\textbf{35.9} & \cellcolor[HTML]{83C0C6}\textbf{35.8} & \cellcolor[HTML]{73BEA9}\textbf{34.9} \\
\bottomrule
\end{tabular}%
}
\end{minipage}
\hfill
\begin{minipage}[t]{0.49\textwidth}
\centering
\caption{Macro average over all unseen tasks -- Avg@8 percentage-point improvement over the base model \texttt{Qwen3-1.7B-Base}}
\label{tab:appendix_crosslingual_qwen3_1_7b_base_2_appendix_unseen_task_groups_macro}
\tiny
\setlength{\tabcolsep}{2pt}
\renewcommand{\arraystretch}{0.92}
\newcommand{\MatrixResizeWidth}{\linewidth}
\resizebox{\MatrixResizeWidth}{!}{%
\begin{tabular}{llrrrrrrrrrrrVr}
\toprule
Train & Reward & en & zh & de & fr & es & ru & ja & th & bn & te & sw & Avg \\
\midrule
\multirow{1}{*}{en} & English & \cellcolor[HTML]{6EBD9F}\textbf{27.2} & \cellcolor[HTML]{8CC2D2}\textbf{24.1} & \cellcolor[HTML]{91C3D9}\textbf{23.4} & \cellcolor[HTML]{81C0C2}\textbf{25.7} & \cellcolor[HTML]{7ABFB7}\textbf{26.9} & \cellcolor[HTML]{95C4DD}\textbf{23.9} & \cellcolor[HTML]{A7CBF0}\textbf{20.9} & \cellcolor[HTML]{A6CBEF}\textbf{21.8} & \cellcolor[HTML]{9DC7E7}\textbf{19.9} & \cellcolor[HTML]{9CC7E6}\textbf{14.8} & \cellcolor[HTML]{EEE8C0}\textbf{5.1} & \cellcolor[HTML]{88C2CD}\textbf{21.2} \\
\midrule
\multirow{2}{*}{zh} & Native & \cellcolor[HTML]{B6D2ED}24.3 & \cellcolor[HTML]{5CBE78}\underline{\textbf{25.9}} & \cellcolor[HTML]{C0D7EB}22.2 & \cellcolor[HTML]{F0F2EC}22.8 & \cellcolor[HTML]{BED6EB}24.4 & \cellcolor[HTML]{B1CFEF}23.0 & \cellcolor[HTML]{88C1CE}22.0 & \cellcolor[HTML]{EFF2EF}20.7 & \cellcolor[HTML]{F5D99A}16.8 & \cellcolor[HTML]{F4D495}12.2 & \cellcolor[HTML]{F2E7B4}\textbf{5.0} & \cellcolor[HTML]{C6DAE9}19.9 \\
 & English & \cellcolor[HTML]{6FBEA2}\textbf{27.2} & \cellcolor[HTML]{7EBFBC}24.6 & \cellcolor[HTML]{6EBEA0}\textbf{24.8} & \cellcolor[HTML]{6CBD9B}\textbf{26.5} & \cellcolor[HTML]{74BEAC}\textbf{27.2} & \cellcolor[HTML]{6CBE9B}\textbf{25.5} & \cellcolor[HTML]{6BBE98}\textbf{23.3} & \cellcolor[HTML]{72BEA8}\textbf{23.3} & \cellcolor[HTML]{73BEAA}\textbf{21.3} & \cellcolor[HTML]{93C4DC}\textbf{15.0} & \cellcolor[HTML]{F3C78D}4.9 & \cellcolor[HTML]{6CBD9B}\textbf{22.1} \\
\midrule
\multirow{2}{*}{de} & Native & \cellcolor[HTML]{ACCDF0}24.5 & \cellcolor[HTML]{EBEAD1}21.3 & \cellcolor[HTML]{7FBFBE}24.1 & \cellcolor[HTML]{BFD7EB}23.9 & \cellcolor[HTML]{A7CBF0}24.9 & \cellcolor[HTML]{D5E2E8}22.2 & \cellcolor[HTML]{CBDDE9}20.2 & \cellcolor[HTML]{ECEEE3}20.4 & \cellcolor[HTML]{F5F7F4}18.4 & \cellcolor[HTML]{D2E1E8}14.0 & \cellcolor[HTML]{B2D0EF}\textbf{5.3} & \cellcolor[HTML]{C6DAEA}19.9 \\
\bottomrule
\end{tabular}%
}
\end{minipage}
\end{table}

\begin{table}[!htbp]
\centering
\begin{minipage}[t]{0.49\textwidth}
\centering
\caption{Average over all unseen tasks from Multilingual Reasoning Gym -- Avg@8 percentage-point improvement over the base model \texttt{Qwen3-1.7B-Base}}
\label{tab:appendix_crosslingual_qwen3_1_7b_base_3_appendix_mrg_unseen_macro}
\tiny
\setlength{\tabcolsep}{2pt}
\renewcommand{\arraystretch}{0.92}
\newcommand{\MatrixResizeWidth}{\linewidth}
\resizebox{\MatrixResizeWidth}{!}{%
\begin{tabular}{llrrrrrrrrrrrrrrVr}
\toprule
Train & Reward & en & zh & de & fr & es & ru & ja & th & bn & te & sw & it & ko & pt & Avg \\
\midrule
\multirow{1}{*}{en} & English & \cellcolor[HTML]{5CBE78}\underline{\textbf{10.6}} & \cellcolor[HTML]{A9CCF1}\textbf{8.2} & \cellcolor[HTML]{EEF0E8}\textbf{7.8} & \cellcolor[HTML]{8BC2D1}\textbf{9.9} & \cellcolor[HTML]{9DC7E7}\textbf{10.0} & \cellcolor[HTML]{E7EDEB}\textbf{7.9} & \cellcolor[HTML]{F5DD9C}\textbf{8.9} & \cellcolor[HTML]{EDF0E6}\textbf{8.3} & \cellcolor[HTML]{F5DD9D}\textbf{8.0} & \cellcolor[HTML]{F6E2A1}\textbf{6.9} & \cellcolor[HTML]{ED8E6F}\textbf{4.5} & \cellcolor[HTML]{8BC2D1}\textbf{9.5} & \cellcolor[HTML]{A1C9EA}\textbf{9.6} & \cellcolor[HTML]{A4CAED}\textbf{10.1} & \cellcolor[HTML]{F4F7F4}\textbf{8.6} \\
\midrule
\multirow{2}{*}{zh} & Native & \cellcolor[HTML]{71BEA5}\textbf{10.3} & \cellcolor[HTML]{5CBE78}\underline{\textbf{8.9}} & \cellcolor[HTML]{F3E7B0}\textbf{7.5} & \cellcolor[HTML]{85C1C8}9.9 & \cellcolor[HTML]{A3C9EC}\textbf{9.9} & \cellcolor[HTML]{E85C5C}6.3 & \cellcolor[HTML]{ECEEE3}9.3 & \cellcolor[HTML]{EEE8C1}8.0 & \cellcolor[HTML]{E85C5C}7.0 & \cellcolor[HTML]{E85C5C}5.9 & \cellcolor[HTML]{EE9974}\textbf{4.5} & \cellcolor[HTML]{8FC3D6}9.5 & \cellcolor[HTML]{F1F3EE}9.0 & \cellcolor[HTML]{E1E9E9}9.7 & \cellcolor[HTML]{F7E6A5}8.3 \\
 & English & \cellcolor[HTML]{A0C8E9}9.7 & \cellcolor[HTML]{ECF1EE}7.9 & \cellcolor[HTML]{F4E7AE}7.5 & \cellcolor[HTML]{6FBEA3}\textbf{10.2} & \cellcolor[HTML]{A9CCF2}9.9 & \cellcolor[HTML]{ECEDDF}\textbf{7.7} & \cellcolor[HTML]{EDEFE4}\textbf{9.3} & \cellcolor[HTML]{ECE8C8}\textbf{8.1} & \cellcolor[HTML]{F5F7F4}\textbf{8.7} & \cellcolor[HTML]{F2C38A}\textbf{6.6} & \cellcolor[HTML]{E85C5C}4.3 & \cellcolor[HTML]{81C0C2}\textbf{9.7} & \cellcolor[HTML]{B9D4EC}\textbf{9.4} & \cellcolor[HTML]{CFDFE8}\textbf{9.8} & \cellcolor[HTML]{EBEBD8}\textbf{8.5} \\
\bottomrule
\end{tabular}%
}
\end{minipage}
\hfill
\begin{minipage}[t]{0.49\textwidth}
\centering
\caption{Main heldout tasks from Multilingual Reasoning Gym -- Avg@8 percentage-point improvement over the base model \texttt{Qwen3-1.7B-Base}}
\label{tab:appendix_crosslingual_qwen3_1_7b_base_4_heldout_filtered}
\tiny
\setlength{\tabcolsep}{2pt}
\renewcommand{\arraystretch}{0.92}
\newcommand{\MatrixResizeWidth}{\linewidth}
\resizebox{\MatrixResizeWidth}{!}{%
\begin{tabular}{llrrrrrrrrrrrrrrVr}
\toprule
Train & Reward & en & zh & de & fr & es & ru & ja & th & bn & te & sw & it & ko & pt & Avg \\
\midrule
\multirow{1}{*}{en} & English & \cellcolor[HTML]{EBEBD8}\textbf{11.0} & \cellcolor[HTML]{EDE8C6}\textbf{10.4} & \cellcolor[HTML]{D9E5E8}\textbf{9.7} & \cellcolor[HTML]{EEE7BF}\textbf{10.2} & \cellcolor[HTML]{F4D193}\textbf{10.4} & \cellcolor[HTML]{ECE9CB}\textbf{9.9} & \cellcolor[HTML]{EEA077}\textbf{10.0} & \cellcolor[HTML]{ECEEE2}\textbf{9.6} & \cellcolor[HTML]{F1B783}\textbf{9.5} & \cellcolor[HTML]{F1B984}\textbf{10.0} & \cellcolor[HTML]{EE9973}\textbf{10.6} & \cellcolor[HTML]{F2BC86}\textbf{9.6} & \cellcolor[HTML]{EE9F77}\textbf{10.6} & \cellcolor[HTML]{F5E7A9}\textbf{10.9} & \cellcolor[HTML]{F2BD86}\textbf{10.2} \\
\midrule
\multirow{2}{*}{zh} & Native & \cellcolor[HTML]{91C3D9}\textbf{12.2} & \cellcolor[HTML]{5CBE78}\underline{\textbf{12.7}} & \cellcolor[HTML]{F2C48B}8.7 & \cellcolor[HTML]{F2F5F1}\textbf{10.7} & \cellcolor[HTML]{F4CC90}\textbf{10.3} & \cellcolor[HTML]{EB7A67}8.5 & \cellcolor[HTML]{F0AE7E}\textbf{10.2} & \cellcolor[HTML]{CDDEE8}\textbf{10.3} & \cellcolor[HTML]{F5DC9C}\textbf{10.1} & \cellcolor[HTML]{F2BA85}\textbf{10.0} & \cellcolor[HTML]{EE9873}\textbf{10.6} & \cellcolor[HTML]{F5DA9A}\textbf{9.9} & \cellcolor[HTML]{F0F2EC}\textbf{12.1} & \cellcolor[HTML]{EBECDC}\textbf{11.5} & \cellcolor[HTML]{F6DF9E}\textbf{10.6} \\
 & English & \cellcolor[HTML]{F4CE91}10.4 & \cellcolor[HTML]{F5E6AC}10.2 & \cellcolor[HTML]{F4CD90}\textbf{8.7} & \cellcolor[HTML]{ECEFE3}10.6 & \cellcolor[HTML]{EE9270}9.5 & \cellcolor[HTML]{F4D394}\textbf{9.4} & \cellcolor[HTML]{EE9572}9.9 & \cellcolor[HTML]{EFE7BD}9.2 & \cellcolor[HTML]{EE9A74}9.1 & \cellcolor[HTML]{ED906F}9.3 & \cellcolor[HTML]{E85C5C}9.7 & \cellcolor[HTML]{F0A87B}9.4 & \cellcolor[HTML]{EA6861}10.1 & \cellcolor[HTML]{EEA077}9.8 & \cellcolor[HTML]{EB7D68}9.7 \\
\bottomrule
\end{tabular}%
}
\end{minipage}
\end{table}

\begin{table}[!htbp]
\centering
\begin{minipage}[t]{0.49\textwidth}
\centering
\caption{English-content tasks from Multilingual Reasoning Gym, not seen during training -- Avg@8 percentage-point improvement over the base model \texttt{Qwen3-1.7B-Base}}
\label{tab:appendix_crosslingual_qwen3_1_7b_base_5_english_content_add}
\tiny
\setlength{\tabcolsep}{2pt}
\renewcommand{\arraystretch}{0.92}
\newcommand{\MatrixResizeWidth}{\linewidth}
\resizebox{\MatrixResizeWidth}{!}{%
\begin{tabular}{llrrrrrrrrrrrrrrVr}
\toprule
Train & Reward & en & zh & de & fr & es & ru & ja & th & bn & te & sw & it & ko & pt & Avg \\
\midrule
\multirow{1}{*}{en} & English & \cellcolor[HTML]{5CBE78}\underline{\textbf{20.8}} & \cellcolor[HTML]{5CBE78}\underline{\textbf{14.2}} & \cellcolor[HTML]{EBEBD6}\textbf{13.6} & \cellcolor[HTML]{8AC1CF}\textbf{19.3} & \cellcolor[HTML]{A3C9EC}\textbf{19.6} & \cellcolor[HTML]{E3EBEA}\textbf{13.7} & \cellcolor[HTML]{CBDDE8}\textbf{16.8} & \cellcolor[HTML]{F1F3EE}\textbf{15.2} & \cellcolor[HTML]{E9EFEC}\textbf{14.4} & \cellcolor[HTML]{96C5DF}\textbf{10.8} & \cellcolor[HTML]{EDEFE4}\textbf{2.8} & \cellcolor[HTML]{92C4DA}\textbf{18.9} & \cellcolor[HTML]{83C0C6}\textbf{18.1} & \cellcolor[HTML]{83C0C5}\textbf{19.2} & \cellcolor[HTML]{87C1CC}\textbf{15.5} \\
\midrule
\multirow{2}{*}{zh} & Native & \cellcolor[HTML]{87C1CC}18.7 & \cellcolor[HTML]{6BBE9A}\textbf{13.9} & \cellcolor[HTML]{EBEDDE}\textbf{13.7} & \cellcolor[HTML]{8FC3D7}19.1 & \cellcolor[HTML]{A6CAEF}19.5 & \cellcolor[HTML]{E85E5D}10.4 & \cellcolor[HTML]{99C6E2}17.7 & \cellcolor[HTML]{F5D696}13.7 & \cellcolor[HTML]{EE9873}10.8 & \cellcolor[HTML]{E85C5C}7.5 & \cellcolor[HTML]{E6ECEA}2.9 & \cellcolor[HTML]{9FC8E8}18.5 & \cellcolor[HTML]{EBEAD1}14.8 & \cellcolor[HTML]{B4D1EE}17.6 & \cellcolor[HTML]{CCDDE9}14.2 \\
 & English & \cellcolor[HTML]{86C1CB}\textbf{18.7} & \cellcolor[HTML]{7ABEB5}13.6 & \cellcolor[HTML]{EBEBD8}13.6 & \cellcolor[HTML]{7CBFB9}\textbf{19.9} & \cellcolor[HTML]{96C5DF}\textbf{20.1} & \cellcolor[HTML]{E6ECEB}\textbf{13.6} & \cellcolor[HTML]{91C4D8}\textbf{18.0} & \cellcolor[HTML]{EBECDC}\textbf{14.9} & \cellcolor[HTML]{8BC2D1}\textbf{16.7} & \cellcolor[HTML]{A4CAED}\textbf{10.5} & \cellcolor[HTML]{D2E1E8}\textbf{3.0} & \cellcolor[HTML]{84C0C6}\textbf{19.6} & \cellcolor[HTML]{82C0C4}\textbf{18.1} & \cellcolor[HTML]{7ABEB5}\textbf{19.6} & \cellcolor[HTML]{81C0C1}\textbf{15.7} \\
\bottomrule
\end{tabular}%
}
\end{minipage}
\hfill
\begin{minipage}[t]{0.49\textwidth}
\centering
\caption{Very difficult tasks from Multilingual Reasoning Gym, not seen during training -- Avg@8 percentage-point improvement over the base model \texttt{Qwen3-1.7B-Base}}
\label{tab:appendix_crosslingual_qwen3_1_7b_base_6_difficulty_addback}
\tiny
\setlength{\tabcolsep}{2pt}
\renewcommand{\arraystretch}{0.92}
\newcommand{\MatrixResizeWidth}{\linewidth}
\resizebox{\MatrixResizeWidth}{!}{%
\begin{tabular}{llrrrrrrrrrrrrrrVr}
\toprule
Train & Reward & en & zh & de & fr & es & ru & ja & th & bn & te & sw & it & ko & pt & Avg \\
\midrule
\multirow{1}{*}{en} & English & \cellcolor[HTML]{E85C5C}\textbf{0.0} & \cellcolor[HTML]{F6F8F5}\textbf{0.1} & \cellcolor[HTML]{E2EAE9}\textbf{0.1} & \cellcolor[HTML]{F7E7A6}\textbf{0.1} & \cellcolor[HTML]{F7E7A6}\textbf{-0.0} & \cellcolor[HTML]{EBEBD7}\textbf{0.1} & \cellcolor[HTML]{F3C68C}\textbf{0.0} & \cellcolor[HTML]{93C4DB}\textbf{0.2} & \cellcolor[HTML]{F7E7A6}\textbf{0.1} & \cellcolor[HTML]{F2E7B3}\textbf{0.1} & \cellcolor[HTML]{EBEBDA}\textbf{0.1} & \cellcolor[HTML]{F4CC91}\textbf{0.0} & \cellcolor[HTML]{ECEDDF}\textbf{0.2} & \cellcolor[HTML]{5CBE78}\underline{\textbf{0.1}} & \cellcolor[HTML]{DCE6E8}\textbf{0.1} \\
\midrule
\multirow{2}{*}{zh} & Native & \cellcolor[HTML]{A9CCF1}0.1 & \cellcolor[HTML]{F2C088}\textbf{0.0} & \cellcolor[HTML]{F4CC91}\textbf{0.0} & \cellcolor[HTML]{F7E7A6}\textbf{0.1} & \cellcolor[HTML]{EBEBDA}\textbf{-0.0} & \cellcolor[HTML]{EB7D68}0.0 & \cellcolor[HTML]{ECEEE1}\textbf{0.1} & \cellcolor[HTML]{ECE9CD}0.1 & \cellcolor[HTML]{EEF0E7}0.2 & \cellcolor[HTML]{F5D797}\textbf{0.1} & \cellcolor[HTML]{73BEAA}\textbf{0.1} & \cellcolor[HTML]{F4CC91}\textbf{0.0} & \cellcolor[HTML]{F4CD90}\textbf{0.1} & \cellcolor[HTML]{F2C38A}-0.0 & \cellcolor[HTML]{F7E7A6}\textbf{0.1} \\
 & English & \cellcolor[HTML]{8DC2D3}\textbf{0.1} & \cellcolor[HTML]{F2C088}\textbf{0.0} & \cellcolor[HTML]{F0AD7D}0.0 & \cellcolor[HTML]{EE9E76}0.0 & \cellcolor[HTML]{F2C38A}-0.0 & \cellcolor[HTML]{F1B582}\textbf{0.1} & \cellcolor[HTML]{ECEEE1}\textbf{0.1} & \cellcolor[HTML]{F6F8F6}\textbf{0.1} & \cellcolor[HTML]{E9EEEC}\textbf{0.2} & \cellcolor[HTML]{F5D797}\textbf{0.1} & \cellcolor[HTML]{A9CCF1}0.1 & \cellcolor[HTML]{F4CC91}\textbf{0.0} & \cellcolor[HTML]{F2BE87}0.1 & \cellcolor[HTML]{EBEBDA}\textbf{0.0} & \cellcolor[HTML]{F5DC9C}0.1 \\
\bottomrule
\end{tabular}%
}
\end{minipage}
\end{table}

\begin{table}[!htbp]
\centering
\begin{minipage}[t]{0.49\textwidth}
\centering
\caption{Easy Math (MGSM + PolyMath low) -- Avg@8 percentage-point improvement over the base model \texttt{Qwen3-1.7B-Base}; $\dagger$: includes only PolyMath low because MGSM is unavailable for this language.}
\label{tab:appendix_crosslingual_qwen3_1_7b_base_7_appendix_easy_math_macro}
\tiny
\setlength{\tabcolsep}{2pt}
\renewcommand{\arraystretch}{0.92}
\newcommand{\MatrixResizeWidth}{\linewidth}
\resizebox{\MatrixResizeWidth}{!}{%
\begin{tabular}{llrrrrrrrrrrrrrrrrrrVr}
\toprule
Train & Reward & en & zh & de & fr & es & ru & ja & th & bn & te & sw & it$^{\dagger}$ & ko$^{\dagger}$ & pt$^{\dagger}$ & ar$^{\dagger}$ & id$^{\dagger}$ & ms$^{\dagger}$ & vi$^{\dagger}$ & Avg \\
\midrule
\multirow{1}{*}{en} & English & \cellcolor[HTML]{78BEB2}\textbf{63.7} & \cellcolor[HTML]{8CC2D2}\textbf{57.2} & \cellcolor[HTML]{8DC2D3}\textbf{55.8} & \cellcolor[HTML]{7DBFBB}\textbf{60.7} & \cellcolor[HTML]{78BEB2}\textbf{63.6} & \cellcolor[HTML]{95C5DF}\textbf{56.8} & \cellcolor[HTML]{9EC8E8}\textbf{47.3} & \cellcolor[HTML]{ABCDF1}\textbf{50.7} & \cellcolor[HTML]{A0C8E9}\textbf{45.2} & \cellcolor[HTML]{C1D8EA}\textbf{31.4} & \cellcolor[HTML]{D9E5E8}\textbf{7.0} & \cellcolor[HTML]{9AC6E4}\textbf{59.4} & \cellcolor[HTML]{A8CBF0}\textbf{51.7} & \cellcolor[HTML]{8BC2D1}\textbf{57.6} & \cellcolor[HTML]{B4D1EE}\textbf{48.1} & \cellcolor[HTML]{88C1CE}\textbf{57.2} & \cellcolor[HTML]{95C5DE}\textbf{52.7} & \cellcolor[HTML]{93C4DB}\textbf{54.2} & \cellcolor[HTML]{8BC2D1}\textbf{51.1} \\
\midrule
\multirow{2}{*}{zh} & Native & \cellcolor[HTML]{CBDDE9}56.1 & \cellcolor[HTML]{5CBE78}\underline{\textbf{61.2}} & \cellcolor[HTML]{B2D0EE}52.8 & \cellcolor[HTML]{EBECDB}52.0 & \cellcolor[HTML]{C8DBE9}56.2 & \cellcolor[HTML]{9CC7E6}56.1 & \cellcolor[HTML]{85C1C8}49.8 & \cellcolor[HTML]{EAEFEC}48.2 & \cellcolor[HTML]{F6E1A0}38.1 & \cellcolor[HTML]{F4D293}26.3 & \cellcolor[HTML]{F5D898}\textbf{6.3} & \cellcolor[HTML]{D1E0E8}55.6 & \cellcolor[HTML]{DDE7E9}49.2 & \cellcolor[HTML]{D4E2E8}52.7 & \cellcolor[HTML]{EDEFE5}44.7 & \cellcolor[HTML]{98C6E1}55.7 & \cellcolor[HTML]{69BD96}57.1 & \cellcolor[HTML]{61BE84}\textbf{58.0} & \cellcolor[HTML]{AECEF0}48.7 \\
 & English & \cellcolor[HTML]{78BFB3}\textbf{63.6} & \cellcolor[HTML]{7CBFBA}58.5 & \cellcolor[HTML]{6CBE9B}\textbf{59.1} & \cellcolor[HTML]{70BEA5}\textbf{61.9} & \cellcolor[HTML]{6ABE98}\textbf{65.3} & \cellcolor[HTML]{69BD96}\textbf{61.8} & \cellcolor[HTML]{5EBE7C}\textbf{54.4} & \cellcolor[HTML]{6ABE97}\textbf{55.5} & \cellcolor[HTML]{7BBFB9}\textbf{47.9} & \cellcolor[HTML]{9BC7E5}\textbf{32.9} & \cellcolor[HTML]{F2B984}6.2 & \cellcolor[HTML]{85C1C8}\textbf{61.8} & \cellcolor[HTML]{72BEA8}\textbf{56.4} & \cellcolor[HTML]{74BFAC}\textbf{59.9} & \cellcolor[HTML]{85C1C8}\textbf{51.3} & \cellcolor[HTML]{7BBFB7}\textbf{58.7} & \cellcolor[HTML]{66BE8F}\textbf{57.4} & \cellcolor[HTML]{6ABE96}57.3 & \cellcolor[HTML]{6BBD9A}\textbf{53.9} \\
\midrule
\multirow{2}{*}{de} & Native & \cellcolor[HTML]{ACCDF1}57.9 & \cellcolor[HTML]{EBEBD6}50.8 & \cellcolor[HTML]{6EBEA0}58.8 & \cellcolor[HTML]{ABCDF1}56.7 & \cellcolor[HTML]{98C6E1}59.8 & \cellcolor[HTML]{B1CFEF}54.5 & \cellcolor[HTML]{A9CBF1}46.3 & \cellcolor[HTML]{EBEDDE}47.1 & \cellcolor[HTML]{E5ECEA}42.1 & \cellcolor[HTML]{E8EEEB}30.1 & \cellcolor[HTML]{96C5DF}\textbf{7.5} & \cellcolor[HTML]{B8D3ED}57.1 & \cellcolor[HTML]{F3F5F1}48.0 & \cellcolor[HTML]{BAD4EC}54.0 & \cellcolor[HTML]{ECEEE0}44.5 & \cellcolor[HTML]{A1C9EA}54.9 & \cellcolor[HTML]{B7D2ED}50.4 & \cellcolor[HTML]{ECE9CA}48.4 & \cellcolor[HTML]{B9D3ED}48.3 \\
 & English & \cellcolor[HTML]{6DBE9D}\textbf{65.0} & \cellcolor[HTML]{7ABFB5}\textbf{58.6} & \cellcolor[HTML]{5DBE7A}\textbf{60.7} & \cellcolor[HTML]{5CBE78}\underline{\textbf{64.1}} & \cellcolor[HTML]{5CBE79}\textbf{67.2} & \cellcolor[HTML]{62BD86}\textbf{62.7} & \cellcolor[HTML]{5CBE78}\underline{\textbf{54.6}} & \cellcolor[HTML]{5EBE7E}\textbf{56.5} & \cellcolor[HTML]{5CBE78}\underline{\textbf{50.6}} & \cellcolor[HTML]{76BFAF}\textbf{35.1} & \cellcolor[HTML]{E6ECEB}7.0 & \cellcolor[HTML]{72BEA8}\textbf{64.1} & \cellcolor[HTML]{5CBE78}\underline{\textbf{58.6}} & \cellcolor[HTML]{5EBE7C}\textbf{62.4} & \cellcolor[HTML]{7ABFB5}\textbf{52.3} & \cellcolor[HTML]{68BE93}\textbf{60.7} & \cellcolor[HTML]{5CBE78}\underline{\textbf{58.5}} & \cellcolor[HTML]{79BFB4}\textbf{56.1} & \cellcolor[HTML]{5CBE78}\underline{\textbf{55.3}} \\
\midrule
\multirow{2}{*}{fr} & Native & \cellcolor[HTML]{A7CBF0}58.3 & \cellcolor[HTML]{E0E8E9}52.9 & \cellcolor[HTML]{A8CBF1}53.3 & \cellcolor[HTML]{8EC3D4}59.1 & \cellcolor[HTML]{86C1CB}61.7 & \cellcolor[HTML]{A2C9EC}55.5 & \cellcolor[HTML]{AACCF1}46.2 & \cellcolor[HTML]{9FC8E8}51.5 & \cellcolor[HTML]{EDE8C4}39.6 & \cellcolor[HTML]{F3E7B0}27.6 & \cellcolor[HTML]{F0A87B}6.0 & \cellcolor[HTML]{A1C8EA}58.8 & \cellcolor[HTML]{DBE6E8}49.3 & \cellcolor[HTML]{97C5E0}56.4 & \cellcolor[HTML]{EEF0E7}44.8 & \cellcolor[HTML]{C2D8EA}53.0 & \cellcolor[HTML]{B0CFEF}50.7 & \cellcolor[HTML]{F0F3F0}50.1 & \cellcolor[HTML]{B0CFEF}48.6 \\
 & English & \cellcolor[HTML]{78BFB3}\textbf{63.6} & \cellcolor[HTML]{8BC2D0}\textbf{57.3} & \cellcolor[HTML]{81C0C2}\textbf{56.9} & \cellcolor[HTML]{78BFB3}\textbf{61.2} & \cellcolor[HTML]{67BD90}\textbf{65.8} & \cellcolor[HTML]{75BEAE}\textbf{60.3} & \cellcolor[HTML]{6FBEA3}\textbf{52.3} & \cellcolor[HTML]{6BBE99}\textbf{55.4} & \cellcolor[HTML]{85C1C8}\textbf{47.1} & \cellcolor[HTML]{C6DAEA}\textbf{31.2} & \cellcolor[HTML]{F1F3EE}\textbf{6.9} & \cellcolor[HTML]{91C3D9}\textbf{60.4} & \cellcolor[HTML]{B0CFEF}\textbf{51.3} & \cellcolor[HTML]{86C1CB}\textbf{58.0} & \cellcolor[HTML]{9DC7E6}\textbf{49.4} & \cellcolor[HTML]{78BEB2}\textbf{59.0} & \cellcolor[HTML]{82C0C4}\textbf{54.5} & \cellcolor[HTML]{7CBFB9}\textbf{55.9} & \cellcolor[HTML]{7ABEB5}\textbf{52.6} \\
\midrule
\multirow{2}{*}{es} & Native & \cellcolor[HTML]{94C4DC}60.3 & \cellcolor[HTML]{AACCF1}55.0 & \cellcolor[HTML]{B1CFEF}52.9 & \cellcolor[HTML]{A2C9EB}57.4 & \cellcolor[HTML]{70BEA4}64.6 & \cellcolor[HTML]{9AC6E3}56.4 & \cellcolor[HTML]{90C3D7}48.7 & \cellcolor[HTML]{E6ECEA}48.4 & \cellcolor[HTML]{E6ECEA}42.1 & \cellcolor[HTML]{ECE8C7}28.3 & \cellcolor[HTML]{9DC8E7}\textbf{7.4} & \cellcolor[HTML]{A3C9ED}58.5 & \cellcolor[HTML]{F3F5F1}48.0 & \cellcolor[HTML]{85C1C9}58.1 & \cellcolor[HTML]{9FC8E8}49.3 & \cellcolor[HTML]{A4CAED}54.6 & \cellcolor[HTML]{98C5E1}52.5 & \cellcolor[HTML]{8DC3D4}54.6 & \cellcolor[HTML]{9BC7E5}49.8 \\
 & English & \cellcolor[HTML]{70BDA3}\textbf{64.6} & \cellcolor[HTML]{8EC2D5}\textbf{57.0} & \cellcolor[HTML]{67BD91}\textbf{59.6} & \cellcolor[HTML]{6CBE9C}\textbf{62.3} & \cellcolor[HTML]{5FBE7F}\textbf{66.8} & \cellcolor[HTML]{6ABE97}\textbf{61.7} & \cellcolor[HTML]{6EBEA0}\textbf{52.4} & \cellcolor[HTML]{78BEB2}\textbf{54.4} & \cellcolor[HTML]{7DBFBB}\textbf{47.8} & \cellcolor[HTML]{99C6E3}\textbf{33.0} & \cellcolor[HTML]{EBEAD2}6.7 & \cellcolor[HTML]{7FC0C0}\textbf{62.5} & \cellcolor[HTML]{83C0C5}\textbf{54.8} & \cellcolor[HTML]{6EBE9F}\textbf{60.6} & \cellcolor[HTML]{8FC3D6}\textbf{50.5} & \cellcolor[HTML]{7EC0BD}\textbf{58.3} & \cellcolor[HTML]{71BEA7}\textbf{56.2} & \cellcolor[HTML]{85C1C7}\textbf{55.2} & \cellcolor[HTML]{6EBEA0}\textbf{53.6} \\
\midrule
\multirow{2}{*}{ru} & Native & \cellcolor[HTML]{A3C9EC}58.7 & \cellcolor[HTML]{EDEFE5}51.5 & \cellcolor[HTML]{E9EFEC}50.0 & \cellcolor[HTML]{F0F4F0}53.3 & \cellcolor[HTML]{CFDFE8}55.7 & \cellcolor[HTML]{6BBD9B}61.5 & \cellcolor[HTML]{C7DBE9}44.7 & \cellcolor[HTML]{9DC7E6}51.6 & \cellcolor[HTML]{EBEBD6}40.3 & \cellcolor[HTML]{F7E4A3}27.2 & \cellcolor[HTML]{F4D092}\textbf{6.3} & \cellcolor[HTML]{E1E9E9}54.6 & \cellcolor[HTML]{D3E1E8}49.7 & \cellcolor[HTML]{B4D1ED}54.3 & \cellcolor[HTML]{F5F7F4}45.3 & \cellcolor[HTML]{B3D1EE}53.7 & \cellcolor[HTML]{82C0C4}54.5 & \cellcolor[HTML]{B7D2ED}52.3 & \cellcolor[HTML]{BED6EB}48.1 \\
 & English & \cellcolor[HTML]{5CBE78}\underline{\textbf{67.2}} & \cellcolor[HTML]{67BD90}\textbf{60.3} & \cellcolor[HTML]{61BE84}\textbf{60.2} & \cellcolor[HTML]{5DBE7A}\textbf{64.0} & \cellcolor[HTML]{60BE82}\textbf{66.6} & \cellcolor[HTML]{5FBE80}\textbf{63.0} & \cellcolor[HTML]{67BE8F}\textbf{53.4} & \cellcolor[HTML]{64BD8B}\textbf{56.0} & \cellcolor[HTML]{80C0C1}\textbf{47.6} & \cellcolor[HTML]{B5D2EE}\textbf{31.8} & \cellcolor[HTML]{E85C5C}5.7 & \cellcolor[HTML]{71BEA7}\textbf{64.2} & \cellcolor[HTML]{73BEA9}\textbf{56.3} & \cellcolor[HTML]{65BE8C}\textbf{61.6} & \cellcolor[HTML]{68BE93}\textbf{53.9} & \cellcolor[HTML]{5CBE78}\underline{\textbf{62.1}} & \cellcolor[HTML]{68BE93}\textbf{57.2} & \cellcolor[HTML]{78BEB2}\textbf{56.2} & \cellcolor[HTML]{60BE82}\textbf{54.8} \\
\midrule
\multirow{2}{*}{ja} & Native & \cellcolor[HTML]{EEF2EF}53.8 & \cellcolor[HTML]{AFCEF0}54.9 & \cellcolor[HTML]{E3EAEA}50.4 & \cellcolor[HTML]{EBEBD6}51.7 & \cellcolor[HTML]{C5D9EA}56.4 & \cellcolor[HTML]{BAD4EC}54.0 & \cellcolor[HTML]{7ABFB6}51.1 & \cellcolor[HTML]{9CC7E6}51.6 & \cellcolor[HTML]{D5E2E8}42.8 & \cellcolor[HTML]{F3C88D}25.9 & \cellcolor[HTML]{ECE9CD}6.7 & \cellcolor[HTML]{CCDDE9}55.9 & \cellcolor[HTML]{9DC7E7}52.5 & \cellcolor[HTML]{B2D0EE}54.4 & \cellcolor[HTML]{EBEBD8}44.1 & \cellcolor[HTML]{DDE7E8}51.5 & \cellcolor[HTML]{B0CFEF}50.7 & \cellcolor[HTML]{E8EEEB}50.4 & \cellcolor[HTML]{C6DAEA}47.7 \\
 & English & \cellcolor[HTML]{5EBE7E}\textbf{67.0} & \cellcolor[HTML]{7EC0BD}\textbf{58.2} & \cellcolor[HTML]{5FBE80}\textbf{60.4} & \cellcolor[HTML]{68BE93}\textbf{62.8} & \cellcolor[HTML]{5DBE79}\textbf{67.2} & \cellcolor[HTML]{5CBE78}\underline{\textbf{63.4}} & \cellcolor[HTML]{66BE8D}\textbf{53.5} & \cellcolor[HTML]{5CBE78}\underline{\textbf{56.7}} & \cellcolor[HTML]{73BEA9}\textbf{48.6} & \cellcolor[HTML]{7CBFBA}\textbf{34.7} & \cellcolor[HTML]{DDE7E9}\textbf{7.0} & \cellcolor[HTML]{62BE88}\textbf{66.1} & \cellcolor[HTML]{77BEB1}\textbf{55.9} & \cellcolor[HTML]{6EBE9F}\textbf{60.6} & \cellcolor[HTML]{65BE8C}\textbf{54.2} & \cellcolor[HTML]{5EBE7E}\textbf{61.8} & \cellcolor[HTML]{66BE8F}\textbf{57.4} & \cellcolor[HTML]{6ABE96}\textbf{57.3} & \cellcolor[HTML]{5DBD7B}\textbf{55.2} \\
\midrule
\multirow{2}{*}{th} & Native & \cellcolor[HTML]{EEF1E9}52.9 & \cellcolor[HTML]{EBEAD0}50.6 & \cellcolor[HTML]{F3E7AF}46.2 & \cellcolor[HTML]{F5DD9C}48.8 & \cellcolor[HTML]{ECEDE0}52.2 & \cellcolor[HTML]{F6E7A7}46.5 & \cellcolor[HTML]{EFF2EB}41.6 & \cellcolor[HTML]{78BFB3}54.4 & \cellcolor[HTML]{F2C58B}36.5 & \cellcolor[HTML]{F2BA85}25.4 & \cellcolor[HTML]{F6F8F5}6.9 & \cellcolor[HTML]{F6DF9E}48.1 & \cellcolor[HTML]{F6E09F}44.1 & \cellcolor[HTML]{EFF2EF}51.2 & \cellcolor[HTML]{ED906F}37.5 & \cellcolor[HTML]{EFA479}41.9 & \cellcolor[HTML]{F0B07F}40.0 & \cellcolor[HTML]{F3C88D}45.5 & \cellcolor[HTML]{F4E7AE}42.8 \\
 & English & \cellcolor[HTML]{63BE89}\textbf{66.2} & \cellcolor[HTML]{80C0C1}\textbf{58.1} & \cellcolor[HTML]{5CBE78}\underline{\textbf{60.8}} & \cellcolor[HTML]{60BD81}\textbf{63.7} & \cellcolor[HTML]{5CBE78}\underline{\textbf{67.2}} & \cellcolor[HTML]{64BD8C}\textbf{62.4} & \cellcolor[HTML]{5CBE78}\underline{\textbf{54.6}} & \cellcolor[HTML]{64BD8B}\textbf{56.0} & \cellcolor[HTML]{6BBD9A}\textbf{49.3} & \cellcolor[HTML]{90C3D7}\textbf{33.5} & \cellcolor[HTML]{EAEFED}\textbf{6.9} & \cellcolor[HTML]{7EC0BC}\textbf{62.7} & \cellcolor[HTML]{78BFB3}\textbf{55.8} & \cellcolor[HTML]{5CBE78}\underline{\textbf{62.6}} & \cellcolor[HTML]{6BBD9A}\textbf{53.6} & \cellcolor[HTML]{5DBE7A}\textbf{62.0} & \cellcolor[HTML]{61BE85}\textbf{57.9} & \cellcolor[HTML]{83C0C6}\textbf{55.3} & \cellcolor[HTML]{5FBD81}\textbf{54.9} \\
\midrule
\multirow{2}{*}{bn} & Native & \cellcolor[HTML]{CADCE9}56.1 & \cellcolor[HTML]{EA7164}43.6 & \cellcolor[HTML]{F5DB9B}44.9 & \cellcolor[HTML]{EE9C75}45.1 & \cellcolor[HTML]{ECEEE3}52.4 & \cellcolor[HTML]{F2C58B}43.7 & \cellcolor[HTML]{F6E0A0}37.3 & \cellcolor[HTML]{EA7465}39.8 & \cellcolor[HTML]{92C4DA}46.2 & \cellcolor[HTML]{ECEDDF}29.1 & \cellcolor[HTML]{62BE87}8.0 & \cellcolor[HTML]{ECE8C8}50.8 & \cellcolor[HTML]{F0B581}41.5 & \cellcolor[HTML]{F4D294}45.5 & \cellcolor[HTML]{E8645F}35.8 & \cellcolor[HTML]{F2BD86}43.3 & \cellcolor[HTML]{F5D898}42.4 & \cellcolor[HTML]{F5D495}46.1 & \cellcolor[HTML]{F5DC9B}41.8 \\
 & English & \cellcolor[HTML]{86C1C9}\textbf{61.9} & \cellcolor[HTML]{7CBFB9}\textbf{58.5} & \cellcolor[HTML]{66BE8F}\textbf{59.7} & \cellcolor[HTML]{86C1CB}\textbf{59.8} & \cellcolor[HTML]{6EBD9F}\textbf{64.9} & \cellcolor[HTML]{74BEAD}\textbf{60.4} & \cellcolor[HTML]{70BEA4}\textbf{52.2} & \cellcolor[HTML]{6BBE99}\textbf{55.4} & \cellcolor[HTML]{65BE8C}\textbf{49.9} & \cellcolor[HTML]{5CBE78}\underline{\textbf{36.9}} & \cellcolor[HTML]{5EBE7D}\textbf{8.1} & \cellcolor[HTML]{B7D2ED}\textbf{57.2} & \cellcolor[HTML]{71BEA5}\textbf{56.5} & \cellcolor[HTML]{8AC2CF}\textbf{57.7} & \cellcolor[HTML]{71BEA5}\textbf{53.1} & \cellcolor[HTML]{82C0C3}\textbf{57.9} & \cellcolor[HTML]{6FBEA3}\textbf{56.4} & \cellcolor[HTML]{E8EEEB}\textbf{50.4} & \cellcolor[HTML]{73BEA9}\textbf{53.2} \\
\midrule
\multirow{2}{*}{te} & Native & \cellcolor[HTML]{E85C5C}39.7 & \cellcolor[HTML]{EA7866}43.8 & \cellcolor[HTML]{E85C5C}38.2 & \cellcolor[HTML]{E85C5C}42.2 & \cellcolor[HTML]{E85C5C}39.5 & \cellcolor[HTML]{E85C5C}37.9 & \cellcolor[HTML]{E85C5C}29.5 & \cellcolor[HTML]{E85C5C}39.0 & \cellcolor[HTML]{F1B984}35.9 & \cellcolor[HTML]{F2C088}25.6 & \cellcolor[HTML]{5CBE78}\underline{\textbf{8.1}} & \cellcolor[HTML]{E85C5C}39.8 & \cellcolor[HTML]{E85C5C}37.6 & \cellcolor[HTML]{E85C5C}39.3 & \cellcolor[HTML]{ED906F}37.5 & \cellcolor[HTML]{E85C5C}38.4 & \cellcolor[HTML]{E85C5C}36.1 & \cellcolor[HTML]{E85C5C}41.5 & \cellcolor[HTML]{E85C5C}36.1 \\
 & English & \cellcolor[HTML]{84C1C7}\textbf{62.1} & \cellcolor[HTML]{8DC3D4}\textbf{57.1} & \cellcolor[HTML]{91C3D8}\textbf{55.4} & \cellcolor[HTML]{93C4DB}\textbf{58.6} & \cellcolor[HTML]{73BEA9}\textbf{64.2} & \cellcolor[HTML]{7EC0BC}\textbf{59.4} & \cellcolor[HTML]{6DBE9E}\textbf{52.6} & \cellcolor[HTML]{6FBEA3}\textbf{55.0} & \cellcolor[HTML]{78BEB2}\textbf{48.3} & \cellcolor[HTML]{5FBE7F}\textbf{36.7} & \cellcolor[HTML]{A1C8EA}7.3 & \cellcolor[HTML]{8CC2D3}\textbf{61.0} & \cellcolor[HTML]{96C5DF}\textbf{53.1} & \cellcolor[HTML]{7BBFB8}\textbf{59.2} & \cellcolor[HTML]{7ABFB5}\textbf{52.3} & \cellcolor[HTML]{94C4DD}\textbf{56.1} & \cellcolor[HTML]{8EC3D5}\textbf{53.4} & \cellcolor[HTML]{89C1CE}\textbf{54.9} & \cellcolor[HTML]{7ABEB5}\textbf{52.6} \\
\midrule
\multirow{2}{*}{sw} & Native & \cellcolor[HTML]{F7E7A6}48.9 & \cellcolor[HTML]{E85C5C}42.9 & \cellcolor[HTML]{EA6961}38.8 & \cellcolor[HTML]{F4D193}48.0 & \cellcolor[HTML]{EBEBD5}51.5 & \cellcolor[HTML]{F0AE7E}42.2 & \cellcolor[HTML]{F2C189}35.0 & \cellcolor[HTML]{EE9471}41.0 & \cellcolor[HTML]{E85C5C}32.3 & \cellcolor[HTML]{E85C5C}22.4 & \cellcolor[HTML]{F6DE9D}6.4 & \cellcolor[HTML]{F7E6A5}48.8 & \cellcolor[HTML]{E85F5D}37.7 & \cellcolor[HTML]{F6E09F}46.5 & \cellcolor[HTML]{E85C5C}35.5 & \cellcolor[HTML]{EBEAD4}48.6 & \cellcolor[HTML]{EDEFE4}46.6 & \cellcolor[HTML]{E8655F}41.8 & \cellcolor[HTML]{F1B682}39.7 \\
 & English & \cellcolor[HTML]{71BEA6}\textbf{64.5} & \cellcolor[HTML]{76BFAF}\textbf{58.9} & \cellcolor[HTML]{74BFAC}\textbf{58.2} & \cellcolor[HTML]{7DBFBB}\textbf{60.7} & \cellcolor[HTML]{6BBE98}\textbf{65.3} & \cellcolor[HTML]{77BFB1}\textbf{60.2} & \cellcolor[HTML]{6FBEA3}\textbf{52.3} & \cellcolor[HTML]{67BE92}\textbf{55.7} & \cellcolor[HTML]{7EC0BD}\textbf{47.7} & \cellcolor[HTML]{8AC2D0}\textbf{33.9} & \cellcolor[HTML]{77BFAF}\textbf{7.8} & \cellcolor[HTML]{83C0C6}\textbf{62.0} & \cellcolor[HTML]{93C4DB}\textbf{53.4} & \cellcolor[HTML]{6EBE9F}\textbf{60.6} & \cellcolor[HTML]{7EBFBD}\textbf{51.9} & \cellcolor[HTML]{6CBE9B}\textbf{60.3} & \cellcolor[HTML]{87C1CC}\textbf{54.0} & \cellcolor[HTML]{5CBE78}\underline{\textbf{58.4}} & \cellcolor[HTML]{6EBD9F}\textbf{53.6} \\
\midrule
\multirow{2}{*}{multi} & Native & \cellcolor[HTML]{7ABFB6}63.4 & \cellcolor[HTML]{7CBFBA}58.4 & \cellcolor[HTML]{B6D2ED}52.7 & \cellcolor[HTML]{9DC7E7}57.7 & \cellcolor[HTML]{85C1C8}61.9 & \cellcolor[HTML]{8CC2D2}57.9 & \cellcolor[HTML]{B1CFEF}45.8 & \cellcolor[HTML]{ACCDF1}50.7 & \cellcolor[HTML]{ECE9CA}39.8 & \cellcolor[HTML]{E8625E}22.6 & \cellcolor[HTML]{EDE8C4}6.6 & \cellcolor[HTML]{B2D0EF}57.5 & \cellcolor[HTML]{B4D1EE}51.2 & \cellcolor[HTML]{98C6E1}56.3 & \cellcolor[HTML]{CDDEE8}47.0 & \cellcolor[HTML]{A0C8EA}55.0 & \cellcolor[HTML]{AFCEF0}50.8 & \cellcolor[HTML]{69BD95}\textbf{57.4} & \cellcolor[HTML]{9FC8E9}49.6 \\
 & English & \cellcolor[HTML]{66BE8E}\textbf{66.0} & \cellcolor[HTML]{6DBE9D}\textbf{59.7} & \cellcolor[HTML]{67BD90}\textbf{59.6} & \cellcolor[HTML]{62BE88}\textbf{63.3} & \cellcolor[HTML]{61BE84}\textbf{66.5} & \cellcolor[HTML]{5DBE7B}\textbf{63.3} & \cellcolor[HTML]{5FBE7F}\textbf{54.2} & \cellcolor[HTML]{5CBE78}\textbf{56.7} & \cellcolor[HTML]{75BEAE}\textbf{48.4} & \cellcolor[HTML]{90C3D7}\textbf{33.5} & \cellcolor[HTML]{E0E8E9}\textbf{7.0} & \cellcolor[HTML]{5CBE78}\underline{\textbf{67.0}} & \cellcolor[HTML]{6FBEA3}\textbf{56.6} & \cellcolor[HTML]{66BE8D}\textbf{61.6} & \cellcolor[HTML]{5CBE78}\underline{\textbf{55.0}} & \cellcolor[HTML]{60BE83}\textbf{61.6} & \cellcolor[HTML]{5DBE7A}\textbf{58.4} & \cellcolor[HTML]{7CBFB9}55.9 & \cellcolor[HTML]{5CBE78}\textbf{55.2} \\
\bottomrule
\end{tabular}%
}
\end{minipage}
\hfill
\begin{minipage}[t]{0.49\textwidth}
\centering
\caption{Hard Math (PolyMath medium, high, and top) -- Avg@8 percentage-point improvement over the base model \texttt{Qwen3-1.7B-Base}}
\label{tab:appendix_crosslingual_qwen3_1_7b_base_8_appendix_hard_math_macro}
\tiny
\setlength{\tabcolsep}{2pt}
\renewcommand{\arraystretch}{0.92}
\newcommand{\MatrixResizeWidth}{\linewidth}
\resizebox{\MatrixResizeWidth}{!}{%
\begin{tabular}{llrrrrrrrrrrrrrrrrrrVr}
\toprule
Train & Reward & en & zh & de & fr & es & ru & ja & th & bn & te & sw & it & ko & pt & ar & id & ms & vi & Avg \\
\midrule
\multirow{1}{*}{en} & English & \cellcolor[HTML]{7DBFBC}\textbf{7.4} & \cellcolor[HTML]{83C1C5}\textbf{7.0} & \cellcolor[HTML]{9AC6E4}\textbf{6.7} & \cellcolor[HTML]{9CC7E6}\textbf{6.4} & \cellcolor[HTML]{7EC0BD}\textbf{7.0} & \cellcolor[HTML]{75BEAD}\textbf{6.9} & \cellcolor[HTML]{A6CAEF}\textbf{6.5} & \cellcolor[HTML]{8DC2D3}\textbf{6.4} & \cellcolor[HTML]{90C3D7}\textbf{6.6} & \cellcolor[HTML]{64BD8C}\textbf{6.1} & \cellcolor[HTML]{B0CFEF}\textbf{3.7} & \cellcolor[HTML]{79BFB4}\textbf{7.2} & \cellcolor[HTML]{A5CAEE}\textbf{7.1} & \cellcolor[HTML]{9EC8E8}\textbf{6.9} & \cellcolor[HTML]{A5CAEE}\textbf{6.6} & \cellcolor[HTML]{6EBEA0}\textbf{7.2} & \cellcolor[HTML]{9CC7E6}\textbf{6.3} & \cellcolor[HTML]{77BEB1}\textbf{7.1} & \cellcolor[HTML]{85C1C9}\textbf{6.6} \\
\midrule
\multirow{2}{*}{zh} & Native & \cellcolor[HTML]{A7CBF0}6.5 & \cellcolor[HTML]{68BE93}\textbf{7.5} & \cellcolor[HTML]{BFD6EB}6.3 & \cellcolor[HTML]{9EC8E7}6.4 & \cellcolor[HTML]{7EC0BD}\textbf{7.0} & \cellcolor[HTML]{8BC2D1}6.5 & \cellcolor[HTML]{8EC3D4}\textbf{6.9} & \cellcolor[HTML]{C0D7EA}5.8 & \cellcolor[HTML]{EBEAD3}5.3 & \cellcolor[HTML]{F4F6F3}4.6 & \cellcolor[HTML]{63BE8B}4.2 & \cellcolor[HTML]{A0C8EA}6.4 & \cellcolor[HTML]{EDEEE3}6.0 & \cellcolor[HTML]{91C3D8}7.1 & \cellcolor[HTML]{87C1CC}7.1 & \cellcolor[HTML]{7ABFB7}7.0 & \cellcolor[HTML]{A4CAED}6.2 & \cellcolor[HTML]{96C5E0}6.5 & \cellcolor[HTML]{99C6E2}6.3 \\
 & English & \cellcolor[HTML]{5CBE78}\underline{\textbf{8.2}} & \cellcolor[HTML]{6BBD9A}7.5 & \cellcolor[HTML]{63BE89}\textbf{7.9} & \cellcolor[HTML]{6DBE9D}\textbf{7.4} & \cellcolor[HTML]{A2C9EB}6.3 & \cellcolor[HTML]{68BE92}\textbf{7.1} & \cellcolor[HTML]{C5DAEA}6.1 & \cellcolor[HTML]{91C3D8}\textbf{6.3} & \cellcolor[HTML]{6EBEA0}\textbf{7.2} & \cellcolor[HTML]{8BC2D1}\textbf{5.5} & \cellcolor[HTML]{5CBE78}\underline{\textbf{4.3}} & \cellcolor[HTML]{81C0C2}\textbf{7.0} & \cellcolor[HTML]{B5D2EE}\textbf{6.9} & \cellcolor[HTML]{77BFAF}\textbf{7.6} & \cellcolor[HTML]{68BE93}\textbf{7.6} & \cellcolor[HTML]{73BEAA}\textbf{7.1} & \cellcolor[HTML]{87C1CC}\textbf{6.7} & \cellcolor[HTML]{93C4DB}\textbf{6.5} & \cellcolor[HTML]{78BEB2}\textbf{6.8} \\
\midrule
\multirow{2}{*}{de} & Native & \cellcolor[HTML]{BCD5EC}6.3 & \cellcolor[HTML]{ECEDE0}5.3 & \cellcolor[HTML]{EDEFE4}5.5 & \cellcolor[HTML]{EDE8C4}5.0 & \cellcolor[HTML]{F0E7B9}4.9 & \cellcolor[HTML]{F3E7B2}4.7 & \cellcolor[HTML]{EBEAD1}5.2 & \cellcolor[HTML]{EDE8C3}4.9 & \cellcolor[HTML]{F3E7AE}5.0 & \cellcolor[HTML]{EEE7BF}4.2 & \cellcolor[HTML]{F6F8F5}3.4 & \cellcolor[HTML]{DFE8E9}5.8 & \cellcolor[HTML]{F6DF9E}5.2 & \cellcolor[HTML]{EFE7BD}5.5 & \cellcolor[HTML]{BFD7EB}6.4 & \cellcolor[HTML]{EFF2EB}5.4 & \cellcolor[HTML]{D6E3E8}5.7 & \cellcolor[HTML]{EBEAD3}5.1 & \cellcolor[HTML]{EBEBD7}5.2 \\
 & English & \cellcolor[HTML]{B0CFEF}\textbf{6.4} & \cellcolor[HTML]{71BEA7}\textbf{7.3} & \cellcolor[HTML]{88C1CD}\textbf{7.1} & \cellcolor[HTML]{86C1CB}\textbf{6.8} & \cellcolor[HTML]{67BD90}\textbf{7.5} & \cellcolor[HTML]{66BE8F}\textbf{7.2} & \cellcolor[HTML]{69BD94}\textbf{7.7} & \cellcolor[HTML]{86C1CA}\textbf{6.5} & \cellcolor[HTML]{8EC2D5}\textbf{6.6} & \cellcolor[HTML]{7DBFBB}\textbf{5.7} & \cellcolor[HTML]{B9D3ED}\textbf{3.6} & \cellcolor[HTML]{69BE94}\textbf{7.5} & \cellcolor[HTML]{D4E2E8}\textbf{6.6} & \cellcolor[HTML]{6ABD96}\textbf{7.9} & \cellcolor[HTML]{80C0C0}\textbf{7.2} & \cellcolor[HTML]{7CBFBA}\textbf{6.9} & \cellcolor[HTML]{98C6E1}\textbf{6.4} & \cellcolor[HTML]{7FC0C0}\textbf{6.9} & \cellcolor[HTML]{7CBFB9}\textbf{6.8} \\
\midrule
\multirow{2}{*}{fr} & Native & \cellcolor[HTML]{E1EAE9}5.8 & \cellcolor[HTML]{D3E1E8}5.9 & \cellcolor[HTML]{EBEBD8}5.4 & \cellcolor[HTML]{D6E3E8}\textbf{5.8} & \cellcolor[HTML]{B5D2EE}6.1 & \cellcolor[HTML]{E6ECEB}5.4 & \cellcolor[HTML]{EEF2EE}\textbf{5.7} & \cellcolor[HTML]{EBEBD6}5.1 & \cellcolor[HTML]{F5DD9C}4.8 & \cellcolor[HTML]{BFD7EB}5.0 & \cellcolor[HTML]{98C6E1}\textbf{3.8} & \cellcolor[HTML]{EEE7BE}5.0 & \cellcolor[HTML]{F6E6A8}5.4 & \cellcolor[HTML]{CFDFE8}6.4 & \cellcolor[HTML]{EEF2EF}5.9 & \cellcolor[HTML]{ADCDF0}6.1 & \cellcolor[HTML]{EEE8C3}4.9 & \cellcolor[HTML]{8DC3D4}\textbf{6.6} & \cellcolor[HTML]{EAEFEC}5.5 \\
 & English & \cellcolor[HTML]{ACCDF0}\textbf{6.4} & \cellcolor[HTML]{D0DFE8}\textbf{5.9} & \cellcolor[HTML]{8FC3D6}\textbf{7.0} & \cellcolor[HTML]{D9E5E8}5.7 & \cellcolor[HTML]{A2C9EB}\textbf{6.3} & \cellcolor[HTML]{7ABFB7}\textbf{6.8} & \cellcolor[HTML]{F0F2EC}5.5 & \cellcolor[HTML]{AACCF1}\textbf{6.0} & \cellcolor[HTML]{F0F3F0}\textbf{5.6} & \cellcolor[HTML]{B1CFEF}\textbf{5.1} & \cellcolor[HTML]{EBF0ED}3.4 & \cellcolor[HTML]{9CC7E6}\textbf{6.5} & \cellcolor[HTML]{DDE7E9}\textbf{6.5} & \cellcolor[HTML]{A4CAED}\textbf{6.8} & \cellcolor[HTML]{8AC2CF}\textbf{7.0} & \cellcolor[HTML]{8FC3D6}\textbf{6.6} & \cellcolor[HTML]{DDE7E9}\textbf{5.6} & \cellcolor[HTML]{A8CBF1}6.2 & \cellcolor[HTML]{ABCDF1}\textbf{6.1} \\
\midrule
\multirow{2}{*}{es} & Native & \cellcolor[HTML]{A0C8EA}6.6 & \cellcolor[HTML]{83C1C5}\textbf{7.0} & \cellcolor[HTML]{BBD5EC}6.3 & \cellcolor[HTML]{96C5E0}6.5 & \cellcolor[HTML]{81C0C3}6.9 & \cellcolor[HTML]{9FC8E8}6.1 & \cellcolor[HTML]{A1C8EA}\textbf{6.6} & \cellcolor[HTML]{CBDDE8}5.7 & \cellcolor[HTML]{A9CCF1}6.2 & \cellcolor[HTML]{75BEAF}\textbf{5.8} & \cellcolor[HTML]{6CBE9B}\textbf{4.1} & \cellcolor[HTML]{BED6EB}6.1 & \cellcolor[HTML]{EBEBD7}5.9 & \cellcolor[HTML]{82C0C4}7.4 & \cellcolor[HTML]{BFD7EB}6.4 & \cellcolor[HTML]{9FC8E8}6.3 & \cellcolor[HTML]{F2F4EF}5.3 & \cellcolor[HTML]{95C4DD}6.5 & \cellcolor[HTML]{9FC8E8}6.2 \\
 & English & \cellcolor[HTML]{7BBFB7}\textbf{7.4} & \cellcolor[HTML]{A2C9EB}6.4 & \cellcolor[HTML]{5CBE78}\underline{\textbf{8.1}} & \cellcolor[HTML]{5CBE78}\underline{\textbf{7.7}} & \cellcolor[HTML]{6EBEA0}\textbf{7.3} & \cellcolor[HTML]{83C1C5}\textbf{6.6} & \cellcolor[HTML]{A6CAEF}6.5 & \cellcolor[HTML]{86C1CA}\textbf{6.5} & \cellcolor[HTML]{90C3D7}\textbf{6.6} & \cellcolor[HTML]{78BFB3}5.8 & \cellcolor[HTML]{85C1C9}3.9 & \cellcolor[HTML]{75BEAE}\textbf{7.2} & \cellcolor[HTML]{95C5DE}\textbf{7.4} & \cellcolor[HTML]{70BEA3}\textbf{7.8} & \cellcolor[HTML]{6ABE97}\textbf{7.6} & \cellcolor[HTML]{5FBE7F}\textbf{7.5} & \cellcolor[HTML]{8BC2D1}\textbf{6.6} & \cellcolor[HTML]{75BEAE}\textbf{7.1} & \cellcolor[HTML]{74BEAD}\textbf{6.9} \\
\midrule
\multirow{2}{*}{ru} & Native & \cellcolor[HTML]{D9E5E8}5.9 & \cellcolor[HTML]{C6DAE9}6.0 & \cellcolor[HTML]{B8D3ED}6.3 & \cellcolor[HTML]{F3F5F2}5.5 & \cellcolor[HTML]{83C1C5}6.9 & \cellcolor[HTML]{B5D2EE}5.9 & \cellcolor[HTML]{DFE8E9}5.8 & \cellcolor[HTML]{D7E3E8}5.6 & \cellcolor[HTML]{C7DAE9}6.0 & \cellcolor[HTML]{7FBFBF}5.7 & \cellcolor[HTML]{EBF0ED}3.4 & \cellcolor[HTML]{C5DAEA}6.0 & \cellcolor[HTML]{EBE9CF}5.8 & \cellcolor[HTML]{BAD4EC}6.6 & \cellcolor[HTML]{C3D8EA}6.3 & \cellcolor[HTML]{96C5DF}6.5 & \cellcolor[HTML]{C5DAE9}5.8 & \cellcolor[HTML]{DCE6E8}5.7 & \cellcolor[HTML]{C1D7EA}5.9 \\
 & English & \cellcolor[HTML]{73BEAA}\textbf{7.6} & \cellcolor[HTML]{5CBE78}\underline{\textbf{7.8}} & \cellcolor[HTML]{6DBD9F}\textbf{7.7} & \cellcolor[HTML]{80C0C0}\textbf{7.0} & \cellcolor[HTML]{60BE82}\textbf{7.6} & \cellcolor[HTML]{5CBE78}\underline{\textbf{7.4}} & \cellcolor[HTML]{7EC0BC}\textbf{7.3} & \cellcolor[HTML]{6BBD9A}\textbf{7.0} & \cellcolor[HTML]{5CBE78}\underline{\textbf{7.5}} & \cellcolor[HTML]{6BBE9A}\textbf{6.0} & \cellcolor[HTML]{5CBE78}\underline{\textbf{4.3}} & \cellcolor[HTML]{8FC3D6}\textbf{6.7} & \cellcolor[HTML]{9DC7E7}\textbf{7.2} & \cellcolor[HTML]{68BE93}\textbf{7.9} & \cellcolor[HTML]{94C4DC}\textbf{6.9} & \cellcolor[HTML]{5FBE7F}\textbf{7.5} & \cellcolor[HTML]{7DBFBB}\textbf{6.9} & \cellcolor[HTML]{6FBEA1}\textbf{7.2} & \cellcolor[HTML]{6BBE98}\textbf{7.1} \\
\midrule
\multirow{2}{*}{ja} & Native & \cellcolor[HTML]{F3F6F2}5.6 & \cellcolor[HTML]{EBE9D1}5.2 & \cellcolor[HTML]{EBEAD2}5.3 & \cellcolor[HTML]{EDE8C7}5.0 & \cellcolor[HTML]{ECEDE0}5.3 & \cellcolor[HTML]{EBEAD3}5.0 & \cellcolor[HTML]{EBEBD7}5.3 & \cellcolor[HTML]{F5DC9C}4.6 & \cellcolor[HTML]{F2C289}4.5 & \cellcolor[HTML]{EEF0E8}4.5 & \cellcolor[HTML]{F7E5A4}3.1 & \cellcolor[HTML]{EDEEE3}5.4 & \cellcolor[HTML]{F1B884}4.7 & \cellcolor[HTML]{F0F3F0}6.0 & \cellcolor[HTML]{EBEAD3}5.6 & \cellcolor[HTML]{ECEEE0}5.3 & \cellcolor[HTML]{F6F8F5}5.4 & \cellcolor[HTML]{D8E4E8}5.7 & \cellcolor[HTML]{ECE8CA}5.1 \\
 & English & \cellcolor[HTML]{5FBE7E}\textbf{8.1} & \cellcolor[HTML]{70BEA3}\textbf{7.4} & \cellcolor[HTML]{6BBE98}\textbf{7.7} & \cellcolor[HTML]{67BD90}\textbf{7.5} & \cellcolor[HTML]{5CBE78}\underline{\textbf{7.7}} & \cellcolor[HTML]{5CBE78}\underline{\textbf{7.4}} & \cellcolor[HTML]{71BEA7}\textbf{7.5} & \cellcolor[HTML]{66BE8F}\textbf{7.1} & \cellcolor[HTML]{74BEAC}\textbf{7.1} & \cellcolor[HTML]{5CBE78}\underline{\textbf{6.2}} & \cellcolor[HTML]{70BEA4}\textbf{4.1} & \cellcolor[HTML]{5CBE78}\underline{\textbf{7.8}} & \cellcolor[HTML]{5CBE78}\underline{\textbf{8.7}} & \cellcolor[HTML]{7CBFB9}\textbf{7.5} & \cellcolor[HTML]{62BE88}\textbf{7.7} & \cellcolor[HTML]{5CBE78}\underline{\textbf{7.6}} & \cellcolor[HTML]{5CBE78}\underline{\textbf{7.5}} & \cellcolor[HTML]{5CBE78}\underline{\textbf{7.6}} & \cellcolor[HTML]{5CBE78}\underline{\textbf{7.3}} \\
\midrule
\multirow{2}{*}{th} & Native & \cellcolor[HTML]{ECE9CB}5.2 & \cellcolor[HTML]{EDE8C7}5.1 & \cellcolor[HTML]{ECEDDE}5.5 & \cellcolor[HTML]{F2E7B2}4.8 & \cellcolor[HTML]{F4E6AE}4.8 & \cellcolor[HTML]{EFF2EB}5.2 & \cellcolor[HTML]{F7E5A5}4.8 & \cellcolor[HTML]{EBEAD3}5.1 & \cellcolor[HTML]{EBEAD3}5.3 & \cellcolor[HTML]{EBEAD2}4.4 & \cellcolor[HTML]{D3E1E8}3.5 & \cellcolor[HTML]{F4D193}4.5 & \cellcolor[HTML]{F1B682}4.7 & \cellcolor[HTML]{F4D092}4.9 & \cellcolor[HTML]{F3C78D}4.8 & \cellcolor[HTML]{EEF0E7}5.4 & \cellcolor[HTML]{F3CB90}4.3 & \cellcolor[HTML]{F5D999}4.5 & \cellcolor[HTML]{F6E7A8}4.8 \\
 & English & \cellcolor[HTML]{6EBD9F}\textbf{7.7} & \cellcolor[HTML]{91C3D9}\textbf{6.7} & \cellcolor[HTML]{8AC2CF}\textbf{7.1} & \cellcolor[HTML]{7BBFB8}\textbf{7.1} & \cellcolor[HTML]{80C0C0}\textbf{7.0} & \cellcolor[HTML]{8DC2D3}\textbf{6.4} & \cellcolor[HTML]{5CBE78}\underline{\textbf{8.0}} & \cellcolor[HTML]{73BEA9}\textbf{6.8} & \cellcolor[HTML]{90C3D7}\textbf{6.6} & \cellcolor[HTML]{62BD87}\textbf{6.1} & \cellcolor[HTML]{74BFAC}\textbf{4.1} & \cellcolor[HTML]{6DBE9E}\textbf{7.4} & \cellcolor[HTML]{A3C9EC}\textbf{7.1} & \cellcolor[HTML]{9CC7E6}\textbf{6.9} & \cellcolor[HTML]{5CBE78}\underline{\textbf{7.9}} & \cellcolor[HTML]{7CBFBA}\textbf{6.9} & \cellcolor[HTML]{96C5DF}\textbf{6.4} & \cellcolor[HTML]{86C1CA}\textbf{6.8} & \cellcolor[HTML]{78BFB3}\textbf{6.8} \\
\midrule
\multirow{2}{*}{bn} & Native & \cellcolor[HTML]{F0AE7E}3.9 & \cellcolor[HTML]{E85C5C}3.2 & \cellcolor[HTML]{EEA277}4.0 & \cellcolor[HTML]{E85C5C}3.2 & \cellcolor[HTML]{F0AA7C}3.9 & \cellcolor[HTML]{F0B481}3.9 & \cellcolor[HTML]{EC836B}3.5 & \cellcolor[HTML]{E8615E}3.5 & \cellcolor[HTML]{EEA277}4.2 & \cellcolor[HTML]{F4D193}3.8 & \cellcolor[HTML]{F0A77A}2.8 & \cellcolor[HTML]{F2C189}4.3 & \cellcolor[HTML]{E85F5D}3.8 & \cellcolor[HTML]{EB7C68}4.1 & \cellcolor[HTML]{E85C5C}3.9 & \cellcolor[HTML]{EA7164}3.5 & \cellcolor[HTML]{E8645F}3.3 & \cellcolor[HTML]{E85C5C}3.2 & \cellcolor[HTML]{EA7264}3.7 \\
 & English & \cellcolor[HTML]{ECF1EE}\textbf{5.7} & \cellcolor[HTML]{C9DCE9}\textbf{6.0} & \cellcolor[HTML]{DDE7E9}\textbf{5.9} & \cellcolor[HTML]{AACCF1}\textbf{6.2} & \cellcolor[HTML]{E5ECEA}\textbf{5.6} & \cellcolor[HTML]{A7CBEF}\textbf{6.0} & \cellcolor[HTML]{CBDDE9}\textbf{6.1} & \cellcolor[HTML]{EBEAD3}\textbf{5.1} & \cellcolor[HTML]{D9E5E8}\textbf{5.8} & \cellcolor[HTML]{C9DCE9}\textbf{4.9} & \cellcolor[HTML]{B0CFEF}\textbf{3.7} & \cellcolor[HTML]{A4CAED}\textbf{6.4} & \cellcolor[HTML]{E2EAE9}\textbf{6.4} & \cellcolor[HTML]{CCDDE8}\textbf{6.4} & \cellcolor[HTML]{A3C9EC}\textbf{6.6} & \cellcolor[HTML]{CBDDE9}\textbf{5.9} & \cellcolor[HTML]{D9E5E8}\textbf{5.6} & \cellcolor[HTML]{AFCFEF}\textbf{6.1} & \cellcolor[HTML]{CADDE9}\textbf{5.8} \\
\midrule
\multirow{2}{*}{te} & Native & \cellcolor[HTML]{E85C5C}3.1 & \cellcolor[HTML]{EA6860}3.3 & \cellcolor[HTML]{E85C5C}3.3 & \cellcolor[HTML]{EFA479}3.8 & \cellcolor[HTML]{E85C5C}3.2 & \cellcolor[HTML]{E85C5C}3.2 & \cellcolor[HTML]{EE9D75}3.8 & \cellcolor[HTML]{E85C5C}3.4 & \cellcolor[HTML]{F1B783}4.4 & \cellcolor[HTML]{EC7F69}3.2 & \cellcolor[HTML]{F5DA9A}3.0 & \cellcolor[HTML]{E85C5C}3.3 & \cellcolor[HTML]{E85C5C}3.7 & \cellcolor[HTML]{E8645F}3.9 & \cellcolor[HTML]{EE9B74}4.4 & \cellcolor[HTML]{E85C5C}3.4 & \cellcolor[HTML]{E85C5C}3.2 & \cellcolor[HTML]{EC7F69}3.5 & \cellcolor[HTML]{E85C5C}3.5 \\
 & English & \cellcolor[HTML]{D9E5E8}\textbf{5.9} & \cellcolor[HTML]{74BEAD}\textbf{7.3} & \cellcolor[HTML]{F3F5F2}\textbf{5.7} & \cellcolor[HTML]{9EC8E7}\textbf{6.4} & \cellcolor[HTML]{D6E3E8}\textbf{5.8} & \cellcolor[HTML]{94C4DD}\textbf{6.3} & \cellcolor[HTML]{B4D1ED}\textbf{6.3} & \cellcolor[HTML]{AFCFEF}\textbf{5.9} & \cellcolor[HTML]{C3D8EA}\textbf{6.0} & \cellcolor[HTML]{CDDEE9}\textbf{4.9} & \cellcolor[HTML]{EBEBD9}\textbf{3.3} & \cellcolor[HTML]{81C0C2}\textbf{7.0} & \cellcolor[HTML]{EBECDD}\textbf{6.0} & \cellcolor[HTML]{F2F5F1}\textbf{6.0} & \cellcolor[HTML]{BBD4EC}\textbf{6.4} & \cellcolor[HTML]{CBDDE9}\textbf{5.9} & \cellcolor[HTML]{F2F4EF}\textbf{5.3} & \cellcolor[HTML]{A6CBEF}\textbf{6.2} & \cellcolor[HTML]{BCD5EC}\textbf{5.9} \\
\midrule
\multirow{2}{*}{sw} & Native & \cellcolor[HTML]{F0AE7E}3.9 & \cellcolor[HTML]{EEA177}3.9 & \cellcolor[HTML]{EE9571}3.8 & \cellcolor[HTML]{EE9E76}3.8 & \cellcolor[HTML]{EC826A}3.5 & \cellcolor[HTML]{F0B07F}3.9 & \cellcolor[HTML]{E85C5C}3.2 & \cellcolor[HTML]{EB7C68}3.7 & \cellcolor[HTML]{E85C5C}3.7 & \cellcolor[HTML]{F0B280}3.6 & \cellcolor[HTML]{EA7163}2.6 & \cellcolor[HTML]{E8605D}3.4 & \cellcolor[HTML]{EA6E63}3.9 & \cellcolor[HTML]{E85C5C}3.8 & \cellcolor[HTML]{EA7766}4.1 & \cellcolor[HTML]{F4D495}4.5 & \cellcolor[HTML]{EC846B}3.5 & \cellcolor[HTML]{EA7465}3.4 & \cellcolor[HTML]{EA7264}3.7 \\
 & English & \cellcolor[HTML]{C2D8EA}\textbf{6.2} & \cellcolor[HTML]{B9D3ED}\textbf{6.1} & \cellcolor[HTML]{9CC7E6}\textbf{6.7} & \cellcolor[HTML]{D6E3E8}\textbf{5.8} & \cellcolor[HTML]{D6E3E8}\textbf{5.8} & \cellcolor[HTML]{88C1CE}\textbf{6.5} & \cellcolor[HTML]{CEDEE8}\textbf{6.0} & \cellcolor[HTML]{EBEBD6}\textbf{5.1} & \cellcolor[HTML]{A4CAED}\textbf{6.3} & \cellcolor[HTML]{A0C8E9}\textbf{5.3} & \cellcolor[HTML]{5CBE78}\underline{\textbf{4.3}} & \cellcolor[HTML]{C1D8EA}\textbf{6.1} & \cellcolor[HTML]{ECE8C8}\textbf{5.7} & \cellcolor[HTML]{E2EAEA}\textbf{6.2} & \cellcolor[HTML]{AFCFEF}\textbf{6.5} & \cellcolor[HTML]{C7DBE9}\textbf{5.9} & \cellcolor[HTML]{CFDFE8}\textbf{5.7} & \cellcolor[HTML]{8CC2D2}\textbf{6.7} & \cellcolor[HTML]{BAD4EC}\textbf{5.9} \\
\midrule
\multirow{2}{*}{multi} & Native & \cellcolor[HTML]{6FBEA2}\textbf{7.7} & \cellcolor[HTML]{EDEFE5}5.4 & \cellcolor[HTML]{EDEFE4}5.5 & \cellcolor[HTML]{EFE7BC}4.9 & \cellcolor[HTML]{ACCDF0}6.2 & \cellcolor[HTML]{DFE8E9}5.5 & \cellcolor[HTML]{F6E2A1}4.7 & \cellcolor[HTML]{F3CB8F}4.4 & \cellcolor[HTML]{F0A77A}4.3 & \cellcolor[HTML]{E85C5C}3.0 & \cellcolor[HTML]{E85C5C}2.5 & \cellcolor[HTML]{E7EDEB}5.7 & \cellcolor[HTML]{F2BD86}4.8 & \cellcolor[HTML]{F4F6F3}6.0 & \cellcolor[HTML]{F3E7B2}5.3 & \cellcolor[HTML]{D5E2E8}5.8 & \cellcolor[HTML]{F7E3A2}4.6 & \cellcolor[HTML]{E5ECEA}5.6 & \cellcolor[HTML]{ECE9CC}5.1 \\
 & English & \cellcolor[HTML]{91C3D9}6.9 & \cellcolor[HTML]{60BE82}\textbf{7.7} & \cellcolor[HTML]{8BC2D0}\textbf{7.0} & \cellcolor[HTML]{96C5E0}\textbf{6.5} & \cellcolor[HTML]{76BFAF}\textbf{7.2} & \cellcolor[HTML]{64BD8B}\textbf{7.2} & \cellcolor[HTML]{85C1C8}\textbf{7.1} & \cellcolor[HTML]{5CBE78}\underline{\textbf{7.2}} & \cellcolor[HTML]{E6ECEB}\textbf{5.7} & \cellcolor[HTML]{D6E3E8}\textbf{4.8} & \cellcolor[HTML]{EDEFE5}\textbf{3.4} & \cellcolor[HTML]{6FBEA3}\textbf{7.3} & \cellcolor[HTML]{E5ECEA}\textbf{6.4} & \cellcolor[HTML]{5CBE78}\underline{\textbf{8.2}} & \cellcolor[HTML]{9BC7E5}\textbf{6.8} & \cellcolor[HTML]{83C1C5}\textbf{6.8} & \cellcolor[HTML]{A2C9EC}\textbf{6.2} & \cellcolor[HTML]{8BC2D1}\textbf{6.7} & \cellcolor[HTML]{85C1C7}\textbf{6.6} \\
\bottomrule
\end{tabular}%
}
\end{minipage}
\end{table}

\FloatBarrier
\subsection{Qwen3-4B-Base}
\label{app:matrices-qwen3-4b-base-sec}
\paragraph{Ineffective rewards.}
For Qwen3-4B-Base, we mark some reward rows as ineffective. ``Ineffective'' means the base model had at least 2.5\% support for the target reasoning language, but the trained model did not converge to that language.
\begin{itemize}
\item For training in Chinese (zh), English reward is marked \textit{ineffective}. On Chinese (zh)-prompt rollouts, the base model produced 51.1\% Chinese (zh), 24.0\% English, and 24.9\% other languages; the trained model produced 90.2\% Chinese (zh), 2.8\% English, and 7.0\% other languages.
\item For training in Japanese (ja), English reward is marked \textit{ineffective}. On Japanese (ja)-prompt rollouts, the base model produced 49.5\% Japanese (ja), 23.8\% English, and 26.7\% other languages; the trained model produced 96.4\% Japanese (ja), 1.7\% English, and 1.8\% other languages.
\end{itemize}

\begin{table}[!htbp]
\centering
\begin{minipage}[t]{0.49\textwidth}
\centering
\caption{Tasks seen during training from Multilingual Reasoning Gym -- Avg@8 percentage-point improvement over the base model \texttt{Qwen3-4B-Base}}
\label{tab:appendix_crosslingual_qwen3_4b_base_1_train_filtered_noheldout}
\tiny
\setlength{\tabcolsep}{2pt}
\renewcommand{\arraystretch}{0.92}
\newcommand{\MatrixResizeWidth}{\linewidth}
\resizebox{\MatrixResizeWidth}{!}{%
\begin{tabular}{llrrrrrrrrrrrrrrVr}
\toprule
Train & Reward & en & zh & de & fr & es & ru & ja & th & bn & te & sw & it & ko & pt & Avg \\
\midrule
\multirow{1}{*}{en} & English & \cellcolor[HTML]{5CBE78}\underline{\textbf{50.8}} & \cellcolor[HTML]{95C5DE}\textbf{41.3} & \cellcolor[HTML]{8DC2D3}\textbf{48.4} & \cellcolor[HTML]{B9D4EC}\textbf{44.0} & \cellcolor[HTML]{ADCDF0}\textbf{44.5} & \cellcolor[HTML]{9DC7E6}\textbf{42.9} & \cellcolor[HTML]{B1D0EF}\textbf{41.9} & \cellcolor[HTML]{C2D8EA}\textbf{46.0} & \cellcolor[HTML]{EBEBD8}\textbf{43.8} & \cellcolor[HTML]{F6E7A7}\textbf{39.7} & \cellcolor[HTML]{F0AE7E}\textbf{31.6} & \cellcolor[HTML]{D5E2E8}\textbf{43.3} & \cellcolor[HTML]{A0C8EA}\textbf{43.9} & \cellcolor[HTML]{A6CBEF}\textbf{43.4} & \cellcolor[HTML]{98C6E1}\textbf{43.2} \\
\midrule
\multirow{2}{*}{zh} & Native & \cellcolor[HTML]{ABCDF1}45.1 & \cellcolor[HTML]{60BE82}45.5 & \cellcolor[HTML]{F4F6F2}43.3 & \cellcolor[HTML]{EFE7BD}39.5 & \cellcolor[HTML]{F6E7A7}38.4 & \cellcolor[HTML]{ECEDE0}38.2 & \cellcolor[HTML]{9AC6E4}43.0 & \cellcolor[HTML]{EBEAD5}42.7 & \cellcolor[HTML]{F0B581}\textbf{39.7} & \cellcolor[HTML]{F1B783}\textbf{37.0} & \cellcolor[HTML]{F0A67A}\textbf{31.1} & \cellcolor[HTML]{EFE7BC}40.3 & \cellcolor[HTML]{E4EBEA}41.4 & \cellcolor[HTML]{ECEEE2}39.7 & \cellcolor[HTML]{F3F5F2}40.4 \\
 & English (ineffective) & \cellcolor[HTML]{9BC7E5}\textbf{46.0} & \cellcolor[HTML]{5CBE78}\underline{\textbf{45.9}} & \cellcolor[HTML]{E2EAE9}\textbf{44.1} & \cellcolor[HTML]{ECEEE1}\textbf{41.0} & \cellcolor[HTML]{EDE8C4}\textbf{39.6} & \cellcolor[HTML]{ECF0ED}\textbf{39.3} & \cellcolor[HTML]{8AC1CF}\textbf{44.1} & \cellcolor[HTML]{EEF2EE}\textbf{44.1} & \cellcolor[HTML]{F0AA7C}39.3 & \cellcolor[HTML]{F0B581}36.9 & \cellcolor[HTML]{EC8D6E}29.8 & \cellcolor[HTML]{EBEBD9}\textbf{41.3} & \cellcolor[HTML]{D3E1E8}\textbf{42.0} & \cellcolor[HTML]{F0F3F0}\textbf{40.5} & \cellcolor[HTML]{DFE8E9}\textbf{41.0} \\
\bottomrule
\end{tabular}%
}
\end{minipage}
\hfill
\begin{minipage}[t]{0.49\textwidth}
\centering
\caption{Macro average over all unseen tasks -- Avg@8 percentage-point improvement over the base model \texttt{Qwen3-4B-Base}}
\label{tab:appendix_crosslingual_qwen3_4b_base_2_appendix_unseen_task_groups_macro}
\tiny
\setlength{\tabcolsep}{2pt}
\renewcommand{\arraystretch}{0.92}
\newcommand{\MatrixResizeWidth}{\linewidth}
\resizebox{\MatrixResizeWidth}{!}{%
\begin{tabular}{llrrrrrrrrrrrVr}
\toprule
Train & Reward & en & zh & de & fr & es & ru & ja & th & bn & te & sw & Avg \\
\midrule
\multirow{1}{*}{en} & English & \cellcolor[HTML]{7CBFB9}\textbf{32.8} & \cellcolor[HTML]{73BEA9}\textbf{29.1} & \cellcolor[HTML]{72BEA8}\textbf{32.7} & \cellcolor[HTML]{74BEAC}\textbf{32.4} & \cellcolor[HTML]{6CBD9B}\textbf{31.6} & \cellcolor[HTML]{78BEB2}\textbf{32.3} & \cellcolor[HTML]{7EC0BD}\textbf{32.5} & \cellcolor[HTML]{79BFB3}\textbf{31.6} & \cellcolor[HTML]{94C4DD}\textbf{32.9} & \cellcolor[HTML]{A4CAED}\textbf{26.2} & \cellcolor[HTML]{CADCE9}\textbf{13.8} & \cellcolor[HTML]{7DBFBB}\textbf{29.8} \\
\midrule
\multirow{2}{*}{zh} & Native & \cellcolor[HTML]{E1E9E9}30.4 & \cellcolor[HTML]{8AC1CF}\textbf{28.1} & \cellcolor[HTML]{B4D1ED}30.3 & \cellcolor[HTML]{B1CFEF}\textbf{30.4} & \cellcolor[HTML]{A3C9EC}\textbf{29.5} & \cellcolor[HTML]{A9CCF1}\textbf{30.4} & \cellcolor[HTML]{B4D1EE}30.7 & \cellcolor[HTML]{B5D2EE}\textbf{29.3} & \cellcolor[HTML]{F1E7B6}\textbf{29.8} & \cellcolor[HTML]{ECEDDE}\textbf{24.3} & \cellcolor[HTML]{F3CA8F}12.1 & \cellcolor[HTML]{CEDFE8}\textbf{27.8} \\
 & English (ineffective) & \cellcolor[HTML]{DBE5E8}\textbf{30.5} & \cellcolor[HTML]{8BC2D1}28.0 & \cellcolor[HTML]{B1CFEF}\textbf{30.3} & \cellcolor[HTML]{D3E1E8}29.8 & \cellcolor[HTML]{E4EBEA}28.2 & \cellcolor[HTML]{B6D2ED}30.1 & \cellcolor[HTML]{ACCDF1}\textbf{30.8} & \cellcolor[HTML]{BFD7EB}29.1 & \cellcolor[HTML]{F3E7B1}29.7 & \cellcolor[HTML]{EDE8C4}23.8 & \cellcolor[HTML]{F3E7B2}\textbf{12.7} & \cellcolor[HTML]{D9E5E8}27.6 \\
\midrule
\multirow{2}{*}{de} & Native & \cellcolor[HTML]{CADCE9}30.8 & \cellcolor[HTML]{B9D4EC}26.5 & \cellcolor[HTML]{9CC7E6}30.9 & \cellcolor[HTML]{A9CCF1}30.6 & \cellcolor[HTML]{9FC8E9}29.6 & \cellcolor[HTML]{A7CBF0}30.4 & \cellcolor[HTML]{D6E3E8}29.9 & \cellcolor[HTML]{9BC7E5}30.1 & \cellcolor[HTML]{F5E7AC}29.7 & \cellcolor[HTML]{F5D999}22.9 & \cellcolor[HTML]{DFE8E9}13.6 & \cellcolor[HTML]{CFDFE8}27.7 \\
\bottomrule
\end{tabular}%
}
\end{minipage}
\end{table}

\begin{table}[!htbp]
\centering
\begin{minipage}[t]{0.49\textwidth}
\centering
\caption{Average over all unseen tasks from Multilingual Reasoning Gym -- Avg@8 percentage-point improvement over the base model \texttt{Qwen3-4B-Base}}
\label{tab:appendix_crosslingual_qwen3_4b_base_3_appendix_mrg_unseen_macro}
\tiny
\setlength{\tabcolsep}{2pt}
\renewcommand{\arraystretch}{0.92}
\newcommand{\MatrixResizeWidth}{\linewidth}
\resizebox{\MatrixResizeWidth}{!}{%
\begin{tabular}{llrrrrrrrrrrrrrrVr}
\toprule
Train & Reward & en & zh & de & fr & es & ru & ja & th & bn & te & sw & it & ko & pt & Avg \\
\midrule
\multirow{1}{*}{en} & English & \cellcolor[HTML]{E9EEEC}\textbf{16.1} & \cellcolor[HTML]{92C4DA}\textbf{15.9} & \cellcolor[HTML]{B8D3ED}\textbf{15.2} & \cellcolor[HTML]{C4D9EA}\textbf{16.1} & \cellcolor[HTML]{8CC2D2}\textbf{15.9} & \cellcolor[HTML]{ADCDF0}\textbf{15.6} & \cellcolor[HTML]{A9CCF1}\textbf{16.4} & \cellcolor[HTML]{A1C9EA}\textbf{15.9} & \cellcolor[HTML]{A8CBF1}\textbf{16.5} & \cellcolor[HTML]{EDEFE6}\textbf{11.9} & \cellcolor[HTML]{EDE8C6}\textbf{9.1} & \cellcolor[HTML]{90C4D8}\textbf{15.4} & \cellcolor[HTML]{AECEF0}\textbf{17.1} & \cellcolor[HTML]{A5CAEE}\textbf{16.8} & \cellcolor[HTML]{BFD7EB}\textbf{15.3} \\
\midrule
\multirow{2}{*}{zh} & Native & \cellcolor[HTML]{EDEEE3}\textbf{15.7} & \cellcolor[HTML]{C1D7EA}15.0 & \cellcolor[HTML]{EBEBD6}\textbf{13.5} & \cellcolor[HTML]{EAEFED}\textbf{15.6} & \cellcolor[HTML]{BFD7EB}\textbf{15.1} & \cellcolor[HTML]{B2D0EF}\textbf{15.5} & \cellcolor[HTML]{F0E7BA}14.5 & \cellcolor[HTML]{F6E1A0}\textbf{13.3} & \cellcolor[HTML]{EDE8C8}\textbf{15.1} & \cellcolor[HTML]{CCDDE9}\textbf{12.4} & \cellcolor[HTML]{F2C48B}\textbf{8.5} & \cellcolor[HTML]{EDEFE4}\textbf{13.8} & \cellcolor[HTML]{F1E7B6}\textbf{15.6} & \cellcolor[HTML]{F3F6F2}\textbf{15.7} & \cellcolor[HTML]{EBEAD0}\textbf{14.2} \\
 & English (ineffective) & \cellcolor[HTML]{EDE8C3}15.2 & \cellcolor[HTML]{BDD5EB}\textbf{15.0} & \cellcolor[HTML]{ECE9CB}13.3 & \cellcolor[HTML]{F3E7AF}14.7 & \cellcolor[HTML]{F2E7B3}13.7 & \cellcolor[HTML]{E2EAE9}14.8 & \cellcolor[HTML]{EBEBD6}\textbf{14.9} & \cellcolor[HTML]{F6DF9E}13.2 & \cellcolor[HTML]{F6E7AA}14.8 & \cellcolor[HTML]{F5E7A9}11.3 & \cellcolor[HTML]{EC856B}8.0 & \cellcolor[HTML]{F6E1A0}12.9 & \cellcolor[HTML]{ED9070}14.1 & \cellcolor[HTML]{ECE9CA}15.2 & \cellcolor[HTML]{F6E0A0}13.7 \\
\bottomrule
\end{tabular}%
}
\end{minipage}
\hfill
\begin{minipage}[t]{0.49\textwidth}
\centering
\caption{Main heldout tasks from Multilingual Reasoning Gym -- Avg@8 percentage-point improvement over the base model \texttt{Qwen3-4B-Base}}
\label{tab:appendix_crosslingual_qwen3_4b_base_4_heldout_filtered}
\tiny
\setlength{\tabcolsep}{2pt}
\renewcommand{\arraystretch}{0.92}
\newcommand{\MatrixResizeWidth}{\linewidth}
\resizebox{\MatrixResizeWidth}{!}{%
\begin{tabular}{llrrrrrrrrrrrrrrVr}
\toprule
Train & Reward & en & zh & de & fr & es & ru & ja & th & bn & te & sw & it & ko & pt & Avg \\
\midrule
\multirow{1}{*}{en} & English & \cellcolor[HTML]{ECEDDE}\textbf{17.2} & \cellcolor[HTML]{9FC8E9}\textbf{16.4} & \cellcolor[HTML]{C7DBE9}\textbf{18.1} & \cellcolor[HTML]{C7DBE9}\textbf{17.9} & \cellcolor[HTML]{85C1C9}\textbf{18.1} & \cellcolor[HTML]{EBECDD}\textbf{15.3} & \cellcolor[HTML]{9CC7E6}\textbf{19.3} & \cellcolor[HTML]{ECE9CA}\textbf{16.2} & \cellcolor[HTML]{DCE7E8}\textbf{15.1} & \cellcolor[HTML]{D2E1E8}\textbf{15.0} & \cellcolor[HTML]{C5DAE9}\textbf{16.9} & \cellcolor[HTML]{95C5DE}\textbf{18.6} & \cellcolor[HTML]{C6DAE9}\textbf{18.7} & \cellcolor[HTML]{81C0C3}\textbf{19.1} & \cellcolor[HTML]{C6DAEA}\textbf{17.3} \\
\midrule
\multirow{2}{*}{zh} & Native & \cellcolor[HTML]{A2C9EB}\textbf{18.4} & \cellcolor[HTML]{A0C8EA}16.4 & \cellcolor[HTML]{F5D898}16.3 & \cellcolor[HTML]{EDF0E6}17.2 & \cellcolor[HTML]{9FC8E9}17.5 & \cellcolor[HTML]{BED6EB}16.2 & \cellcolor[HTML]{EDE8C3}17.1 & \cellcolor[HTML]{EE9270}14.7 & \cellcolor[HTML]{ECE9CC}14.3 & \cellcolor[HTML]{B9D3EC}\textbf{15.4} & \cellcolor[HTML]{F5D697}14.7 & \cellcolor[HTML]{D8E4E8}\textbf{17.7} & \cellcolor[HTML]{F5E7AB}17.2 & \cellcolor[HTML]{D4E2E8}17.6 & \cellcolor[HTML]{EDE8C4}16.5 \\
 & English (ineffective) & \cellcolor[HTML]{A2C9EB}\textbf{18.4} & \cellcolor[HTML]{5CBE78}\underline{\textbf{18.1}} & \cellcolor[HTML]{BAD4EC}\textbf{18.2} & \cellcolor[HTML]{EDF1EE}\textbf{17.4} & \cellcolor[HTML]{9CC7E6}\textbf{17.6} & \cellcolor[HTML]{B3D1EE}\textbf{16.4} & \cellcolor[HTML]{88C2CE}\textbf{19.9} & \cellcolor[HTML]{F4CF92}\textbf{15.4} & \cellcolor[HTML]{A6CAEF}\textbf{15.9} & \cellcolor[HTML]{D4E2E8}15.0 & \cellcolor[HTML]{EDF0E6}\textbf{16.0} & \cellcolor[HTML]{DEE7E9}17.6 & \cellcolor[HTML]{CEDEE8}\textbf{18.6} & \cellcolor[HTML]{90C3D7}\textbf{18.7} & \cellcolor[HTML]{BAD4EC}\textbf{17.4} \\
\bottomrule
\end{tabular}%
}
\end{minipage}
\end{table}

\begin{table}[!htbp]
\centering
\begin{minipage}[t]{0.49\textwidth}
\centering
\caption{English-content tasks from Multilingual Reasoning Gym, not seen during training -- Avg@8 percentage-point improvement over the base model \texttt{Qwen3-4B-Base}}
\label{tab:appendix_crosslingual_qwen3_4b_base_5_english_content_add}
\tiny
\setlength{\tabcolsep}{2pt}
\renewcommand{\arraystretch}{0.92}
\newcommand{\MatrixResizeWidth}{\linewidth}
\resizebox{\MatrixResizeWidth}{!}{%
\begin{tabular}{llrrrrrrrrrrrrrrVr}
\toprule
Train & Reward & en & zh & de & fr & es & ru & ja & th & bn & te & sw & it & ko & pt & Avg \\
\midrule
\multirow{1}{*}{en} & English & \cellcolor[HTML]{E1E9E9}\textbf{30.6} & \cellcolor[HTML]{98C6E1}\textbf{30.8} & \cellcolor[HTML]{ABCDF1}\textbf{27.2} & \cellcolor[HTML]{CFDFE8}\textbf{30.2} & \cellcolor[HTML]{95C5DD}\textbf{29.3} & \cellcolor[HTML]{96C5E0}\textbf{31.0} & \cellcolor[HTML]{A4CAED}\textbf{29.4} & \cellcolor[HTML]{85C1C8}\textbf{31.1} & \cellcolor[HTML]{82C0C3}\textbf{33.9} & \cellcolor[HTML]{ECEDE0}\textbf{20.4} & \cellcolor[HTML]{F7E4A3}\textbf{10.1} & \cellcolor[HTML]{8AC1CF}\textbf{27.4} & \cellcolor[HTML]{9DC7E6}\textbf{32.2} & \cellcolor[HTML]{BDD6EB}\textbf{30.8} & \cellcolor[HTML]{ACCDF1}\textbf{28.2} \\
\midrule
\multirow{2}{*}{zh} & Native & \cellcolor[HTML]{F0E7B9}\textbf{28.3} & \cellcolor[HTML]{E0E9E9}\textbf{28.1} & \cellcolor[HTML]{F3F6F2}\textbf{24.1} & \cellcolor[HTML]{EDF1EE}\textbf{29.4} & \cellcolor[HTML]{D4E2E8}\textbf{27.3} & \cellcolor[HTML]{A8CBF1}\textbf{30.0} & \cellcolor[HTML]{EBEAD4}\textbf{26.0} & \cellcolor[HTML]{ECE9C9}\textbf{24.9} & \cellcolor[HTML]{C0D7EA}\textbf{30.8} & \cellcolor[HTML]{D3E1E8}\textbf{21.6} & \cellcolor[HTML]{EEE7BF}\textbf{10.6} & \cellcolor[HTML]{EEF1E9}\textbf{23.2} & \cellcolor[HTML]{ECEEE2}\textbf{29.2} & \cellcolor[HTML]{ECF0ED}\textbf{29.6} & \cellcolor[HTML]{EFF1EA}\textbf{25.9} \\
 & English (ineffective) & \cellcolor[HTML]{F5D596}27.0 & \cellcolor[HTML]{EBEBD8}26.6 & \cellcolor[HTML]{F3E7B2}21.4 & \cellcolor[HTML]{F3C98E}26.3 & \cellcolor[HTML]{F2C38A}23.2 & \cellcolor[HTML]{F2F4F1}27.7 & \cellcolor[HTML]{F7E5A4}24.6 & \cellcolor[HTML]{F6E7A6}23.7 & \cellcolor[HTML]{EBECDC}28.3 & \cellcolor[HTML]{F6E1A0}18.8 & \cellcolor[HTML]{E85C5C}7.7 & \cellcolor[HTML]{F5D999}20.7 & \cellcolor[HTML]{E85D5C}23.6 & \cellcolor[HTML]{F3CB8F}26.4 & \cellcolor[HTML]{F4D294}23.3 \\
\bottomrule
\end{tabular}%
}
\end{minipage}
\hfill
\begin{minipage}[t]{0.49\textwidth}
\centering
\caption{Very difficult tasks from Multilingual Reasoning Gym, not seen during training -- Avg@8 percentage-point improvement over the base model \texttt{Qwen3-4B-Base}}
\label{tab:appendix_crosslingual_qwen3_4b_base_6_difficulty_addback}
\tiny
\setlength{\tabcolsep}{2pt}
\renewcommand{\arraystretch}{0.92}
\newcommand{\MatrixResizeWidth}{\linewidth}
\resizebox{\MatrixResizeWidth}{!}{%
\begin{tabular}{llrrrrrrrrrrrrrrVr}
\toprule
Train & Reward & en & zh & de & fr & es & ru & ja & th & bn & te & sw & it & ko & pt & Avg \\
\midrule
\multirow{1}{*}{en} & English & \cellcolor[HTML]{E8EEEB}\textbf{0.5} & \cellcolor[HTML]{DBE6E8}\textbf{0.5} & \cellcolor[HTML]{F3C78D}\textbf{0.2} & \cellcolor[HTML]{F4CF92}\textbf{0.3} & \cellcolor[HTML]{F1E7B7}\textbf{0.4} & \cellcolor[HTML]{E1EAE9}\textbf{0.5} & \cellcolor[HTML]{A7CBEF}\textbf{0.6} & \cellcolor[HTML]{F7E7A6}\textbf{0.5} & \cellcolor[HTML]{F2E7B3}\textbf{0.4} & \cellcolor[HTML]{F2BA85}\textbf{0.2} & \cellcolor[HTML]{EBEAD0}\textbf{0.2} & \cellcolor[HTML]{F0AD7D}\textbf{0.2} & \cellcolor[HTML]{B9D4EC}\textbf{0.6} & \cellcolor[HTML]{CFDFE8}\textbf{0.5} & \cellcolor[HTML]{EEF0E7}\textbf{0.4} \\
\midrule
\multirow{2}{*}{zh} & Native & \cellcolor[HTML]{F6F8F5}\textbf{0.4} & \cellcolor[HTML]{ECEEE2}\textbf{0.4} & \cellcolor[HTML]{ED8F6F}0.2 & \cellcolor[HTML]{F5D999}0.3 & \cellcolor[HTML]{EBECDB}\textbf{0.4} & \cellcolor[HTML]{F5D999}0.3 & \cellcolor[HTML]{CBDDE9}\textbf{0.5} & \cellcolor[HTML]{EC8C6D}0.2 & \cellcolor[HTML]{F4D293}0.3 & \cellcolor[HTML]{EC7F69}\textbf{0.1} & \cellcolor[HTML]{F4D395}0.2 & \cellcolor[HTML]{F6DF9E}\textbf{0.3} & \cellcolor[HTML]{F4D394}\textbf{0.3} & \cellcolor[HTML]{EA7565}0.1 & \cellcolor[HTML]{F6E09F}0.3 \\
 & English (ineffective) & \cellcolor[HTML]{F1E7B7}0.3 & \cellcolor[HTML]{EDE8C4}0.3 & \cellcolor[HTML]{F7E7A6}\textbf{0.3} & \cellcolor[HTML]{EBEAD3}\textbf{0.4} & \cellcolor[HTML]{F6E2A1}0.3 & \cellcolor[HTML]{F6E1A0}\textbf{0.3} & \cellcolor[HTML]{F5DC9C}0.3 & \cellcolor[HTML]{F7E7A6}\textbf{0.5} & \cellcolor[HTML]{F5D898}\textbf{0.3} & \cellcolor[HTML]{E85C5C}0.0 & \cellcolor[HTML]{C2D8EA}\textbf{0.3} & \cellcolor[HTML]{F5D697}0.3 & \cellcolor[HTML]{F4D394}\textbf{0.3} & \cellcolor[HTML]{ECE9CD}\textbf{0.4} & \cellcolor[HTML]{F1E7B4}\textbf{0.3} \\
\bottomrule
\end{tabular}%
}
\end{minipage}
\end{table}

\begin{table}[!htbp]
\centering
\begin{minipage}[t]{0.49\textwidth}
\centering
\caption{Easy Math (MGSM + PolyMath low) -- Avg@8 percentage-point improvement over the base model \texttt{Qwen3-4B-Base}; $\dagger$: includes only PolyMath low because MGSM is unavailable for this language.}
\label{tab:appendix_crosslingual_qwen3_4b_base_7_appendix_easy_math_macro}
\tiny
\setlength{\tabcolsep}{2pt}
\renewcommand{\arraystretch}{0.92}
\newcommand{\MatrixResizeWidth}{\linewidth}
\resizebox{\MatrixResizeWidth}{!}{%
\begin{tabular}{llrrrrrrrrrrrrrrrrrrVr}
\toprule
Train & Reward & en & zh & de & fr & es & ru & ja & th & bn & te & sw & it$^{\dagger}$ & ko$^{\dagger}$ & pt$^{\dagger}$ & ar$^{\dagger}$ & id$^{\dagger}$ & ms$^{\dagger}$ & vi$^{\dagger}$ & Avg \\
\midrule
\multirow{1}{*}{en} & English & \cellcolor[HTML]{68BE92}\textbf{67.5} & \cellcolor[HTML]{64BD8B}\textbf{57.2} & \cellcolor[HTML]{63BE8B}\textbf{67.4} & \cellcolor[HTML]{6DBE9C}\textbf{67.0} & \cellcolor[HTML]{66BE8D}\textbf{64.7} & \cellcolor[HTML]{6FBEA2}\textbf{67.3} & \cellcolor[HTML]{7ABEB5}\textbf{65.4} & \cellcolor[HTML]{76BFAF}\textbf{64.9} & \cellcolor[HTML]{94C4DC}\textbf{68.0} & \cellcolor[HTML]{D8E4E8}\textbf{53.3} & \cellcolor[HTML]{F0E7B9}\textbf{21.5} & \cellcolor[HTML]{5CBE78}\underline{\textbf{72.5}} & \cellcolor[HTML]{7DBFBC}\textbf{67.5} & \cellcolor[HTML]{5EBE7C}\textbf{68.6} & \cellcolor[HTML]{6FBEA2}\textbf{75.5} & \cellcolor[HTML]{71BEA7}\textbf{72.3} & \cellcolor[HTML]{70BEA4}\textbf{69.4} & \cellcolor[HTML]{8BC2D0}\textbf{66.8} & \cellcolor[HTML]{6FBEA1}\textbf{64.3} \\
\midrule
\multirow{2}{*}{zh} & Native & \cellcolor[HTML]{CFDFE8}\textbf{63.9} & \cellcolor[HTML]{76BFAF}56.1 & \cellcolor[HTML]{99C6E3}\textbf{64.3} & \cellcolor[HTML]{ADCDF0}\textbf{63.9} & \cellcolor[HTML]{C3D9EA}\textbf{60.6} & \cellcolor[HTML]{C8DBE9}\textbf{63.2} & \cellcolor[HTML]{9FC7E8}\textbf{63.5} & \cellcolor[HTML]{99C6E2}\textbf{62.6} & \cellcolor[HTML]{EE9270}61.3 & \cellcolor[HTML]{EA6C62}47.9 & \cellcolor[HTML]{E85C5C}19.2 & \cellcolor[HTML]{9CC7E6}\textbf{68.8} & \cellcolor[HTML]{C3D9EA}\textbf{63.8} & \cellcolor[HTML]{C0D7EB}\textbf{64.7} & \cellcolor[HTML]{CCDEE8}\textbf{70.3} & \cellcolor[HTML]{E6ECEB}\textbf{67.0} & \cellcolor[HTML]{CDDEE9}\textbf{64.5} & \cellcolor[HTML]{81C0C2}\textbf{67.4} & \cellcolor[HTML]{C7DAE9}\textbf{60.7} \\
\bottomrule
\end{tabular}%
}
\end{minipage}
\hfill
\begin{minipage}[t]{0.49\textwidth}
\centering
\caption{Hard Math (PolyMath medium, high, and top) -- Avg@8 percentage-point improvement over the base model \texttt{Qwen3-4B-Base}}
\label{tab:appendix_crosslingual_qwen3_4b_base_8_appendix_hard_math_macro}
\tiny
\setlength{\tabcolsep}{2pt}
\renewcommand{\arraystretch}{0.92}
\newcommand{\MatrixResizeWidth}{\linewidth}
\resizebox{\MatrixResizeWidth}{!}{%
\begin{tabular}{llrrrrrrrrrrrrrrrrrrVr}
\toprule
Train & Reward & en & zh & de & fr & es & ru & ja & th & bn & te & sw & it & ko & pt & ar & id & ms & vi & Avg \\
\midrule
\multirow{1}{*}{en} & English & \cellcolor[HTML]{83C0C6}\textbf{14.8} & \cellcolor[HTML]{7BBFB9}\textbf{14.3} & \cellcolor[HTML]{6ABE96}\textbf{15.6} & \cellcolor[HTML]{6EBEA0}\textbf{14.1} & \cellcolor[HTML]{78BEB2}\textbf{14.2} & \cellcolor[HTML]{73BEA9}\textbf{14.1} & \cellcolor[HTML]{6FBEA1}\textbf{15.8} & \cellcolor[HTML]{88C2CD}\textbf{13.9} & \cellcolor[HTML]{9BC7E5}\textbf{14.1} & \cellcolor[HTML]{7CBFBA}\textbf{13.5} & \cellcolor[HTML]{84C0C6}\textbf{10.9} & \cellcolor[HTML]{6DBE9E}\textbf{15.5} & \cellcolor[HTML]{71BEA5}\textbf{15.9} & \cellcolor[HTML]{7EBFBC}\textbf{13.9} & \cellcolor[HTML]{9CC7E6}\textbf{13.1} & \cellcolor[HTML]{7DBFBB}\textbf{14.0} & \cellcolor[HTML]{6CBE9B}\textbf{14.5} & \cellcolor[HTML]{73BEAB}\textbf{14.5} & \cellcolor[HTML]{78BEB2}\textbf{14.3} \\
\midrule
\multirow{2}{*}{zh} & Native & \cellcolor[HTML]{F0F3ED}11.6 & \cellcolor[HTML]{91C3D8}\textbf{13.2} & \cellcolor[HTML]{B2D0EE}12.9 & \cellcolor[HTML]{ADCDF0}11.8 & \cellcolor[HTML]{95C4DD}\textbf{12.9} & \cellcolor[HTML]{99C6E3}12.3 & \cellcolor[HTML]{97C5E0}13.9 & \cellcolor[HTML]{C2D8EA}12.2 & \cellcolor[HTML]{C8DBE9}\textbf{13.1} & \cellcolor[HTML]{8EC3D5}\textbf{12.6} & \cellcolor[HTML]{E9EEEC}8.6 & \cellcolor[HTML]{9DC7E7}\textbf{13.3} & \cellcolor[HTML]{A1C8EA}13.8 & \cellcolor[HTML]{A4CAED}\textbf{12.3} & \cellcolor[HTML]{98C6E1}\textbf{13.2} & \cellcolor[HTML]{B1CFEF}\textbf{11.9} & \cellcolor[HTML]{A6CBEF}12.0 & \cellcolor[HTML]{C3D8EA}11.5 & \cellcolor[HTML]{A6CAEF}\textbf{12.4} \\
\bottomrule
\end{tabular}%
}
\end{minipage}
\end{table}

\FloatBarrier
\subsection{Qwen3-8B-Base}

\begin{table}[!htbp]
\centering
\begin{minipage}[t]{0.49\textwidth}
\centering
\caption{Tasks seen during training from Multilingual Reasoning Gym -- Avg@8 percentage-point improvement over the base model \texttt{Qwen3-8B-Base}}
\label{tab:appendix_crosslingual_qwen3_8b_base_1_train_filtered_noheldout}
\tiny
\setlength{\tabcolsep}{2pt}
\renewcommand{\arraystretch}{0.92}
\newcommand{\MatrixResizeWidth}{\linewidth}
\resizebox{\MatrixResizeWidth}{!}{%
\begin{tabular}{llrrrrrrrrrrrrrrVr}
\toprule
Train & Reward & en & zh & de & fr & es & ru & ja & th & bn & te & sw & it & ko & pt & Avg \\
\midrule
\multirow{1}{*}{en} & English & \cellcolor[HTML]{6EBEA0}\textbf{40.9} & \cellcolor[HTML]{DDE7E8}\textbf{35.9} & \cellcolor[HTML]{B2D0EF}\textbf{40.7} & \cellcolor[HTML]{E0E9E9}\textbf{38.3} & \cellcolor[HTML]{B8D3EC}\textbf{38.7} & \cellcolor[HTML]{CFDFE8}\textbf{35.6} & \cellcolor[HTML]{ECECDD}\textbf{37.8} & \cellcolor[HTML]{F0F2EC}\textbf{41.4} & \cellcolor[HTML]{F2BA84}\textbf{37.2} & \cellcolor[HTML]{F2BA85}\textbf{34.8} & \cellcolor[HTML]{EC8A6D}\textbf{34.4} & \cellcolor[HTML]{E5ECEA}\textbf{38.0} & \cellcolor[HTML]{DDE7E9}\textbf{40.0} & \cellcolor[HTML]{BFD6EB}\textbf{37.4} & \cellcolor[HTML]{CFDFE8}\textbf{37.9} \\
\midrule
\multirow{2}{*}{zh} & Native & \cellcolor[HTML]{A4CAED}37.7 & \cellcolor[HTML]{74BFAC}42.5 & \cellcolor[HTML]{B7D2ED}40.5 & \cellcolor[HTML]{EEF0E7}37.1 & \cellcolor[HTML]{EBEAD2}34.5 & \cellcolor[HTML]{CBDDE9}35.8 & \cellcolor[HTML]{92C4DA}43.8 & \cellcolor[HTML]{E3EAEA}42.3 & \cellcolor[HTML]{F5DB9B}38.9 & \cellcolor[HTML]{F4E7AD}38.1 & \cellcolor[HTML]{EE9873}35.0 & \cellcolor[HTML]{EBECDC}36.6 & \cellcolor[HTML]{C1D7EB}41.2 & \cellcolor[HTML]{EFF2EB}34.8 & \cellcolor[HTML]{BDD6EB}38.5 \\
 & English & \cellcolor[HTML]{7ABFB6}\textbf{40.1} & \cellcolor[HTML]{5CBE78}\underline{\textbf{44.8}} & \cellcolor[HTML]{85C1C8}\textbf{43.6} & \cellcolor[HTML]{A6CAEF}\textbf{40.5} & \cellcolor[HTML]{B8D3ED}\textbf{38.7} & \cellcolor[HTML]{86C0CA}\textbf{40.0} & \cellcolor[HTML]{70BEA4}\textbf{46.7} & \cellcolor[HTML]{92C4DA}\textbf{46.0} & \cellcolor[HTML]{EDF1EE}\textbf{42.9} & \cellcolor[HTML]{F1F4F0}\textbf{41.6} & \cellcolor[HTML]{F5DD9C}\textbf{39.0} & \cellcolor[HTML]{B0CFEF}\textbf{39.9} & \cellcolor[HTML]{6FBEA0}\textbf{46.6} & \cellcolor[HTML]{A0C8EA}\textbf{38.9} & \cellcolor[HTML]{74BEAD}\textbf{42.1} \\
\bottomrule
\end{tabular}%
}
\end{minipage}
\hfill
\begin{minipage}[t]{0.49\textwidth}
\centering
\caption{Macro average over all unseen tasks -- Avg@8 percentage-point improvement over the base model \texttt{Qwen3-8B-Base}}
\label{tab:appendix_crosslingual_qwen3_8b_base_2_appendix_unseen_task_groups_macro}
\tiny
\setlength{\tabcolsep}{2pt}
\renewcommand{\arraystretch}{0.92}
\newcommand{\MatrixResizeWidth}{\linewidth}
\resizebox{\MatrixResizeWidth}{!}{%
\begin{tabular}{llrrrrrrrrrrrVr}
\toprule
Train & Reward & en & zh & de & fr & es & ru & ja & th & bn & te & sw & Avg \\
\midrule
\multirow{1}{*}{en} & English & \cellcolor[HTML]{63BE89}\textbf{25.2} & \cellcolor[HTML]{79BFB3}\textbf{26.9} & \cellcolor[HTML]{6ABE96}\textbf{29.8} & \cellcolor[HTML]{7ABFB6}\textbf{28.3} & \cellcolor[HTML]{6BBE98}\textbf{27.0} & \cellcolor[HTML]{72BEA9}\textbf{27.4} & \cellcolor[HTML]{98C6E2}\textbf{29.7} & \cellcolor[HTML]{88C1CD}\textbf{30.7} & \cellcolor[HTML]{9EC8E7}\textbf{30.1} & \cellcolor[HTML]{ACCDF1}\textbf{27.9} & \cellcolor[HTML]{A7CBEF}\textbf{24.4} & \cellcolor[HTML]{7EC0BD}\textbf{28.0} \\
\midrule
\multirow{2}{*}{zh} & Native & \cellcolor[HTML]{E8EEEB}21.7 & \cellcolor[HTML]{97C5E1}25.6 & \cellcolor[HTML]{9DC7E7}27.5 & \cellcolor[HTML]{C8DBE9}25.9 & \cellcolor[HTML]{D6E3E8}23.6 & \cellcolor[HTML]{D3E1E8}23.8 & \cellcolor[HTML]{9CC7E6}29.5 & \cellcolor[HTML]{DDE7E8}28.5 & \cellcolor[HTML]{ECE8C8}27.1 & \cellcolor[HTML]{EBECDC}25.6 & \cellcolor[HTML]{F0B380}19.8 & \cellcolor[HTML]{D7E3E8}25.3 \\
 & English & \cellcolor[HTML]{7CBFB9}\textbf{24.3} & \cellcolor[HTML]{77BFB0}\textbf{27.0} & \cellcolor[HTML]{6ABE97}\textbf{29.8} & \cellcolor[HTML]{86C1CB}\textbf{27.8} & \cellcolor[HTML]{77BFB0}\textbf{26.5} & \cellcolor[HTML]{74BFAC}\textbf{27.3} & \cellcolor[HTML]{7BBFB8}\textbf{31.1} & \cellcolor[HTML]{7EBFBC}\textbf{31.2} & \cellcolor[HTML]{93C4DB}\textbf{30.5} & \cellcolor[HTML]{85C1C9}\textbf{29.5} & \cellcolor[HTML]{9FC7E8}\textbf{24.6} & \cellcolor[HTML]{7ABEB5}\textbf{28.2} \\
\midrule
\multirow{2}{*}{de} & Native & \cellcolor[HTML]{90C3D7}23.6 & \cellcolor[HTML]{9AC6E3}25.5 & \cellcolor[HTML]{89C1CF}28.3 & \cellcolor[HTML]{9CC7E6}27.0 & \cellcolor[HTML]{9BC7E5}25.0 & \cellcolor[HTML]{9FC8E9}25.3 & \cellcolor[HTML]{9EC8E8}29.5 & \cellcolor[HTML]{AFCFEF}29.4 & \cellcolor[HTML]{D8E4E8}28.7 & \cellcolor[HTML]{B2D0EE}27.8 & \cellcolor[HTML]{EBEAD0}22.0 & \cellcolor[HTML]{A2C8EB}26.6 \\
\bottomrule
\end{tabular}%
}
\end{minipage}
\end{table}

\begin{table}[!htbp]
\centering
\begin{minipage}[t]{0.49\textwidth}
\centering
\caption{Average over all unseen tasks from Multilingual Reasoning Gym -- Avg@8 percentage-point improvement over the base model \texttt{Qwen3-8B-Base}}
\label{tab:appendix_crosslingual_qwen3_8b_base_3_appendix_mrg_unseen_macro}
\tiny
\setlength{\tabcolsep}{2pt}
\renewcommand{\arraystretch}{0.92}
\newcommand{\MatrixResizeWidth}{\linewidth}
\resizebox{\MatrixResizeWidth}{!}{%
\begin{tabular}{llrrrrrrrrrrrrrrVr}
\toprule
Train & Reward & en & zh & de & fr & es & ru & ja & th & bn & te & sw & it & ko & pt & Avg \\
\midrule
\multirow{1}{*}{en} & English & \cellcolor[HTML]{6EBEA0}\textbf{19.7} & \cellcolor[HTML]{7DBFBB}\textbf{19.1} & \cellcolor[HTML]{7BBFB8}\textbf{20.3} & \cellcolor[HTML]{98C6E1}\textbf{20.1} & \cellcolor[HTML]{8CC2D2}\textbf{19.3} & \cellcolor[HTML]{7EC0BC}\textbf{19.9} & \cellcolor[HTML]{A7CBEF}\textbf{20.4} & \cellcolor[HTML]{9FC7E8}\textbf{19.6} & \cellcolor[HTML]{ABCDF1}\textbf{19.4} & \cellcolor[HTML]{B7D2ED}\textbf{16.9} & \cellcolor[HTML]{EBEAD3}\textbf{13.9} & \cellcolor[HTML]{93C4DB}\textbf{18.0} & \cellcolor[HTML]{7EC0BD}\textbf{21.7} & \cellcolor[HTML]{91C4D8}\textbf{20.1} & \cellcolor[HTML]{90C3D8}\textbf{19.2} \\
\midrule
\multirow{2}{*}{zh} & Native & \cellcolor[HTML]{EBEDDE}15.4 & \cellcolor[HTML]{A1C9EA}17.5 & \cellcolor[HTML]{A1C9EA}18.9 & \cellcolor[HTML]{C7DBE9}18.9 & \cellcolor[HTML]{D8E4E8}17.3 & \cellcolor[HTML]{EFF1EA}15.8 & \cellcolor[HTML]{D2E0E8}19.4 & \cellcolor[HTML]{F0F2EC}17.7 & \cellcolor[HTML]{F1F3ED}18.0 & \cellcolor[HTML]{E0E9E9}16.1 & \cellcolor[HTML]{F0AC7D}12.1 & \cellcolor[HTML]{E4EBEA}16.0 & \cellcolor[HTML]{EDEFE3}18.6 & \cellcolor[HTML]{F3F6F2}17.3 & \cellcolor[HTML]{E5ECEA}17.1 \\
 & English & \cellcolor[HTML]{A2C8EB}\textbf{17.7} & \cellcolor[HTML]{84C0C6}\textbf{18.7} & \cellcolor[HTML]{79BFB4}\textbf{20.4} & \cellcolor[HTML]{A0C8EA}\textbf{19.8} & \cellcolor[HTML]{90C3D7}\textbf{19.1} & \cellcolor[HTML]{80C0C1}\textbf{19.8} & \cellcolor[HTML]{98C6E1}\textbf{20.9} & \cellcolor[HTML]{B1D0EF}\textbf{19.1} & \cellcolor[HTML]{85C1C8}\textbf{20.6} & \cellcolor[HTML]{7DBFBC}\textbf{18.8} & \cellcolor[HTML]{E0E9E9}\textbf{14.8} & \cellcolor[HTML]{C0D7EB}\textbf{16.7} & \cellcolor[HTML]{9FC8E9}\textbf{20.6} & \cellcolor[HTML]{96C5E0}\textbf{19.8} & \cellcolor[HTML]{93C4DB}\textbf{19.1} \\
\bottomrule
\end{tabular}%
}
\end{minipage}
\hfill
\begin{minipage}[t]{0.49\textwidth}
\centering
\caption{Main heldout tasks from Multilingual Reasoning Gym -- Avg@8 percentage-point improvement over the base model \texttt{Qwen3-8B-Base}}
\label{tab:appendix_crosslingual_qwen3_8b_base_4_heldout_filtered}
\tiny
\setlength{\tabcolsep}{2pt}
\renewcommand{\arraystretch}{0.92}
\newcommand{\MatrixResizeWidth}{\linewidth}
\resizebox{\MatrixResizeWidth}{!}{%
\begin{tabular}{llrrrrrrrrrrrrrrVr}
\toprule
Train & Reward & en & zh & de & fr & es & ru & ja & th & bn & te & sw & it & ko & pt & Avg \\
\midrule
\multirow{1}{*}{en} & English & \cellcolor[HTML]{5CBE78}\underline{\textbf{24.2}} & \cellcolor[HTML]{8CC2D2}\textbf{20.9} & \cellcolor[HTML]{E7EDEB}\textbf{22.4} & \cellcolor[HTML]{F5D898}\textbf{20.5} & \cellcolor[HTML]{C4D9EA}\textbf{21.3} & \cellcolor[HTML]{87C1CC}\textbf{21.6} & \cellcolor[HTML]{DEE7E9}\textbf{24.4} & \cellcolor[HTML]{F6E1A0}\textbf{19.6} & \cellcolor[HTML]{F0E7B8}\textbf{19.6} & \cellcolor[HTML]{EDE8C7}\textbf{17.6} & \cellcolor[HTML]{EBEBD6}\textbf{17.8} & \cellcolor[HTML]{A0C8EA}\textbf{22.8} & \cellcolor[HTML]{D1E0E8}\textbf{23.1} & \cellcolor[HTML]{CCDDE9}\textbf{21.2} & \cellcolor[HTML]{CFDFE8}\textbf{21.2} \\
\midrule
\multirow{2}{*}{zh} & Native & \cellcolor[HTML]{F4D092}19.2 & \cellcolor[HTML]{A9CCF1}19.8 & \cellcolor[HTML]{EDF1EE}22.3 & \cellcolor[HTML]{A5CAEE}\textbf{23.2} & \cellcolor[HTML]{C0D7EB}21.4 & \cellcolor[HTML]{F2BF87}17.4 & \cellcolor[HTML]{EBEAD0}23.4 & \cellcolor[HTML]{ECE8C9}\textbf{20.1} & \cellcolor[HTML]{F4CD90}18.6 & \cellcolor[HTML]{D8E4E8}19.1 & \cellcolor[HTML]{F0A77A}15.3 & \cellcolor[HTML]{DBE6E8}\textbf{21.7} & \cellcolor[HTML]{EFF2EB}22.4 & \cellcolor[HTML]{ECE9CC}19.6 & \cellcolor[HTML]{EBEAD4}20.3 \\
 & English & \cellcolor[HTML]{B7D2ED}\textbf{21.8} & \cellcolor[HTML]{95C5DE}\textbf{20.6} & \cellcolor[HTML]{B6D2ED}\textbf{23.6} & \cellcolor[HTML]{D1E0E8}22.6 & \cellcolor[HTML]{B8D3ED}\textbf{21.5} & \cellcolor[HTML]{69BD94}\textbf{22.6} & \cellcolor[HTML]{AACCF1}\textbf{25.2} & \cellcolor[HTML]{F3E7B0}19.8 & \cellcolor[HTML]{DEE7E8}\textbf{21.0} & \cellcolor[HTML]{96C5DF}\textbf{20.7} & \cellcolor[HTML]{A9CCF1}\textbf{19.8} & \cellcolor[HTML]{EBEBD9}20.8 & \cellcolor[HTML]{CADCE9}\textbf{23.2} & \cellcolor[HTML]{97C5E0}\textbf{22.4} & \cellcolor[HTML]{A3C9ED}\textbf{21.8} \\
\bottomrule
\end{tabular}%
}
\end{minipage}
\end{table}

\begin{table}[!htbp]
\centering
\begin{minipage}[t]{0.49\textwidth}
\centering
\caption{English-content tasks from Multilingual Reasoning Gym, not seen during training -- Avg@8 percentage-point improvement over the base model \texttt{Qwen3-8B-Base}}
\label{tab:appendix_crosslingual_qwen3_8b_base_5_english_content_add}
\tiny
\setlength{\tabcolsep}{2pt}
\renewcommand{\arraystretch}{0.92}
\newcommand{\MatrixResizeWidth}{\linewidth}
\resizebox{\MatrixResizeWidth}{!}{%
\begin{tabular}{llrrrrrrrrrrrrrrVr}
\toprule
Train & Reward & en & zh & de & fr & es & ru & ja & th & bn & te & sw & it & ko & pt & Avg \\
\midrule
\multirow{1}{*}{en} & English & \cellcolor[HTML]{6DBE9E}\textbf{33.3} & \cellcolor[HTML]{74BEAD}\textbf{34.8} & \cellcolor[HTML]{60BE82}\textbf{37.6} & \cellcolor[HTML]{6ABE97}\textbf{38.6} & \cellcolor[HTML]{8BC2D1}\textbf{34.8} & \cellcolor[HTML]{78BEB2}\textbf{36.5} & \cellcolor[HTML]{8AC1CF}\textbf{36.1} & \cellcolor[HTML]{90C3D7}\textbf{37.2} & \cellcolor[HTML]{83C1C5}\textbf{37.6} & \cellcolor[HTML]{87C1CC}\textbf{32.6} & \cellcolor[HTML]{E2EAE9}\textbf{23.3} & \cellcolor[HTML]{8AC2D0}\textbf{29.6} & \cellcolor[HTML]{74BEAD}\textbf{40.4} & \cellcolor[HTML]{73BEA9}\textbf{37.5} & \cellcolor[HTML]{7CBFBA}\textbf{35.0} \\
\midrule
\multirow{2}{*}{zh} & Native & \cellcolor[HTML]{DDE7E8}26.2 & \cellcolor[HTML]{9AC6E4}31.5 & \cellcolor[HTML]{96C5E0}33.3 & \cellcolor[HTML]{B9D4EC}32.8 & \cellcolor[HTML]{EAEFEC}29.7 & \cellcolor[HTML]{CFDFE8}29.3 & \cellcolor[HTML]{A4CAED}34.0 & \cellcolor[HTML]{F1F4F0}32.4 & \cellcolor[HTML]{ADCDF0}34.4 & \cellcolor[HTML]{C1D7EB}28.8 & \cellcolor[HTML]{EDE8C4}20.8 & \cellcolor[HTML]{D8E4E8}25.2 & \cellcolor[HTML]{EDEFE5}32.9 & \cellcolor[HTML]{CBDDE9}31.7 & \cellcolor[HTML]{C8DBE9}30.2 \\
 & English & \cellcolor[HTML]{94C4DD}\textbf{29.9} & \cellcolor[HTML]{77BFB0}\textbf{34.7} & \cellcolor[HTML]{6EBEA0}\textbf{36.4} & \cellcolor[HTML]{8CC2D2}\textbf{35.7} & \cellcolor[HTML]{8DC3D4}\textbf{34.6} & \cellcolor[HTML]{82C0C4}\textbf{35.3} & \cellcolor[HTML]{8CC2D2}\textbf{35.9} & \cellcolor[HTML]{A2C9EB}\textbf{35.8} & \cellcolor[HTML]{6EBD9F}\textbf{39.5} & \cellcolor[HTML]{71BEA5}\textbf{34.7} & \cellcolor[HTML]{DCE6E8}\textbf{23.5} & \cellcolor[HTML]{9CC7E6}\textbf{28.1} & \cellcolor[HTML]{A0C8E9}\textbf{37.1} & \cellcolor[HTML]{82C0C4}\textbf{36.1} & \cellcolor[HTML]{86C1CB}\textbf{34.1} \\
\bottomrule
\end{tabular}%
}
\end{minipage}
\hfill
\begin{minipage}[t]{0.49\textwidth}
\centering
\caption{Very difficult tasks from Multilingual Reasoning Gym, not seen during training -- Avg@8 percentage-point improvement over the base model \texttt{Qwen3-8B-Base}}
\label{tab:appendix_crosslingual_qwen3_8b_base_6_difficulty_addback}
\tiny
\setlength{\tabcolsep}{2pt}
\renewcommand{\arraystretch}{0.92}
\newcommand{\MatrixResizeWidth}{\linewidth}
\resizebox{\MatrixResizeWidth}{!}{%
\begin{tabular}{llrrrrrrrrrrrrrrVr}
\toprule
Train & Reward & en & zh & de & fr & es & ru & ja & th & bn & te & sw & it & ko & pt & Avg \\
\midrule
\multirow{1}{*}{en} & English & \cellcolor[HTML]{F1E7B4}\textbf{1.7} & \cellcolor[HTML]{F6E6A8}\textbf{1.4} & \cellcolor[HTML]{F2BD86}\textbf{1.0} & \cellcolor[HTML]{ECE9CC}\textbf{1.1} & \cellcolor[HTML]{EBF0ED}\textbf{1.7} & \cellcolor[HTML]{ECE9CC}\textbf{1.7} & \cellcolor[HTML]{ED906F}\textbf{0.7} & \cellcolor[HTML]{E5ECEA}\textbf{2.0} & \cellcolor[HTML]{EEE7BF}\textbf{1.1} & \cellcolor[HTML]{F4D193}\textbf{0.6} & \cellcolor[HTML]{E0E9E9}\textbf{0.8} & \cellcolor[HTML]{ECEDDF}\textbf{1.7} & \cellcolor[HTML]{EEF0E7}\textbf{1.6} & \cellcolor[HTML]{F5E7AB}\textbf{1.6} & \cellcolor[HTML]{EBE9CE}\textbf{1.3} \\
\midrule
\multirow{2}{*}{zh} & Native & \cellcolor[HTML]{EE9572}0.9 & \cellcolor[HTML]{F3CB90}\textbf{1.1} & \cellcolor[HTML]{F4D092}1.1 & \cellcolor[HTML]{F0B07F}0.5 & \cellcolor[HTML]{F2C28A}0.8 & \cellcolor[HTML]{EE9270}0.8 & \cellcolor[HTML]{F0AE7E}0.9 & \cellcolor[HTML]{EC8C6E}0.7 & \cellcolor[HTML]{F4D193}0.8 & \cellcolor[HTML]{EE9E76}0.4 & \cellcolor[HTML]{EB7B68}0.2 & \cellcolor[HTML]{F3C78D}1.0 & \cellcolor[HTML]{EC7E69}0.6 & \cellcolor[HTML]{E8645F}0.7 & \cellcolor[HTML]{F0AB7C}0.8 \\
 & English & \cellcolor[HTML]{F4CD90}\textbf{1.3} & \cellcolor[HTML]{F0B381}0.9 & \cellcolor[HTML]{F6E2A1}\textbf{1.3} & \cellcolor[HTML]{EBEAD3}\textbf{1.2} & \cellcolor[HTML]{EFE7BB}\textbf{1.3} & \cellcolor[HTML]{F4E7AD}\textbf{1.5} & \cellcolor[HTML]{EFE7BD}\textbf{1.6} & \cellcolor[HTML]{EBEAD2}\textbf{1.7} & \cellcolor[HTML]{F1F4F0}\textbf{1.4} & \cellcolor[HTML]{EDF0E6}\textbf{0.9} & \cellcolor[HTML]{99C5E2}\textbf{1.0} & \cellcolor[HTML]{F7E7A6}\textbf{1.3} & \cellcolor[HTML]{EDE8C3}\textbf{1.4} & \cellcolor[HTML]{EE9170}\textbf{1.0} & \cellcolor[HTML]{EDE8C3}\textbf{1.3} \\
\bottomrule
\end{tabular}%
}
\end{minipage}
\end{table}

\begin{table}[!htbp]
\centering
\begin{minipage}[t]{0.49\textwidth}
\centering
\caption{Easy Math (MGSM + PolyMath low) -- Avg@8 percentage-point improvement over the base model \texttt{Qwen3-8B-Base}; $\dagger$: includes only PolyMath low because MGSM is unavailable for this language.}
\label{tab:appendix_crosslingual_qwen3_8b_base_7_appendix_easy_math_macro}
\tiny
\setlength{\tabcolsep}{2pt}
\renewcommand{\arraystretch}{0.92}
\newcommand{\MatrixResizeWidth}{\linewidth}
\resizebox{\MatrixResizeWidth}{!}{%
\begin{tabular}{llrrrrrrrrrrrrrrrrrrVr}
\toprule
Train & Reward & en & zh & de & fr & es & ru & ja & th & bn & te & sw & it$^{\dagger}$ & ko$^{\dagger}$ & pt$^{\dagger}$ & ar$^{\dagger}$ & id$^{\dagger}$ & ms$^{\dagger}$ & vi$^{\dagger}$ & Avg \\
\midrule
\multirow{1}{*}{en} & English & \cellcolor[HTML]{74BEAD}\textbf{40.4} & \cellcolor[HTML]{7CBFB9}\textbf{45.3} & \cellcolor[HTML]{67BD90}\textbf{50.6} & \cellcolor[HTML]{78BEB2}\textbf{47.9} & \cellcolor[HTML]{5CBE78}\underline{\textbf{45.0}} & \cellcolor[HTML]{6BBE99}\textbf{45.9} & \cellcolor[HTML]{8DC3D4}\textbf{52.5} & \cellcolor[HTML]{88C1CD}\textbf{54.5} & \cellcolor[HTML]{84C0C7}\textbf{54.8} & \cellcolor[HTML]{96C5E0}\textbf{51.4} & \cellcolor[HTML]{8CC2D3}\textbf{45.6} & \cellcolor[HTML]{8BC2D1}\textbf{49.7} & \cellcolor[HTML]{77BFAF}\textbf{58.7} & \cellcolor[HTML]{81C0C3}\textbf{43.0} & \cellcolor[HTML]{8BC2D2}\textbf{55.7} & \cellcolor[HTML]{73BEA9}\textbf{53.0} & \cellcolor[HTML]{7ABFB7}\textbf{54.1} & \cellcolor[HTML]{80C0C0}\textbf{43.9} & \cellcolor[HTML]{75BEAE}\textbf{49.6} \\
\midrule
\multirow{2}{*}{zh} & Native & \cellcolor[HTML]{C0D7EB}37.7 & \cellcolor[HTML]{63BE89}\textbf{46.9} & \cellcolor[HTML]{8CC2D2}48.1 & \cellcolor[HTML]{A8CBF1}45.3 & \cellcolor[HTML]{B9D4EC}40.2 & \cellcolor[HTML]{A4CAED}42.2 & \cellcolor[HTML]{7DBFBC}53.7 & \cellcolor[HTML]{8EC2D5}54.2 & \cellcolor[HTML]{EFF1EA}49.6 & \cellcolor[HTML]{F3F5F2}47.4 & \cellcolor[HTML]{F2BB85}36.1 & \cellcolor[HTML]{C3D8EA}47.3 & \cellcolor[HTML]{A3C9ED}56.0 & \cellcolor[HTML]{B4D1ED}40.6 & \cellcolor[HTML]{D0DFE8}52.4 & \cellcolor[HTML]{BAD4EC}49.5 & \cellcolor[HTML]{CFDFE8}50.3 & \cellcolor[HTML]{92C4DA}42.8 & \cellcolor[HTML]{A7CBF0}46.7 \\
\bottomrule
\end{tabular}%
}
\end{minipage}
\hfill
\begin{minipage}[t]{0.49\textwidth}
\centering
\caption{Hard Math (PolyMath medium, high, and top) -- Avg@8 percentage-point improvement over the base model \texttt{Qwen3-8B-Base}}
\label{tab:appendix_crosslingual_qwen3_8b_base_8_appendix_hard_math_macro}
\tiny
\setlength{\tabcolsep}{2pt}
\renewcommand{\arraystretch}{0.92}
\newcommand{\MatrixResizeWidth}{\linewidth}
\resizebox{\MatrixResizeWidth}{!}{%
\begin{tabular}{llrrrrrrrrrrrrrrrrrrVr}
\toprule
Train & Reward & en & zh & de & fr & es & ru & ja & th & bn & te & sw & it & ko & pt & ar & id & ms & vi & Avg \\
\midrule
\multirow{1}{*}{en} & English & \cellcolor[HTML]{6EBEA0}\textbf{15.6} & \cellcolor[HTML]{6CBE9C}\textbf{16.5} & \cellcolor[HTML]{62BE88}\textbf{18.5} & \cellcolor[HTML]{85C1C8}\textbf{17.0} & \cellcolor[HTML]{82C0C4}\textbf{16.7} & \cellcolor[HTML]{86C1CA}\textbf{16.3} & \cellcolor[HTML]{B9D4EC}\textbf{16.3} & \cellcolor[HTML]{6EBEA0}\textbf{18.1} & \cellcolor[HTML]{B8D3ED}\textbf{16.0} & \cellcolor[HTML]{9EC8E7}\textbf{15.5} & \cellcolor[HTML]{87C1CC}\textbf{13.5} & \cellcolor[HTML]{79BFB3}\textbf{16.7} & \cellcolor[HTML]{84C1C7}\textbf{18.5} & \cellcolor[HTML]{73BEAB}\textbf{17.3} & \cellcolor[HTML]{9BC6E4}\textbf{17.7} & \cellcolor[HTML]{68BE93}\textbf{17.9} & \cellcolor[HTML]{7DBFBB}\textbf{17.4} & \cellcolor[HTML]{74BEAD}\textbf{16.8} & \cellcolor[HTML]{7EC0BC}\textbf{16.8} \\
\midrule
\multirow{2}{*}{zh} & Native & \cellcolor[HTML]{EDE8C8}12.0 & \cellcolor[HTML]{F5D797}12.3 & \cellcolor[HTML]{E5ECEA}15.4 & \cellcolor[HTML]{F5E6AB}13.6 & \cellcolor[HTML]{F4E7AD}13.5 & \cellcolor[HTML]{EBEBD9}13.4 & \cellcolor[HTML]{E7EDEB}15.5 & \cellcolor[HTML]{F5D495}13.5 & \cellcolor[HTML]{EEE7C2}13.8 & \cellcolor[HTML]{E3EBEA}13.4 & \cellcolor[HTML]{ECEEE3}11.1 & \cellcolor[HTML]{F5D797}12.9 & \cellcolor[HTML]{F2C58B}13.6 & \cellcolor[HTML]{EFE7BC}13.3 & \cellcolor[HTML]{F3E7B0}14.7 & \cellcolor[HTML]{F0F3ED}14.0 & \cellcolor[HTML]{EBEBD7}13.3 & \cellcolor[HTML]{F2E7B4}13.1 & \cellcolor[HTML]{F5E7AC}13.5 \\
\bottomrule
\end{tabular}%
}
\end{minipage}
\end{table}

\FloatBarrier
\subsection{Qwen3-1.7B}

\begin{table}[!htbp]
\centering
\begin{minipage}[t]{0.49\textwidth}
\centering
\caption{Tasks seen during training from Multilingual Reasoning Gym -- Avg@8 percentage-point improvement over the base model \texttt{Qwen3-1.7B}}
\label{tab:appendix_crosslingual_qwen3_1_7b_1_train_filtered_noheldout}
\tiny
\setlength{\tabcolsep}{2pt}
\renewcommand{\arraystretch}{0.92}
\newcommand{\MatrixResizeWidth}{\linewidth}
\resizebox{\MatrixResizeWidth}{!}{%
\begin{tabular}{llrrrrrrrrrrrrrrVr}
\toprule
Train & Reward & en & zh & de & fr & es & ru & ja & th & bn & te & sw & it & ko & pt & Avg \\
\midrule
\multirow{1}{*}{en} & English & \cellcolor[HTML]{5CBE78}\underline{\textbf{19.1}} & \cellcolor[HTML]{F6E1A1}\textbf{10.6} & \cellcolor[HTML]{9CC7E6}\textbf{12.5} & \cellcolor[HTML]{9AC6E4}\textbf{11.8} & \cellcolor[HTML]{E5ECEA}\textbf{11.8} & \cellcolor[HTML]{F0AF7E}\textbf{7.9} & \cellcolor[HTML]{EDE8C6}\textbf{10.0} & \cellcolor[HTML]{F2F4EF}\textbf{13.3} & \cellcolor[HTML]{F0A77A}\textbf{13.0} & \cellcolor[HTML]{ED9070}\textbf{10.1} & \cellcolor[HTML]{EA6C62}\textbf{12.3} & \cellcolor[HTML]{72BEA8}\textbf{13.9} & \cellcolor[HTML]{70BEA4}\textbf{14.5} & \cellcolor[HTML]{9AC6E3}\textbf{13.8} & \cellcolor[HTML]{8CC2D2}\textbf{12.5} \\
\midrule
\multirow{2}{*}{zh} & Native & \cellcolor[HTML]{F4D092}12.3 & \cellcolor[HTML]{87C1CC}16.1 & \cellcolor[HTML]{EFF1EA}9.9 & \cellcolor[HTML]{ECF0ED}10.1 & \cellcolor[HTML]{ECEDDF}11.1 & \cellcolor[HTML]{EEF0E7}11.5 & \cellcolor[HTML]{F5D99A}9.0 & \cellcolor[HTML]{F2BF87}10.2 & \cellcolor[HTML]{F0B07F}13.2 & \cellcolor[HTML]{EA7063}8.9 & \cellcolor[HTML]{EC896D}13.2 & \cellcolor[HTML]{BBD5EC}\textbf{11.9} & \cellcolor[HTML]{67BD91}14.8 & \cellcolor[HTML]{D0E0E8}\textbf{12.5} & \cellcolor[HTML]{BDD5EB}11.8 \\
 & English & \cellcolor[HTML]{D4E2E8}\textbf{15.2} & \cellcolor[HTML]{5CBE78}\underline{\textbf{18.4}} & \cellcolor[HTML]{D5E2E8}\textbf{10.9} & \cellcolor[HTML]{B4D1EE}\textbf{11.1} & \cellcolor[HTML]{EEF2EE}\textbf{11.6} & \cellcolor[HTML]{C5DAE9}\textbf{13.0} & \cellcolor[HTML]{F1F4F0}\textbf{11.0} & \cellcolor[HTML]{F6E7A8}\textbf{11.6} & \cellcolor[HTML]{F5DD9C}\textbf{14.5} & \cellcolor[HTML]{F0B280}\textbf{11.4} & \cellcolor[HTML]{EFA378}\textbf{14.1} & \cellcolor[HTML]{BCD5EC}11.9 & \cellcolor[HTML]{5CBE78}\underline{\textbf{15.2}} & \cellcolor[HTML]{D7E4E8}12.4 & \cellcolor[HTML]{6FBEA0}\textbf{13.0} \\
\bottomrule
\end{tabular}%
}
\end{minipage}
\hfill
\begin{minipage}[t]{0.49\textwidth}
\centering
\caption{Macro average over all unseen tasks -- Avg@8 percentage-point improvement over the base model \texttt{Qwen3-1.7B}}
\label{tab:appendix_crosslingual_qwen3_1_7b_2_appendix_unseen_task_groups_macro}
\tiny
\setlength{\tabcolsep}{2pt}
\renewcommand{\arraystretch}{0.92}
\newcommand{\MatrixResizeWidth}{\linewidth}
\resizebox{\MatrixResizeWidth}{!}{%
\begin{tabular}{llrrrrrrrrrrrVr}
\toprule
Train & Reward & en & zh & de & fr & es & ru & ja & th & bn & te & sw & Avg \\
\midrule
\multirow{1}{*}{en} & English & \cellcolor[HTML]{5CBE78}\underline{\textbf{2.3}} & \cellcolor[HTML]{5CBE78}\underline{\textbf{3.6}} & \cellcolor[HTML]{5CBE78}\underline{\textbf{4.9}} & \cellcolor[HTML]{5CBE78}\underline{\textbf{3.5}} & \cellcolor[HTML]{5CBE78}\underline{\textbf{4.2}} & \cellcolor[HTML]{75BEAE}\textbf{1.7} & \cellcolor[HTML]{5CBE78}\underline{\textbf{3.4}} & \cellcolor[HTML]{5CBE78}\underline{\textbf{2.4}} & \cellcolor[HTML]{5CBE78}\underline{\textbf{4.0}} & \cellcolor[HTML]{5CBE78}\underline{\textbf{2.0}} & \cellcolor[HTML]{5CBE78}\underline{\textbf{2.2}} & \cellcolor[HTML]{5CBE78}\underline{\textbf{3.1}} \\
\midrule
\multirow{2}{*}{zh} & Native & \cellcolor[HTML]{ED9170}-7.3 & \cellcolor[HTML]{EDE8C3}-1.6 & \cellcolor[HTML]{EFA378}-2.8 & \cellcolor[HTML]{F0A87A}-4.9 & \cellcolor[HTML]{F0B481}-4.7 & \cellcolor[HTML]{F3CB90}-3.1 & \cellcolor[HTML]{F0A67A}-4.9 & \cellcolor[HTML]{E85E5D}-5.0 & \cellcolor[HTML]{E85C5C}-2.4 & \cellcolor[HTML]{E85C5C}-4.5 & \cellcolor[HTML]{EEA277}-1.2 & \cellcolor[HTML]{EC896D}-3.9 \\
 & English & \cellcolor[HTML]{EBECDD}\textbf{-3.6} & \cellcolor[HTML]{A6CBEF}\textbf{0.8} & \cellcolor[HTML]{BAD4EC}\textbf{1.6} & \cellcolor[HTML]{C8DBE9}\textbf{-0.5} & \cellcolor[HTML]{ECF0EE}\textbf{-1.1} & \cellcolor[HTML]{8AC2CF}\textbf{1.0} & \cellcolor[HTML]{ECF0ED}\textbf{-1.3} & \cellcolor[HTML]{D0E0E8}\textbf{-0.7} & \cellcolor[HTML]{D8E4E8}\textbf{1.2} & \cellcolor[HTML]{A8CBF1}\textbf{-0.1} & \cellcolor[HTML]{DCE7E8}\textbf{0.5} & \cellcolor[HTML]{D7E3E8}\textbf{-0.2} \\
\bottomrule
\end{tabular}%
}
\end{minipage}
\end{table}

\begin{table}[!htbp]
\centering
\begin{minipage}[t]{0.49\textwidth}
\centering
\caption{Average over all unseen tasks from Multilingual Reasoning Gym -- Avg@8 percentage-point improvement over the base model \texttt{Qwen3-1.7B}}
\label{tab:appendix_crosslingual_qwen3_1_7b_3_appendix_mrg_unseen_macro}
\tiny
\setlength{\tabcolsep}{2pt}
\renewcommand{\arraystretch}{0.92}
\newcommand{\MatrixResizeWidth}{\linewidth}
\resizebox{\MatrixResizeWidth}{!}{%
\begin{tabular}{llrrrrrrrrrrrrrrVr}
\toprule
Train & Reward & en & zh & de & fr & es & ru & ja & th & bn & te & sw & it & ko & pt & Avg \\
\midrule
\multirow{1}{*}{en} & English & \cellcolor[HTML]{74BFAC}\textbf{9.3} & \cellcolor[HTML]{C3D9EA}\textbf{9.7} & \cellcolor[HTML]{67BD90}\textbf{10.7} & \cellcolor[HTML]{83C0C6}\textbf{9.6} & \cellcolor[HTML]{71BEA5}\textbf{11.6} & \cellcolor[HTML]{F2C48B}\textbf{7.1} & \cellcolor[HTML]{5CBE78}\underline{\textbf{7.4}} & \cellcolor[HTML]{91C3D9}\textbf{7.7} & \cellcolor[HTML]{5CBE78}\underline{\textbf{8.3}} & \cellcolor[HTML]{F3F6F2}\textbf{5.8} & \cellcolor[HTML]{CBDDE9}\textbf{6.0} & \cellcolor[HTML]{7BBFB8}\textbf{11.9} & \cellcolor[HTML]{5CBE78}\textbf{8.7} & \cellcolor[HTML]{70BEA3}\textbf{11.7} & \cellcolor[HTML]{7DBFBC}\textbf{9.0} \\
\midrule
\multirow{2}{*}{zh} & Native & \cellcolor[HTML]{EEF0E7}5.7 & \cellcolor[HTML]{A9CCF1}10.2 & \cellcolor[HTML]{88C1CE}9.3 & \cellcolor[HTML]{BDD5EB}8.5 & \cellcolor[HTML]{93C4DB}10.4 & \cellcolor[HTML]{80C0C0}12.5 & \cellcolor[HTML]{9AC6E3}5.0 & \cellcolor[HTML]{C9DCE9}6.8 & \cellcolor[HTML]{82C0C3}7.4 & \cellcolor[HTML]{EC8C6E}3.8 & \cellcolor[HTML]{C6DAEA}6.0 & \cellcolor[HTML]{9CC7E6}10.8 & \cellcolor[HTML]{98C6E1}6.9 & \cellcolor[HTML]{74BEAD}11.5 & \cellcolor[HTML]{99C6E3}8.2 \\
 & English & \cellcolor[HTML]{7DBFBB}\textbf{9.0} & \cellcolor[HTML]{5CBE78}\underline{\textbf{13.7}} & \cellcolor[HTML]{6ABE97}\textbf{10.5} & \cellcolor[HTML]{62BE88}\textbf{10.6} & \cellcolor[HTML]{6DBD9F}\textbf{11.7} & \cellcolor[HTML]{62BD86}\textbf{13.7} & \cellcolor[HTML]{65BD8C}\textbf{7.0} & \cellcolor[HTML]{87C1CC}\textbf{7.9} & \cellcolor[HTML]{7EC0BC}\textbf{7.5} & \cellcolor[HTML]{87C1CB}\textbf{7.2} & \cellcolor[HTML]{97C5E0}\textbf{6.4} & \cellcolor[HTML]{5CBE78}\underline{\textbf{13.0}} & \cellcolor[HTML]{5FBE7F}\textbf{8.6} & \cellcolor[HTML]{5CBE78}\underline{\textbf{12.5}} & \cellcolor[HTML]{5CBE78}\underline{\textbf{10.0}} \\
\bottomrule
\end{tabular}%
}
\end{minipage}
\hfill
\begin{minipage}[t]{0.49\textwidth}
\centering
\caption{Main heldout tasks from Multilingual Reasoning Gym -- Avg@8 percentage-point improvement over the base model \texttt{Qwen3-1.7B}}
\label{tab:appendix_crosslingual_qwen3_1_7b_4_heldout_filtered}
\tiny
\setlength{\tabcolsep}{2pt}
\renewcommand{\arraystretch}{0.92}
\newcommand{\MatrixResizeWidth}{\linewidth}
\resizebox{\MatrixResizeWidth}{!}{%
\begin{tabular}{llrrrrrrrrrrrrrrVr}
\toprule
Train & Reward & en & zh & de & fr & es & ru & ja & th & bn & te & sw & it & ko & pt & Avg \\
\midrule
\multirow{1}{*}{en} & English & \cellcolor[HTML]{5CBE78}\underline{\textbf{19.8}} & \cellcolor[HTML]{EDEFE5}\textbf{15.2} & \cellcolor[HTML]{5CBE78}\underline{\textbf{20.2}} & \cellcolor[HTML]{87C1CC}\textbf{15.4} & \cellcolor[HTML]{72BEA9}\textbf{21.4} & \cellcolor[HTML]{F2C289}\textbf{10.2} & \cellcolor[HTML]{5CBE78}\underline{\textbf{14.1}} & \cellcolor[HTML]{60BE82}\textbf{15.9} & \cellcolor[HTML]{7FC0C0}\textbf{13.0} & \cellcolor[HTML]{F7E5A4}\textbf{9.1} & \cellcolor[HTML]{CCDEE8}\textbf{12.2} & \cellcolor[HTML]{62BD87}\textbf{17.2} & \cellcolor[HTML]{6CBE9B}\textbf{14.1} & \cellcolor[HTML]{73BEA9}\textbf{22.5} & \cellcolor[HTML]{74BFAC}\textbf{15.7} \\
\midrule
\multirow{2}{*}{zh} & Native & \cellcolor[HTML]{EEF0E7}10.1 & \cellcolor[HTML]{F4D193}13.0 & \cellcolor[HTML]{A9CCF1}16.0 & \cellcolor[HTML]{ECEEE1}12.7 & \cellcolor[HTML]{D4E2E8}18.9 & \cellcolor[HTML]{99C6E2}16.4 & \cellcolor[HTML]{E5ECEA}9.2 & \cellcolor[HTML]{E4EBEA}11.8 & \cellcolor[HTML]{ABCDF1}\textbf{11.3} & \cellcolor[HTML]{EDE8C3}9.6 & \cellcolor[HTML]{B2D0EE}12.8 & \cellcolor[HTML]{F2F4EF}13.4 & \cellcolor[HTML]{BFD6EB}10.4 & \cellcolor[HTML]{9DC7E6}20.8 & \cellcolor[HTML]{D6E3E8}13.3 \\
 & English & \cellcolor[HTML]{6EBEA0}\textbf{18.1} & \cellcolor[HTML]{5CBE78}\underline{\textbf{20.9}} & \cellcolor[HTML]{7CBFB9}\textbf{18.3} & \cellcolor[HTML]{60BE83}\textbf{16.9} & \cellcolor[HTML]{6CBE9C}\textbf{21.7} & \cellcolor[HTML]{5CBE78}\underline{\textbf{19.8}} & \cellcolor[HTML]{6CBE9C}\textbf{13.2} & \cellcolor[HTML]{9EC8E8}\textbf{13.4} & \cellcolor[HTML]{DFE8E9}10.2 & \cellcolor[HTML]{DEE8E9}\textbf{11.0} & \cellcolor[HTML]{88C2CD}\textbf{14.1} & \cellcolor[HTML]{5CBE78}\underline{\textbf{17.5}} & \cellcolor[HTML]{6ABE97}\textbf{14.2} & \cellcolor[HTML]{5CBE78}\underline{\textbf{23.6}} & \cellcolor[HTML]{5CBE78}\underline{\textbf{16.6}} \\
\bottomrule
\end{tabular}%
}
\end{minipage}
\end{table}

\begin{table}[!htbp]
\centering
\begin{minipage}[t]{0.49\textwidth}
\centering
\caption{English-content tasks from Multilingual Reasoning Gym, not seen during training -- Avg@8 percentage-point improvement over the base model \texttt{Qwen3-1.7B}}
\label{tab:appendix_crosslingual_qwen3_1_7b_5_english_content_add}
\tiny
\setlength{\tabcolsep}{2pt}
\renewcommand{\arraystretch}{0.92}
\newcommand{\MatrixResizeWidth}{\linewidth}
\resizebox{\MatrixResizeWidth}{!}{%
%
%
}
\end{minipage}
\hfill
\begin{minipage}[t]{0.49\textwidth}
\centering
\caption{Very difficult tasks from Multilingual Reasoning Gym, not seen during training -- Avg@8 percentage-point improvement over the base model \texttt{Qwen3-1.7B}}
\label{tab:appendix_crosslingual_qwen3_1_7b_6_difficulty_addback}
\tiny
\setlength{\tabcolsep}{2pt}
\renewcommand{\arraystretch}{0.92}
\newcommand{\MatrixResizeWidth}{\linewidth}
\resizebox{\MatrixResizeWidth}{!}{%
%
%
}
\end{minipage}
\end{table}

\begin{table}[!htbp]
\centering
\begin{minipage}[t]{0.49\textwidth}
\centering
\caption{Easy Math (MGSM + PolyMath low) -- Avg@8 percentage-point improvement over the base model \texttt{Qwen3-1.7B}; $\dagger$: includes only PolyMath low because MGSM is unavailable for this language.}
\label{tab:appendix_crosslingual_qwen3_1_7b_7_appendix_easy_math_macro}
\tiny
\setlength{\tabcolsep}{2pt}
\renewcommand{\arraystretch}{0.92}
\newcommand{\MatrixResizeWidth}{\linewidth}
\resizebox{\MatrixResizeWidth}{!}{%
%
%
}
\end{minipage}
\hfill
\begin{minipage}[t]{0.49\textwidth}
\centering
\caption{Hard Math (PolyMath medium, high, and top) -- Avg@8 percentage-point improvement over the base model \texttt{Qwen3-1.7B}}
\label{tab:appendix_crosslingual_qwen3_1_7b_8_appendix_hard_math_macro}
\tiny
\setlength{\tabcolsep}{2pt}
\renewcommand{\arraystretch}{0.92}
\newcommand{\MatrixResizeWidth}{\linewidth}
\resizebox{\MatrixResizeWidth}{!}{%
%
%
}
\end{minipage}
\end{table}

\FloatBarrier
\subsection{Qwen3-4B}

\paragraph{Ineffective rewards.}
For Qwen3-4B, we mark some reward rows as ineffective. ``Ineffective'' means the base model had at least 2.5\% support for the target reasoning language, but the trained model did not converge to that language.
\begin{itemize}
\item For training in Chinese (zh), English reward is marked \textit{ineffective}. On Chinese (zh)-prompt rollouts, the base model produced 93.7\% Chinese (zh), 2.7\% English, and 3.6\% other languages; the trained model produced 93.0\% Chinese (zh), 2.7\% English, and 4.3\% other languages.
\end{itemize}

\begin{table}[!htbp]
\centering
\begin{minipage}[t]{0.49\textwidth}
\centering
\caption{Tasks seen during training from Multilingual Reasoning Gym -- Avg@8 percentage-point improvement over the base model \texttt{Qwen3-4B}}
\label{tab:appendix_crosslingual_qwen3_4b_1_train_filtered_noheldout}
\tiny
\setlength{\tabcolsep}{2pt}
\renewcommand{\arraystretch}{0.92}
\newcommand{\MatrixResizeWidth}{\linewidth}
\resizebox{\MatrixResizeWidth}{!}{%
%
}
\end{minipage}
\hfill
\begin{minipage}[t]{0.49\textwidth}
\centering
\caption{Macro average over all unseen tasks -- Avg@8 percentage-point improvement over the base model \texttt{Qwen3-4B}}
\label{tab:appendix_crosslingual_qwen3_4b_2_appendix_unseen_task_groups_macro}
\tiny
\setlength{\tabcolsep}{2pt}
\renewcommand{\arraystretch}{0.92}
\newcommand{\MatrixResizeWidth}{\linewidth}
\resizebox{\MatrixResizeWidth}{!}{%
%
%
}
\end{minipage}
\end{table}

\begin{table}[!htbp]
\centering
\begin{minipage}[t]{0.49\textwidth}
\centering
\caption{Average over all unseen tasks from Multilingual Reasoning Gym -- Avg@8 percentage-point improvement over the base model \texttt{Qwen3-4B}}
\label{tab:appendix_crosslingual_qwen3_4b_3_appendix_mrg_unseen_macro}
\tiny
\setlength{\tabcolsep}{2pt}
\renewcommand{\arraystretch}{0.92}
\newcommand{\MatrixResizeWidth}{\linewidth}
\resizebox{\MatrixResizeWidth}{!}{%
%
%
}
\end{minipage}
\hfill
\begin{minipage}[t]{0.49\textwidth}
\centering
\caption{Main heldout tasks from Multilingual Reasoning Gym -- Avg@8 percentage-point improvement over the base model \texttt{Qwen3-4B}}
\label{tab:appendix_crosslingual_qwen3_4b_4_heldout_filtered}
\tiny
\setlength{\tabcolsep}{2pt}
\renewcommand{\arraystretch}{0.92}
\newcommand{\MatrixResizeWidth}{\linewidth}
\resizebox{\MatrixResizeWidth}{!}{%
%
%
}
\end{minipage}
\end{table}

\begin{table}[!htbp]
\centering
\begin{minipage}[t]{0.49\textwidth}
\centering
\caption{English-content tasks from Multilingual Reasoning Gym, not seen during training -- Avg@8 percentage-point improvement over the base model \texttt{Qwen3-4B}}
\label{tab:appendix_crosslingual_qwen3_4b_5_english_content_add}
\tiny
\setlength{\tabcolsep}{2pt}
\renewcommand{\arraystretch}{0.92}
\newcommand{\MatrixResizeWidth}{\linewidth}
\resizebox{\MatrixResizeWidth}{!}{%
%
%
}
\end{minipage}
\hfill
\begin{minipage}[t]{0.49\textwidth}
\centering
\caption{Very difficult tasks from Multilingual Reasoning Gym, not seen during training -- Avg@8 percentage-point improvement over the base model \texttt{Qwen3-4B}}
\label{tab:appendix_crosslingual_qwen3_4b_6_difficulty_addback}
\tiny
\setlength{\tabcolsep}{2pt}
\renewcommand{\arraystretch}{0.92}
\newcommand{\MatrixResizeWidth}{\linewidth}
\resizebox{\MatrixResizeWidth}{!}{%
%
%
}
\end{minipage}
\end{table}

\begin{table}[!htbp]
\centering
\begin{minipage}[t]{0.49\textwidth}
\centering
\caption{Easy Math (MGSM + PolyMath low) -- Avg@8 percentage-point improvement over the base model \texttt{Qwen3-4B}; $\dagger$: includes only PolyMath low because MGSM is unavailable for this language.}
\label{tab:appendix_crosslingual_qwen3_4b_7_appendix_easy_math_macro}
\tiny
\setlength{\tabcolsep}{2pt}
\renewcommand{\arraystretch}{0.92}
\newcommand{\MatrixResizeWidth}{\linewidth}
\resizebox{\MatrixResizeWidth}{!}{%
%
%
}
\end{minipage}
\hfill
\begin{minipage}[t]{0.49\textwidth}
\centering
\caption{Hard Math (PolyMath medium, high, and top) -- Avg@8 percentage-point improvement over the base model \texttt{Qwen3-4B}}
\label{tab:appendix_crosslingual_qwen3_4b_8_appendix_hard_math_macro}
\tiny
\setlength{\tabcolsep}{2pt}
\renewcommand{\arraystretch}{0.92}
\newcommand{\MatrixResizeWidth}{\linewidth}
\resizebox{\MatrixResizeWidth}{!}{%
%
%
}
\end{minipage}
\end{table}

\FloatBarrier
\subsection{Qwen3-8B}

\begin{table}[!htbp]
\centering
\begin{minipage}[t]{0.49\textwidth}
\centering
\caption{Tasks seen during training from Multilingual Reasoning Gym -- Avg@8 percentage-point improvement over the base model \texttt{Qwen3-8B}}
\label{tab:appendix_crosslingual_qwen3_8b_1_train_filtered_noheldout}
\tiny
\setlength{\tabcolsep}{2pt}
\renewcommand{\arraystretch}{0.92}
\newcommand{\MatrixResizeWidth}{\linewidth}
\resizebox{\MatrixResizeWidth}{!}{%
%
%
}
\end{minipage}
\hfill
\begin{minipage}[t]{0.49\textwidth}
\centering
\caption{Macro average over all unseen tasks -- Avg@8 percentage-point improvement over the base model \texttt{Qwen3-8B}}
\label{tab:appendix_crosslingual_qwen3_8b_2_appendix_unseen_task_groups_macro}
\tiny
\setlength{\tabcolsep}{2pt}
\renewcommand{\arraystretch}{0.92}
\newcommand{\MatrixResizeWidth}{\linewidth}
\resizebox{\MatrixResizeWidth}{!}{%
%
%
}
\end{minipage}
\end{table}

\begin{table}[!htbp]
\centering
\begin{minipage}[t]{0.49\textwidth}
\centering
\caption{Average over all unseen tasks from Multilingual Reasoning Gym -- Avg@8 percentage-point improvement over the base model \texttt{Qwen3-8B}}
\label{tab:appendix_crosslingual_qwen3_8b_3_appendix_mrg_unseen_macro}
\tiny
\setlength{\tabcolsep}{2pt}
\renewcommand{\arraystretch}{0.92}
\newcommand{\MatrixResizeWidth}{\linewidth}
\resizebox{\MatrixResizeWidth}{!}{%
%
%
}
\end{minipage}
\hfill
\begin{minipage}[t]{0.49\textwidth}
\centering
\caption{Main heldout tasks from Multilingual Reasoning Gym -- Avg@8 percentage-point improvement over the base model \texttt{Qwen3-8B}}
\label{tab:appendix_crosslingual_qwen3_8b_4_heldout_filtered}
\tiny
\setlength{\tabcolsep}{2pt}
\renewcommand{\arraystretch}{0.92}
\newcommand{\MatrixResizeWidth}{\linewidth}
\resizebox{\MatrixResizeWidth}{!}{%
%
%
}
\end{minipage}
\end{table}

\begin{table}[!htbp]
\centering
\begin{minipage}[t]{0.49\textwidth}
\centering
\caption{English-content tasks from Multilingual Reasoning Gym, not seen during training -- Avg@8 percentage-point improvement over the base model \texttt{Qwen3-8B}}
\label{tab:appendix_crosslingual_qwen3_8b_5_english_content_add}
\tiny
\setlength{\tabcolsep}{2pt}
\renewcommand{\arraystretch}{0.92}
\newcommand{\MatrixResizeWidth}{\linewidth}
\resizebox{\MatrixResizeWidth}{!}{%
%
%
}
\end{minipage}
\hfill
\begin{minipage}[t]{0.49\textwidth}
\centering
\caption{Very difficult tasks from Multilingual Reasoning Gym, not seen during training -- Avg@8 percentage-point improvement over the base model \texttt{Qwen3-8B}}
\label{tab:appendix_crosslingual_qwen3_8b_6_difficulty_addback}
\tiny
\setlength{\tabcolsep}{2pt}
\renewcommand{\arraystretch}{0.92}
\newcommand{\MatrixResizeWidth}{\linewidth}
\resizebox{\MatrixResizeWidth}{!}{%
%
%
}
\end{minipage}
\end{table}

\begin{table}[!htbp]
\centering
\begin{minipage}[t]{0.49\textwidth}
\centering
\caption{Easy Math (MGSM + PolyMath low) -- Avg@8 percentage-point improvement over the base model \texttt{Qwen3-8B}; $\dagger$: includes only PolyMath low because MGSM is unavailable for this language.}
\label{tab:appendix_crosslingual_qwen3_8b_7_appendix_easy_math_macro}
\tiny
\setlength{\tabcolsep}{2pt}
\renewcommand{\arraystretch}{0.92}
\newcommand{\MatrixResizeWidth}{\linewidth}
\resizebox{\MatrixResizeWidth}{!}{%
%
%
}
\end{minipage}
\hfill
\begin{minipage}[t]{0.49\textwidth}
\centering
\caption{Hard Math (PolyMath medium, high, and top) -- Avg@8 percentage-point improvement over the base model \texttt{Qwen3-8B}}
\label{tab:appendix_crosslingual_qwen3_8b_8_appendix_hard_math_macro}
\tiny
\setlength{\tabcolsep}{2pt}
\renewcommand{\arraystretch}{0.92}
\newcommand{\MatrixResizeWidth}{\linewidth}
\resizebox{\MatrixResizeWidth}{!}{%
%
%
}
\end{minipage}
\end{table}

\FloatBarrier
\subsection{gemma-3-1b-it}

\paragraph{Ineffective rewards.}
For gemma-3-1b-it, we mark some reward rows as ineffective. ``Ineffective'' means the base model had at least 2.5\% support for the target reasoning language, but the trained model did not converge to that language.
\begin{itemize}
\item For training in Chinese (zh), English reward is marked \textit{ineffective}. On Chinese (zh)-prompt rollouts, the base model produced 5.8\% Chinese (zh), 78.1\% English, and 16.1\% other languages; the trained model produced 80.4\% Chinese (zh), 6.1\% English, and 13.5\% other languages.
\item For training in German (de), English reward is marked \textit{ineffective}. On German (de)-prompt rollouts, the base model produced 25.6\% German (de), 56.9\% English, and 17.4\% other languages; the trained model produced 92.2\% German (de), 4.5\% English, and 3.4\% other languages.
\item For training in French (fr), English reward is marked \textit{ineffective}. On French (fr)-prompt rollouts, the base model produced 26.1\% French (fr), 62.1\% English, and 11.8\% other languages; the trained model produced 89.4\% French (fr), 4.5\% English, and 6.1\% other languages.
\item For training in Japanese (ja), English reward is marked \textit{ineffective}. On Japanese (ja)-prompt rollouts, the base model produced 12.6\% Japanese (ja), 68.1\% English, and 19.2\% other languages; the trained model produced 84.7\% Japanese (ja), 10.7\% English, and 4.6\% other languages.
\item For training in Bengali (bn), English reward is marked \textit{ineffective}. On Bengali (bn)-prompt rollouts, the base model produced 24.2\% Bengali (bn), 62.0\% English, and 13.8\% other languages; the trained model produced 81.3\% Bengali (bn), 3.7\% English, and 15.0\% other languages.
\item For training in Telugu (te), English reward is marked \textit{ineffective}. On Telugu (te)-prompt rollouts, the base model produced 7.2\% Telugu (te), 62.1\% English, and 30.7\% other languages; the trained model produced 86.3\% Telugu (te), 12.0\% English, and 1.7\% other languages.
\end{itemize}

\begin{table}[!htbp]
\centering
\begin{minipage}[t]{0.49\textwidth}
\centering
\caption{Tasks seen during training from Multilingual Reasoning Gym -- Avg@8 percentage-point improvement over the base model \texttt{gemma-3-1b-it}}
\label{tab:appendix_crosslingual_gemma_3_1b_it_1_train_filtered_noheldout}
\tiny
\setlength{\tabcolsep}{2pt}
\renewcommand{\arraystretch}{0.92}
\newcommand{\MatrixResizeWidth}{\linewidth}
\resizebox{\MatrixResizeWidth}{!}{%
%
}
\end{minipage}
\hfill
\begin{minipage}[t]{0.49\textwidth}
\centering
\caption{Macro average over all unseen tasks -- Avg@8 percentage-point improvement over the base model \texttt{gemma-3-1b-it}}
\label{tab:appendix_crosslingual_gemma_3_1b_it_2_appendix_unseen_task_groups_macro}
\tiny
\setlength{\tabcolsep}{2pt}
\renewcommand{\arraystretch}{0.92}
\newcommand{\MatrixResizeWidth}{\linewidth}
\resizebox{\MatrixResizeWidth}{!}{%
%
%
}
\end{minipage}
\end{table}

\begin{table}[!htbp]
\centering
\begin{minipage}[t]{0.49\textwidth}
\centering
\caption{Average over all unseen tasks from Multilingual Reasoning Gym -- Avg@8 percentage-point improvement over the base model \texttt{gemma-3-1b-it}}
\label{tab:appendix_crosslingual_gemma_3_1b_it_3_appendix_mrg_unseen_macro}
\tiny
\setlength{\tabcolsep}{2pt}
\renewcommand{\arraystretch}{0.92}
\newcommand{\MatrixResizeWidth}{\linewidth}
\resizebox{\MatrixResizeWidth}{!}{%
%
%
}
\end{minipage}
\hfill
\begin{minipage}[t]{0.49\textwidth}
\centering
\caption{Main heldout tasks from Multilingual Reasoning Gym -- Avg@8 percentage-point improvement over the base model \texttt{gemma-3-1b-it}}
\label{tab:appendix_crosslingual_gemma_3_1b_it_4_heldout_filtered}
\tiny
\setlength{\tabcolsep}{2pt}
\renewcommand{\arraystretch}{0.92}
\newcommand{\MatrixResizeWidth}{\linewidth}
\resizebox{\MatrixResizeWidth}{!}{%
%
%
}
\end{minipage}
\end{table}

\begin{table}[!htbp]
\centering
\begin{minipage}[t]{0.49\textwidth}
\centering
\caption{English-content tasks from Multilingual Reasoning Gym, not seen during training -- Avg@8 percentage-point improvement over the base model \texttt{gemma-3-1b-it}}
\label{tab:appendix_crosslingual_gemma_3_1b_it_5_english_content_add}
\tiny
\setlength{\tabcolsep}{2pt}
\renewcommand{\arraystretch}{0.92}
\newcommand{\MatrixResizeWidth}{\linewidth}
\resizebox{\MatrixResizeWidth}{!}{%
%
%
}
\end{minipage}
\hfill
\begin{minipage}[t]{0.49\textwidth}
\centering
\caption{Very difficult tasks from Multilingual Reasoning Gym, not seen during training -- Avg@8 percentage-point improvement over the base model \texttt{gemma-3-1b-it}}
\label{tab:appendix_crosslingual_gemma_3_1b_it_6_difficulty_addback}
\tiny
\setlength{\tabcolsep}{2pt}
\renewcommand{\arraystretch}{0.92}
\newcommand{\MatrixResizeWidth}{\linewidth}
\resizebox{\MatrixResizeWidth}{!}{%
%
%
}
\end{minipage}
\end{table}

\begin{table}[!htbp]
\centering
\begin{minipage}[t]{0.49\textwidth}
\centering
\caption{Easy Math (MGSM + PolyMath low) -- Avg@8 percentage-point improvement over the base model \texttt{gemma-3-1b-it}; $\dagger$: includes only PolyMath low because MGSM is unavailable for this language.}
\label{tab:appendix_crosslingual_gemma_3_1b_it_7_appendix_easy_math_macro}
\tiny
\setlength{\tabcolsep}{2pt}
\renewcommand{\arraystretch}{0.92}
\newcommand{\MatrixResizeWidth}{\linewidth}
\resizebox{\MatrixResizeWidth}{!}{%
%
}
\end{minipage}
\hfill
\begin{minipage}[t]{0.49\textwidth}
\centering
\caption{Hard Math (PolyMath medium, high, and top) -- Avg@8 percentage-point improvement over the base model \texttt{gemma-3-1b-it}}
\label{tab:appendix_crosslingual_gemma_3_1b_it_8_appendix_hard_math_macro}
\tiny
\setlength{\tabcolsep}{2pt}
\renewcommand{\arraystretch}{0.92}
\newcommand{\MatrixResizeWidth}{\linewidth}
\resizebox{\MatrixResizeWidth}{!}{%
%
%
}
\end{minipage}
\end{table}

\FloatBarrier
\subsection{gemma-3-4b-it}

\begin{table}[!htbp]
\centering
\begin{minipage}[t]{0.49\textwidth}
\centering
\caption{Tasks seen during training from Multilingual Reasoning Gym -- Avg@8 percentage-point improvement over the base model \texttt{gemma-3-4b-it}}
\label{tab:appendix_crosslingual_gemma_3_4b_it_1_train_filtered_noheldout}
\tiny
\setlength{\tabcolsep}{2pt}
\renewcommand{\arraystretch}{0.92}
\newcommand{\MatrixResizeWidth}{\linewidth}
\resizebox{\MatrixResizeWidth}{!}{%
%
%
}
\end{minipage}
\hfill
\begin{minipage}[t]{0.49\textwidth}
\centering
\caption{Macro average over all unseen tasks -- Avg@8 percentage-point improvement over the base model \texttt{gemma-3-4b-it}}
\label{tab:appendix_crosslingual_gemma_3_4b_it_2_appendix_unseen_task_groups_macro}
\tiny
\setlength{\tabcolsep}{2pt}
\renewcommand{\arraystretch}{0.92}
\newcommand{\MatrixResizeWidth}{\linewidth}
\resizebox{\MatrixResizeWidth}{!}{%
%
%
}
\end{minipage}
\end{table}

\begin{table}[!htbp]
\centering
\begin{minipage}[t]{0.49\textwidth}
\centering
\caption{Average over all unseen tasks from Multilingual Reasoning Gym -- Avg@8 percentage-point improvement over the base model \texttt{gemma-3-4b-it}}
\label{tab:appendix_crosslingual_gemma_3_4b_it_3_appendix_mrg_unseen_macro}
\tiny
\setlength{\tabcolsep}{2pt}
\renewcommand{\arraystretch}{0.92}
\newcommand{\MatrixResizeWidth}{\linewidth}
\resizebox{\MatrixResizeWidth}{!}{%
%
%
}
\end{minipage}
\hfill
\begin{minipage}[t]{0.49\textwidth}
\centering
\caption{Main heldout tasks from Multilingual Reasoning Gym -- Avg@8 percentage-point improvement over the base model \texttt{gemma-3-4b-it}}
\label{tab:appendix_crosslingual_gemma_3_4b_it_4_heldout_filtered}
\tiny
\setlength{\tabcolsep}{2pt}
\renewcommand{\arraystretch}{0.92}
\newcommand{\MatrixResizeWidth}{\linewidth}
\resizebox{\MatrixResizeWidth}{!}{%
%
%
}
\end{minipage}
\end{table}

\begin{table}[!htbp]
\centering
\begin{minipage}[t]{0.49\textwidth}
\centering
\caption{English-content tasks from Multilingual Reasoning Gym, not seen during training -- Avg@8 percentage-point improvement over the base model \texttt{gemma-3-4b-it}}
\label{tab:appendix_crosslingual_gemma_3_4b_it_5_english_content_add}
\tiny
\setlength{\tabcolsep}{2pt}
\renewcommand{\arraystretch}{0.92}
\newcommand{\MatrixResizeWidth}{\linewidth}
\resizebox{\MatrixResizeWidth}{!}{%
%
%
}
\end{minipage}
\hfill
\begin{minipage}[t]{0.49\textwidth}
\centering
\caption{Very difficult tasks from Multilingual Reasoning Gym, not seen during training -- Avg@8 percentage-point improvement over the base model \texttt{gemma-3-4b-it}}
\label{tab:appendix_crosslingual_gemma_3_4b_it_6_difficulty_addback}
\tiny
\setlength{\tabcolsep}{2pt}
\renewcommand{\arraystretch}{0.92}
\newcommand{\MatrixResizeWidth}{\linewidth}
\resizebox{\MatrixResizeWidth}{!}{%
%
%
}
\end{minipage}
\end{table}

\begin{table}[!htbp]
\centering
\begin{minipage}[t]{0.49\textwidth}
\centering
\caption{Easy Math (MGSM + PolyMath low) -- Avg@8 percentage-point improvement over the base model \texttt{gemma-3-4b-it}; $\dagger$: includes only PolyMath low because MGSM is unavailable for this language.}
\label{tab:appendix_crosslingual_gemma_3_4b_it_7_appendix_easy_math_macro}
\tiny
\setlength{\tabcolsep}{2pt}
\renewcommand{\arraystretch}{0.92}
\newcommand{\MatrixResizeWidth}{\linewidth}
\resizebox{\MatrixResizeWidth}{!}{%
%
}
\end{minipage}
\hfill
\begin{minipage}[t]{0.49\textwidth}
\centering
\caption{Hard Math (PolyMath medium, high, and top) -- Avg@8 percentage-point improvement over the base model \texttt{gemma-3-4b-it}}
\label{tab:appendix_crosslingual_gemma_3_4b_it_8_appendix_hard_math_macro}
\tiny
\setlength{\tabcolsep}{2pt}
\renewcommand{\arraystretch}{0.92}
\newcommand{\MatrixResizeWidth}{\linewidth}
\resizebox{\MatrixResizeWidth}{!}{%
%
%
}
\end{minipage}
\end{table}

\FloatBarrier
\subsection{SmolLM3-3B}

\begin{table}[!htbp]
\centering
\begin{minipage}[t]{0.49\textwidth}
\centering
\caption{Tasks seen during training from Multilingual Reasoning Gym -- Avg@8 percentage-point improvement over the base model \texttt{SmolLM3-3B}}
\label{tab:appendix_crosslingual_smollm3_3b_1_train_filtered_noheldout}
\tiny
\setlength{\tabcolsep}{2pt}
\renewcommand{\arraystretch}{0.92}
\newcommand{\MatrixResizeWidth}{\linewidth}
\resizebox{\MatrixResizeWidth}{!}{%
%
%
}
\end{minipage}
\hfill
\begin{minipage}[t]{0.49\textwidth}
\centering
\caption{Macro average over all unseen tasks -- Avg@8 percentage-point improvement over the base model \texttt{SmolLM3-3B}}
\label{tab:appendix_crosslingual_smollm3_3b_2_appendix_unseen_task_groups_macro}
\tiny
\setlength{\tabcolsep}{2pt}
\renewcommand{\arraystretch}{0.92}
\newcommand{\MatrixResizeWidth}{\linewidth}
\resizebox{\MatrixResizeWidth}{!}{%
%
%
}
\end{minipage}
\end{table}

\begin{table}[!htbp]
\centering
\begin{minipage}[t]{0.49\textwidth}
\centering
\caption{Average over all unseen tasks from Multilingual Reasoning Gym -- Avg@8 percentage-point improvement over the base model \texttt{SmolLM3-3B}}
\label{tab:appendix_crosslingual_smollm3_3b_3_appendix_mrg_unseen_macro}
\tiny
\setlength{\tabcolsep}{2pt}
\renewcommand{\arraystretch}{0.92}
\newcommand{\MatrixResizeWidth}{\linewidth}
\resizebox{\MatrixResizeWidth}{!}{%
%
%
}
\end{minipage}
\hfill
\begin{minipage}[t]{0.49\textwidth}
\centering
\caption{Main heldout tasks from Multilingual Reasoning Gym -- Avg@8 percentage-point improvement over the base model \texttt{SmolLM3-3B}}
\label{tab:appendix_crosslingual_smollm3_3b_4_heldout_filtered}
\tiny
\setlength{\tabcolsep}{2pt}
\renewcommand{\arraystretch}{0.92}
\newcommand{\MatrixResizeWidth}{\linewidth}
\resizebox{\MatrixResizeWidth}{!}{%
%
%
}
\end{minipage}
\end{table}

\begin{table}[!htbp]
\centering
\begin{minipage}[t]{0.49\textwidth}
\centering
\caption{English-content tasks from Multilingual Reasoning Gym, not seen during training -- Avg@8 percentage-point improvement over the base model \texttt{SmolLM3-3B}}
\label{tab:appendix_crosslingual_smollm3_3b_5_english_content_add}
\tiny
\setlength{\tabcolsep}{2pt}
\renewcommand{\arraystretch}{0.92}
\newcommand{\MatrixResizeWidth}{\linewidth}
\resizebox{\MatrixResizeWidth}{!}{%
%
%
}
\end{minipage}
\hfill
\begin{minipage}[t]{0.49\textwidth}
\centering
\caption{Very difficult tasks from Multilingual Reasoning Gym, not seen during training -- Avg@8 percentage-point improvement over the base model \texttt{SmolLM3-3B}}
\label{tab:appendix_crosslingual_smollm3_3b_6_difficulty_addback}
\tiny
\setlength{\tabcolsep}{2pt}
\renewcommand{\arraystretch}{0.92}
\newcommand{\MatrixResizeWidth}{\linewidth}
\resizebox{\MatrixResizeWidth}{!}{%
%
%
}
\end{minipage}
\end{table}

\begin{table}[!htbp]
\centering
\begin{minipage}[t]{0.49\textwidth}
\centering
\caption{Easy Math (MGSM + PolyMath low) -- Avg@8 percentage-point improvement over the base model \texttt{SmolLM3-3B}; $\dagger$: includes only PolyMath low because MGSM is unavailable for this language.}
\label{tab:appendix_crosslingual_smollm3_3b_7_appendix_easy_math_macro}
\tiny
\setlength{\tabcolsep}{2pt}
\renewcommand{\arraystretch}{0.92}
\newcommand{\MatrixResizeWidth}{\linewidth}
\resizebox{\MatrixResizeWidth}{!}{%
%
%
}
\end{minipage}
\hfill
\begin{minipage}[t]{0.49\textwidth}
\centering
\caption{Hard Math (PolyMath medium, high, and top) -- Avg@8 percentage-point improvement over the base model \texttt{SmolLM3-3B}}
\label{tab:appendix_crosslingual_smollm3_3b_8_appendix_hard_math_macro}
\tiny
\setlength{\tabcolsep}{2pt}
\renewcommand{\arraystretch}{0.92}
\newcommand{\MatrixResizeWidth}{\linewidth}
\resizebox{\MatrixResizeWidth}{!}{%
%
%
}
\end{minipage}
\end{table}

\FloatBarrier

\section{Base Model Performance}
\label{app:base-model-performance}

\paragraph{Reading the tables.}
Each cell reports the untrained base model's avg@8 accuracy for the task group and evaluation language, in percentage points. The final Avg column reports the mean over displayed evaluation-language columns. Cells in each column are colored relative to the base models in that table, and bold entries mark the best value in the column.

\paragraph{Macro definitions.}
Easy Math is the macro average of MGSM and PolyMath-low. Hard Math is the macro average of PolyMath-medium, PolyMath-high, and PolyMath-top. Average over all unseen tasks from Multilingual Reasoning Gym (MRG) is the equal-weight macro average of MRG heldout, English-content, and very difficult MRG tasks. Macro average over all unseen tasks is the equal-weight macro average of MRG all-unseen, Easy Math, and Hard Math.

\begin{table}[!htbp]
\centering
\caption{Base model performance on tasks seen during training from Multilingual Reasoning Gym.}
\vspace{-2mm}\label{tab:base_model_performance_1_mrg_train}
\scriptsize
\setlength{\tabcolsep}{3pt}\renewcommand{\arraystretch}{0.92}
\renewcommand{\arraystretch}{0.92}
\resizebox{0.7\textwidth}{!}{%
%
}
\end{table}

\begin{table}[!htbp]
\vspace{-1mm}

\centering
\caption{Base model performance on a macro average over all unseen tasks.}
\vspace{-2mm}\label{tab:base_model_performance_2_all_unseen_macro}
\scriptsize
\setlength{\tabcolsep}{3pt}
\renewcommand{\arraystretch}{1.05}
\resizebox{0.7\textwidth}{!}{%
%
%
}
\end{table}

\begin{table}[!htbp]
\vspace{-1mm}

\centering
\caption{Base model performance on the average over all unseen tasks from Multilingual Reasoning Gym.}
\vspace{-2mm}\label{tab:base_model_performance_3_mrg_unseen_macro}
\scriptsize
\setlength{\tabcolsep}{3pt}\renewcommand{\arraystretch}{0.92}
\renewcommand{\arraystretch}{1.05}
\resizebox{0.7\textwidth}{!}{%
%
%
}
\end{table}

\begin{table}[!htbp]
\vspace{-1mm}

\centering
\caption{Base model performance on the main heldout tasks from Multilingual Reasoning Gym.}
\vspace{-2mm}\label{tab:base_model_performance_4_mrg_heldout_8}
\scriptsize
\setlength{\tabcolsep}{3pt}\renewcommand{\arraystretch}{0.92}
\renewcommand{\arraystretch}{1.05}
\resizebox{0.7\textwidth}{!}{%
%
%
}
\end{table}

\begin{table}[!htbp]
\vspace{-1mm}

\centering
\caption{Base model performance on English-content tasks from Multilingual Reasoning Gym.}
\vspace{-2mm}\label{tab:base_model_performance_5_mrg_english_content}
\scriptsize
\setlength{\tabcolsep}{3pt}\renewcommand{\arraystretch}{0.92}
\renewcommand{\arraystretch}{1.05}
\resizebox{0.7\textwidth}{!}{%
%
%
}
\end{table}

\begin{table}[!htbp]
\vspace{-1mm}

\centering
\caption{Base model performance on very difficult tasks from Multilingual Reasoning Gym.}
\vspace{-2mm}\label{tab:base_model_performance_6_mrg_difficulty_addback}
\scriptsize
\setlength{\tabcolsep}{3pt}\renewcommand{\arraystretch}{0.92}
\renewcommand{\arraystretch}{1.05}
\resizebox{0.7\textwidth}{!}{%
%
%
}
\end{table}

\begin{table}[!htbp]
\vspace{-1mm}

\centering
\caption{Base model performance on Easy Math (MGSM + PolyMath-low); $\dagger$: includes only PolyMath-low.}
\vspace{-2mm}\label{tab:base_model_performance_7_easy_math}
\scriptsize
\setlength{\tabcolsep}{3pt}\renewcommand{\arraystretch}{0.92}
\renewcommand{\arraystretch}{1.05}
\resizebox{0.7\textwidth}{!}{%
%
%
}
\end{table}

\begin{table}[!htbp]
\vspace{-1mm}

\centering
\caption{Base model performance on Hard Math (PolyMath medium, high, and top).}
\vspace{-2mm}\label{tab:base_model_performance_8_hard_math}
\scriptsize
\setlength{\tabcolsep}{3pt}\renewcommand{\arraystretch}{0.92}
\renewcommand{\arraystretch}{1.05}
\resizebox{0.7\textwidth}{!}{%
%
%
}
\end{table}

\FloatBarrier

\applefootnote{ \textcolor{textgray}{\sffamily Apple and the Apple logo are trademarks of Apple Inc., registered in the U.S. and other countries and regions.}}

\end{document}